\documentclass[preprint,12pt]{elsarticle}

\usepackage[table,xcdraw]{xcolor}
\usepackage{amssymb}
\usepackage{graphicx}
\usepackage{multirow}
\usepackage{makecell}
\usepackage{enumerate}
\usepackage[shortlabels]{enumitem}
\usepackage{xcolor}
\usepackage{mathrsfs}

\setenumerate[1]{itemsep=0pt,partopsep=0pt,parsep=\parskip,topsep=5pt}
\setitemize[1]{itemsep=0pt,partopsep=0pt,parsep=\parskip,topsep=5pt}
\setdescription{itemsep=0pt,partopsep=0pt,parsep=\parskip,topsep=5pt}
  
\usepackage{array}
\usepackage{booktabs} %调整表格线与上下内容的间隔
\usepackage{hyperref}
\usepackage{amsthm}
\usepackage{amsmath} 
\usepackage{float}
\usepackage{bm}
\usepackage[multiple]{footmisc}
\biboptions{comma,square,sort&compress}

\journal{Artificial Intelligence}
\journalissue{Artificial Intelligence Special Issue on XAI}

\begin{document}

\begin{frontmatter}



\title{Show or Suppress? Managing Input Uncertainty in Machine Learning Model Explanations}


\author{Danding Wang}
\ead{wangdanding@u.nus.edu}
\author{Wencan Zhang}
\ead{wencanz@u.nus.edu}
\author{Brian Y. Lim}
\ead{brianlim@comp.nus.edu.sg}
\address{School of Computing, National University of Singapore, Singapore}

\begin{abstract}
\label{section:abstract}
Feature attribution is widely used in interpretable machine learning to explain how influential each measured input feature value is for an output inference. However, measurements can be uncertain, and it is unclear how the awareness of input uncertainty can affect the trust in explanations. We propose and study two approaches to help users to manage their perception of uncertainty in a model explanation: 1) transparently show uncertainty in feature attributions to allow users to reflect on, and 2) suppress attribution to features with uncertain measurements and shift attribution to other features by regularizing with an uncertainty penalty. Through simulation experiments, qualitative interviews, and quantitative user evaluations, we identified the benefits of moderately suppressing attribution uncertainty, and concerns regarding showing attribution uncertainty. This work adds to the understanding of handling and communicating uncertainty for model interpretability.
\end{abstract}


% \begin{highlights}
% \end{highlights}

\begin{keyword}
Trust \sep Uncertainty \sep Interpretable Machine Learning



\end{keyword}

\end{frontmatter}


\section{Introduction}
\label{section:Introduction}
\begin{figure}
    \centering
    \includegraphics[width=13cm]{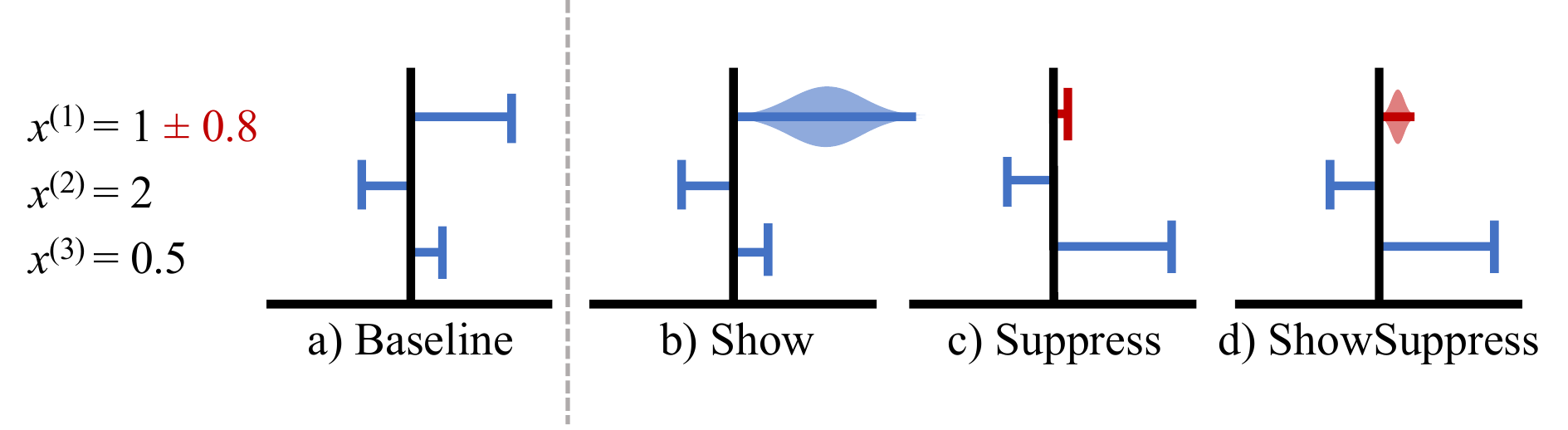}
    \caption{Baseline attribution explanation and three proposed attribution explanations that account for uncertainty. Each horizontal bar in a tornado plot is a feature attribution, and the sum of all bars indicates the total attribution for the model output. Suppose uncertainty in input $x^{(1)}$ is known to be large. This will propagate through the model and can be accounted in its attribution as: a) Baseline explanation which does not handle the uncertainty in $x^{(1)}$; b) Show which visualizes the attribution uncertainty (in this case, with a violin plot); c) Suppress which reduces the attribution to $x^{(1)}$ and reallocates attribution to other features  (in this case, $x^{(3)}$); d) ShowSuppress which combines Show and Suppress to suppress attributions towards uncertain features and shows the suppressed attribution uncertainty.}
    \label{fig:sample}
\end{figure}

The increasing prevalence of machine learning (ML) has called attention to make it more transparent and trustworthy \cite{Lipton2018TheInterpretability}. The burgeoning research area of Explainable AI (XAI) provides a basis to improve understanding and trust in ML model inferences \cite{Abdul2018TrendsAgenda,datta2016algorithmic,doshi2017towards,Lipton2018TheInterpretability}. Explaining with feature attributions (also called attribution explanations) is one popular technique to interpret machine learning models (e.g., \cite{Lundberg2017APredictions,Ribeiro2016WhyClassifier,bach2015pixel}) by attributing the model's inference to input features. This indicates whether a feature influences a decision outcome positively or negatively and by how much. An attribution explanation typically involves approximating a linear relationship between input and output, which can also be considered a weighted sum, and is commonly visualized as a tornado plot (see Baseline in Figure \ref{fig:sample}). 
Each bar indicates the influence of the feature on the model's inference. Bars to the right indicate positive influence and bars to the left indicate negative influence; longer bars indicate larger attribution. This intuitive explanation technique illustrates how each feature contributes to the model inference, using basic bar chart literacy.\par

%The increasing prevalence of machine learning (ML) in society has called attention to make it more trustworthy \cite{Lipton2018TheInterpretability}. The burgeoning research area of Explainable AI provides a basis to improve understanding and trust in ML model inferences \cite{Abdul2018TrendsAgenda,datta2016algorithmic,doshi2017towards,Lipton2018TheInterpretability}. Explaining with feature attributions is a popular technique to interpret machine learning models (e.g., \cite{Lundberg2017APredictions,Ribeiro2016WhyClassifier}) by attributing model explanation to relevant input features. This typically involves approximating a linear relation between input and output, and presenting the explanation as a tornado plot (see Baseline in Figure \ref{fig:sample}), where longer bars indicate higher attribution or influence for that feature. This allows users to understand which features are more important for a given instance inference.

Attribution explanation techniques focus on conveying salient signals and assume that the input feature values are reliable. In contrast, input measurements are often noisy or uncertain \cite{czarnecki2013machine}, e.g., a temperature reading at a regular time may be affected by a temporary gust of wind, or a sensor may momentarily drop data due to network communication issues. With these fluctuating noisy inputs, the model inference may also fluctuate, and may correspond explanations of these inferences. Thus, attribution explanations should account for and communicate the impact of this uncertainty on users. In this paper, we present two approaches for attribution explanations to be “uncertainty-aware” to leverage uncertainty information to augment explanation visualization or re-compute the attributions. 
We are specifically interested in the research questions: In what circumstances will a user trust or not trust an explanation that is based on a set of measurements that are of questionable validity? How can we manage to communicate input uncertainty at inference time to improve trust in explainable AI? \par
Communicating uncertainty has been an active research topic in supporting people's use of intelligent systems, such as investigating the impact of showing uncertainty in estimated values \cite{Rukzio2006VisualizationApplications} in various simpler formats \cite{Kay2016WhenSystems}, creating novel visualizations for uncertainty \cite{hullman2015hypothetical,Kale2019HypotheticalData}, and communicating uncertainty in model inference \cite{Lim2011InvestigatingApplications,Lipton2018TheInterpretability}. These approaches demonstrate that showing uncertainty helps to improve user trust in automation and smart systems. We extend the prior research by communicating the uncertainty in the attribution explanation. Specifically, we first propose a Show Uncertainty explanation which propagates the uncertainty of inputs to attribution explanation and visualizes the attribution uncertainty (Figure \ref{fig:sample}, Show). Interestingly, while showing uncertainty allows users to reflect and calibrate their decisions, some people have an aversion to uncertainty \cite{Wollard2012ThinkingSlow, simpkin2016tolerating, Boukhelifa2017HowStudy, Ellsberg1961RiskAxioms}, and tend to cope with uncertainty by ignoring, denying, or reducing it by acquiring more information \cite{Lipshitz1997CopingAnalysis}. To support them, we propose an alternative Suppress Uncertainty explanation (Figure \ref{fig:sample}, Suppress), which reduces the attributions to inputs with high uncertainty to make the explanation less dependent on uncertain information and shift attributions to more certain inputs instead. The interaction design with Suppressed Uncertainty explanations is as follows: i) the user will see that there is uncertainty in some input feature values, ii) be shown an attribution explanation and be informed that the attributions have been suppressed to be less reliant on uncertain features, iii) and be able to toggle the explanation view to see the unsuppressed attribution explanations. From this comparison, the user will learn how attributions of uncertain features have been reduced in magnitude, and partially reallocated to other features. For a wine quality score rating example, due to an input uncertainty of $\pm$1.0\% for Alcohol, its attribution is reduced from +2.5 (Figure \ref{fig:walk_through}c) to +1.3 (Figure \ref{fig:walk_through}b); similarly an input uncertainty of $\pm$0.2 for Vinegar Taint led to a reduction in its attribution from $-$1.2 to $-$0.5; the reduced attributions have been reallocated from $-$0.4, 0.5, 0.9 to $-$0.3, 0.7, 1.4 for pH, Sulphates, SO2, respectively. To suppress uncertainty, we propose two technical methods that regularize the model-agnostic explainer or the predictor model itself. Finally, combining two strategies of showing and suppressing, we propose the ShowSuppress Uncertainty explanation  (Figure \ref{fig:sample}, ShowSuppress) that first suppresses attribution to uncertain inputs and visualizes any remaining attribution uncertainty. Note that suppressing uncertainty (Suppress) involves shifting the attribution values in the explanation to minimize the influence of inputs with high uncertainty. This does not mean that uncertainty is concealed; the attribution uncertainty can still be shown (ShowSuppress), but they will be smaller than before (Show). This is different from simply not showing attribution uncertainty, which could be the default (Baseline) or with uncertainty suppressed (Suppress).\par

Given the opposing objectives of these uncertainty management techniques, we investigated the relative benefits and limitations of showing or suppressing uncertainty in explanations compared to baseline attribution explanations. We define a stochastic metric --- \textit{Expected Faithfulness Distance} --- based on Explanation Faithfulness\footnote{Faithfulness is the similarity in predictions between the predictor model and explanation (explainer) model that explains the predictor.} \cite{Ribeiro2016WhyClassifier} as a measure of the how poorly (well) an explanation globally agrees with the predictor model when explaining instances with uncertain measurements. In a simulation study, we show how Suppress can reduce Expected Faithfulness Distance. In a qualitative interview study, we learn how participants variously used the different explanation techniques (Baseline, Show, Suppress, ShowSuppress), appreciated the information they provided, but had divergent opinions about their usefulness. In a controlled quantitative user study, we evaluated whether and by how much each explanation technique affected decision quality, confidence, trust in the model predictions, and perceived helpfulness of the explanations. We found that showing uncertainty helps users to understand the system but costs more time, and the decision of users with higher uncertainty tolerance is closer to the system when showing uncertain; and suppressing attribution uncertainty increases explanation faithfulness, user trust, confidence, and decision quality. In summary, our contributions are:
\begin{itemize}
    \item Three approaches to manage and communicate attribution uncertainty due to input uncertainty: 1) showing attribution uncertainty, 2) suppressing uncertainty by regularizing explainer or predictor models, 3) and a hybrid of showing and suppressing.
    \item Experiment methods and instruments to evaluate interpretable ML models under uncertainty by employing stochastic metrics, such as expected faithfulness of hypothetically uncertain instances, and a within-subjects experiment design to evaluate explanation-assisted decision making with input, attribution, and model uncertainties. 
    \item Empirical findings from i) a characterization simulation study, ii) formative qualitative interviews, and iii) a summative quantitative user study to show a) how showing attribution uncertainty improves understanding and trust, but is not helpful to improve decision making, and b) suppressing attribution due to uncertainty improves trust, helpfulness and decision making.
\end{itemize}

This work adds to the research on explainable AI under uncertainty and highlights the importance to manage and communicate attribution uncertainty to users. From the quantitative and qualitative results of our user studies, we identified the benefits of moderately suppressing attribution uncertainty, and concerns regarding showing attribution uncertainty.\par

\section{Related Work}
\label{section:Related_Work}
This work studies the implication of considering uncertainty within explanations of machine learning (ML) models. First, we define the scope of uncertainty that we study, point out that current interpretable ML techniques do not handle uncertainty, and summarize existing methods to handle uncertainty by showing or suppressing it.\par
There are different concepts and aspects of uncertainty in machine learning: stochasticity and insufficiency in training data \cite{Kendall2017WhatVision}, possible deviations of a model output \cite{Kay2016WhenSystems}, probabilistic models with uncertain model parameters \cite{Friedman1997BayesianAlgorithms} and noisy inputs to the model during inference \cite{Abdelaziz2015UncertaintyNetworks}. In this paper, we focus on the last source of uncertainty from noisy input data; this is usually caused by measurement and estimation error. We focus on how input uncertainty should be handled during explaining model inference rather than during explaining model training. Although there are some models that intrinsically modeling or calculating input uncertainty, we focus on more prevalent models that do not model uncertainty and yet face input uncertainty in model inference time. Our paper handles attribution uncertainty due to uncertain input and differentiates from works that identified or handled attribution uncertainty due to the instability of explanation generation process \cite{Guidotti2019OnModels,gosiewska2019not,zhang2019should,zafar2019dlime}.

We note that there may be different reasons for uncertainty in data, such as miscalibrated sensors, lost data connection, adverse or unexpected environmental changes. Determining these causes is beyond the scope of the machine learning modeling we discuss in this paper.

\subsection{Improving Trust With Intelligible, Interpretable Models}
\label{subsection:improving_trust}
The recent interest in explainable AI and interpretable machine learning has given rise to many types of interpretable models and model explainers, e.g. \cite{Caruana2015IntelligibleReadmission,kim2018interpretability,koh2017understanding,Krause2016InteractingModels,Lundberg2017APredictions,Ribeiro2016WhyClassifier}. Typically, attribution explanation generates feature importance by attributing model output to input features. Local Interpretable Model-agnostic Explanation (LIME) \cite{Ribeiro2016WhyClassifier} is a popular method that generates an instance-level attribution explanation by approximating predictor model locally around the query instance with a linear model as an explainer. Besides the model-agnostic attribution explanation LIME, Integrated Gradients (IG) \cite{Sundararajan2017AxiomaticNetworks} is an attribution explanation specific to explaining neural networks by integrating the gradient of model output along each input dimension. However, these explanation methods assume that there is no error in the input features and do not consider the attribution uncertainty. Therefore, we propose approaches to providing uncertainty-aware attribution explanations by showing or suppressing strategies. In this paper, for the model-agnostic explanation, we extend LIME by visualizing and regularizing explanation 6 by input uncertainty. To make model-specific attribution explanation uncertainty-aware, we show an example with IG. \par
%However, these models and explainers assume that there is no error in the input feature values. Recently, methods to consider model stability and robustness \cite{doshi2017towards,gosiewska2019ibreakdown,koh2017understanding,Zhang2019WhyExplanations} and to integrate Bayesian inference \cite{Kendall2017WhatVision} can account for uncertainty, but these typically focus on model uncertainty and not explanation uncertainty. Furthermore, they are not widely applied or adopted. 
The aforementioned works focus on technical innovation to generate explanations, but it is important to ensure that explanations are human-centered to support human reasoning \cite{Abdul2018TrendsAgenda,Lipton2018TheInterpretability,Wang2019DesigningAI}, and be integrated into the iterative design pipeline \cite{Eiband2018BringingPractice}. Many insights can be learned from observing how users use and interact with explanations. Lim and Dey investigated how users interacted with different query types to seek information and learned how little time users spent on consuming explanations \cite{Lim2011DesignApplication}. In a design probe study with data scientists, Hohman et al. identified the importance of supporting interaction in explanations \cite{Hohman2019Gamut:Models}, and in a lab study, Cheng et al. found that interactive explanations help to improve user understanding \cite{Cheng2019ExplainingStakeholders}. Cai et al. identified what concepts pathologists care in medical images and explains the model by providing similar images based on the similarity of user-defined concepts \cite{Cai2019Human-centeredDecision-making}. Wang et al. explored how tailored explainable AI can help clinicians to improve medical decision making by mitigating specific cognitive biases \cite{Wang2019DesigningAI}.\par
In this work, we employ a quantitative empirical controlled user experiments and conduct a qualitative interview to study how users interact with model explanations under input uncertainty.\par

\subsection{Handling Uncertainty by Showing}
\label{subsection:uncertainty_by_show}
Researchers have found that showing inference uncertainty, confidence or accuracy can affect user trust and task performance \cite{Antifakos2005TowardsConfidence,Guidotti2019OnModels,Lipton2018TheInterpretability,Rukzio2006VisualizationApplications}. Antifakos et al. \cite{Antifakos2005TowardsConfidence} found that showing uncertainty can decrease task time when uncertainty is low, while Rukzio et al. \cite{Rukzio2006VisualizationApplications} found that this can hurt task time. Lim and Dey \cite{Lim2011InvestigatingApplications} found that low uncertainty can raise user trust, but showing high uncertainty will lower trust, even if the model behavior is correct. Yin et al. \cite{Yin2019UnderstandingModels} found that communicating the model stated and observed accuracy can have a compounded effect on trust. \par
Many studies communicate uncertainty by showing a numeric score, or ordinal labels (high/low text, traffic light icons, etc.). However, uncertainty can be complex, and several visualization methods have been proposed to display them in interfaces and charts \cite{Bica2019CommunicatingDiffusion,Jung2015DisplayedCar,Kay2013TheresInterface,Sacha2016TheAnalytics}. Hypothetical outcome plots (HOP) help users to understand stochastic events by animating instances on a graph, sampled at random based on its uncertainty distribution \cite{hullman2015hypothetical}. Kay and colleagues found that quantile dot plots help to improve user perception of probabilities and help with decision making \cite{Fernandes2018UncertaintyDecision-making,Kay2016WhenSystems}.

\subsection{Handling Uncertainty by Suppression}
\label{subsection:uncertainty_by_suppress}
The aforementioned studies identified cases where showing uncertainty is useful, but can also sometimes be harmful. Indeed, it may be better to suppress uncertainty, since some people have lower tolerances and poorer coping strategies regarding uncertainty \cite{Lipshitz1997CopingAnalysis}. Even domain experts, such as data scientists who are experts in data uncertainty \cite{Boukhelifa2017HowStudy} and clinicians regularly make decisions under uncertainty \cite{simpkin2016tolerating} seek to reduce uncertainty. Lipshitz and Strauss \cite{Lipshitz1997CopingAnalysis} identified several uncertainty-coping strategies such as acknowledging, reducing and suppressing. They found that people acknowledge uncertainty, seek more information to reduce uncertainty, and ignore or deny uncertainty. These works suggest that suppressing uncertainty in explanation can better help the users who are not comfortable with uncertainty in decision making. The suppressing technique is inspired by the Dempster–Shafer evidence theory \cite{shafer1976mathematical}. When sources of evidence (input features) have different reliability (various input uncertainty), to combine the effects of unequal-reliable evidence, there are several methods \cite{Yang2013DiscountedDisagreement,klein2010automatic,schubert2011conflict} that discount the effects of unreliable evidence. In the domain of interpretable machine learning, no research has been found to use an uncertainty suppression strategy to support users who are uncertainty intolerant. Our paper aims to fill this gap by proposing a Suppressed Uncertainty Explanation and evaluating how it affects user trust of machine learning and confidence in decision making.\par
%% Correll et al. \cite{Correll2018Value-SuppressingPalettes} proposed a Value-Suppressing Uncertainty Palettes that suppresses the visual channel by uncertainty. They found that users tended to make risk-avoiding decisions when they used the suppressed uncertainty visualization.  
Some recent researches on explanation robustness relate to our uncertainty suppression approach, but they vary in their scope and objectives. Ghorbi et al. \cite{Ghorbani2019InterpretationFragile} demonstrated that attribution explanations of deep learning are sensitive to adversary perturbation. Handling this, several works aimed to improve the model and explanation robustness. Chen et al. \cite{Chen2019RobustRegularization} improved robustness by adding a regularization term to the training loss function to penalize the difference between attribution explanations of neighboring instances; this makes explanations of neighboring instances consistent, no sudden changes. Singh et al. \cite{singh2019benefits} improved robustness against adversarial attacks by regularizing their model training loss to penalize differences between attribution explanations of instances and nearby adversarial examples. In contrast, we desensitize the attribution explanation of each instance by adding a regularization term to the training loss function to penalize attributions to input features with high uncertainty; this makes explanations less dependent on features with uncertainty, because attributions to these features will be reduced in magnitude. Furthermore, our aim is not to improve the robustness of a certain type of model, but handling attribution uncertainty to support people who are uncertainty intolerant. Model robustness is a by-product of our Regularized Predictor method. These methods also suppress attribution uncertainty by regularization, but the difference is that they suppress the uncertainty in explanation generation while we suppress the attribution uncertainty due to input uncertainty.\par
In summary, we focus on the uncertainty in machine learning model explanations due to uncertain inputs, and we aim to extend the research field of human-centered explainable AI by investigating how attribution uncertainty facilitates user understanding of model inference and uncertainty and how attribution uncertainty affects user decision making, trust and confidence. Drawing on and extending prior work from the perception of uncertainty, uncertainty visualization, and regularization in machine learning, we propose two uncertainty handling approaches, showing and suppressing attribution explanation uncertainty, for users with different uncertainty tolerance.

\section{Technical Approach}
\label{section:Technical_Approach}
% We extend the popular model-agnostic feature attribution explanation technique, LIME \cite{Ribeiro2016WhyClassifier}, to develop two comparable and compatible approaches to manage uncertainty in feature attribution explanations. This allows us to explicitly compare the two key techniques – Visualized Uncertainty and Regularized Uncertainty, and yet combine them into a hybrid Regularized+Visualized Uncertainty technique. 
In this section, we introduce how we show attribution explanation uncertainty, henceforth called attribution uncertainty, by sampling hypothetical instances and how we suppress uncertainty by regularization in both model-agnostic and model-specific explanations. For the model-agnostic explanation, we base our two uncertainty-aware explanation techniques on LIME \cite{Ribeiro2016WhyClassifier}, a popular model-agnostic attribution explanation method. This allows post-hoc explanations to be made uncertainty-aware without needing to retraining the predictor model. We first briefly introduce LIME, focusing on how it can provide linear attribution explanations, then describe our method to transparently visualize uncertainty in attributions (Show Uncertainty), and describe how we added a regularization term to penalize on input uncertainty (Suppress Uncertainty). These techniques can be combined to provide the hybrid ShowSuppress explanation. While using model-agnostic explanations support convenient implementation and wider adoption of explanations, making model-specific explanations can further improve explanation faithfulness. Thus, we also propose regularizing the training of the predictor model by input uncertainty to suppress attribution uncertainty, and we show an example of a regularized neural network model explained by Integrated Gradients \cite{Sundararajan2017AxiomaticNetworks}.

\subsection{Instance-Based Attribution Explanation}
\label{subsection:attribution_explanation}
LIME \cite{Ribeiro2016WhyClassifier} is an instance-based, model-agnostic explainer $g_f(\cdot)$ that explains the inference for each single instance $x_0$ without needing to know the internal mathematical mechanics of the underlying predictor model $f(\cdot)$. The explainer $g_f(\cdot)$, usually a linear model, is trained by sampling data points in the neighborhood of the query instance $x_0$ to faithfully describe the inferred outcome from predictor $f$.  Formally, the linear explainer model for LIME is simply defined as 
\begin{equation}
    g_f\left(x\right)={w}^{\top} x=\sum_{d} {w}^{(d)} x^{(d)} ,
\end{equation}
where $x^{(d)}$ is the $d$-th feature value, and $w^{(d)}$ is its linear weight.  $w^{(d)}x_0^{(d)}$ is the feature attribution that indicates how influential the $d$-th feature is for inference on instance $x_0$. These feature attributions are commonly visualized in a tornado plot (see Figure \ref{fig:sample}, Baseline), which are shown to be good for visualizing the influence of multiple variables \cite{Eschenbach1992SpiderplotsAnalysis}. In this paper, we follow the established use \cite{Kulesza2015PrinciplesLearning,Lim2011DesignApplication,Lim2011InvestigatingApplications,Ribeiro2016WhyClassifier} of tornado plots for feature attribution explanations. To train the local explainer, a training dataset $s_{x_0}$ is first constructed by sampling in the neighborhood of the query instance $x_0$. Neighbors that are closer to $x_0$ will have a higher influence on the explainer training, encoded as weight $c_{x_0}$. LIME minimizes the difference between predictor outputs and explainer outputs in the neighborhood subset, which means the linear explainer model should be faithful to the predictor model within the neighborhood of the query instance, and the explainer should have consistent inference with the predictor:

\begin{equation}
     \sum_{x \in s_{x_0}} c_{x_0}(x) \cdot \left(f(x)- g_f(x)\right)^{2}+\lambda\|{w}\|_{2}^{2} ,
\label{eq:LIME}
\end{equation}
where $\lambda \left\| {w} \right\|_2^2$ is the L2-norm regularization on the weights ${w}$ of the explainer to control for sparsity. Higher $\lambda$ makes the explainer model sparser with smaller feature attributions. Next, we describe two approaches to manage uncertainty with LIME.

\subsection{Showing Uncertainty in Attribution Explanation}
\label{subsection:Show_in_attribution}
Consider feature $x_0^{(d)}$ with input uncertainty $\epsilon^{(d)}$, i.e., $x_0^{(d)}\pm\epsilon^{(d)}$. We aim to propagate this uncertainty to the feature attribution in the linear model explanation, i.e., $w^{(d)} \cdot (x_0^{(d) }\pm\epsilon^{(d)})$. We generate hypothetical instances using Monte-Carlo Sampling on the probability distribution $P(\epsilon)$ of the error $\epsilon$ and compute the feature attribution for each hypothetical instance. We repeat this for each feature with uncertainty.\par
To illustrate the distribution of attributions to the readers, we use violin plots \cite{Hintze1998ViolinSynergism} to visualize the distribution of uncertainty in attributions. Figure \ref{fig:sample} b) shows an example of Visualized uncertainty as a violin plot for each feature in a tornado plot, specifically, the uncertainty in the attribution for the first feature. Instead of using a T-shape bar to indicate a single-point value for attribution, the violin plot illustrates the probability density of attribution. 

\subsection{Suppressing Attribution Uncertainty in the Explainer}
\label{subsection:Suppress_in_explainer}
\begin{figure}[ht]
     \centering
     \includegraphics[width=13.5cm]{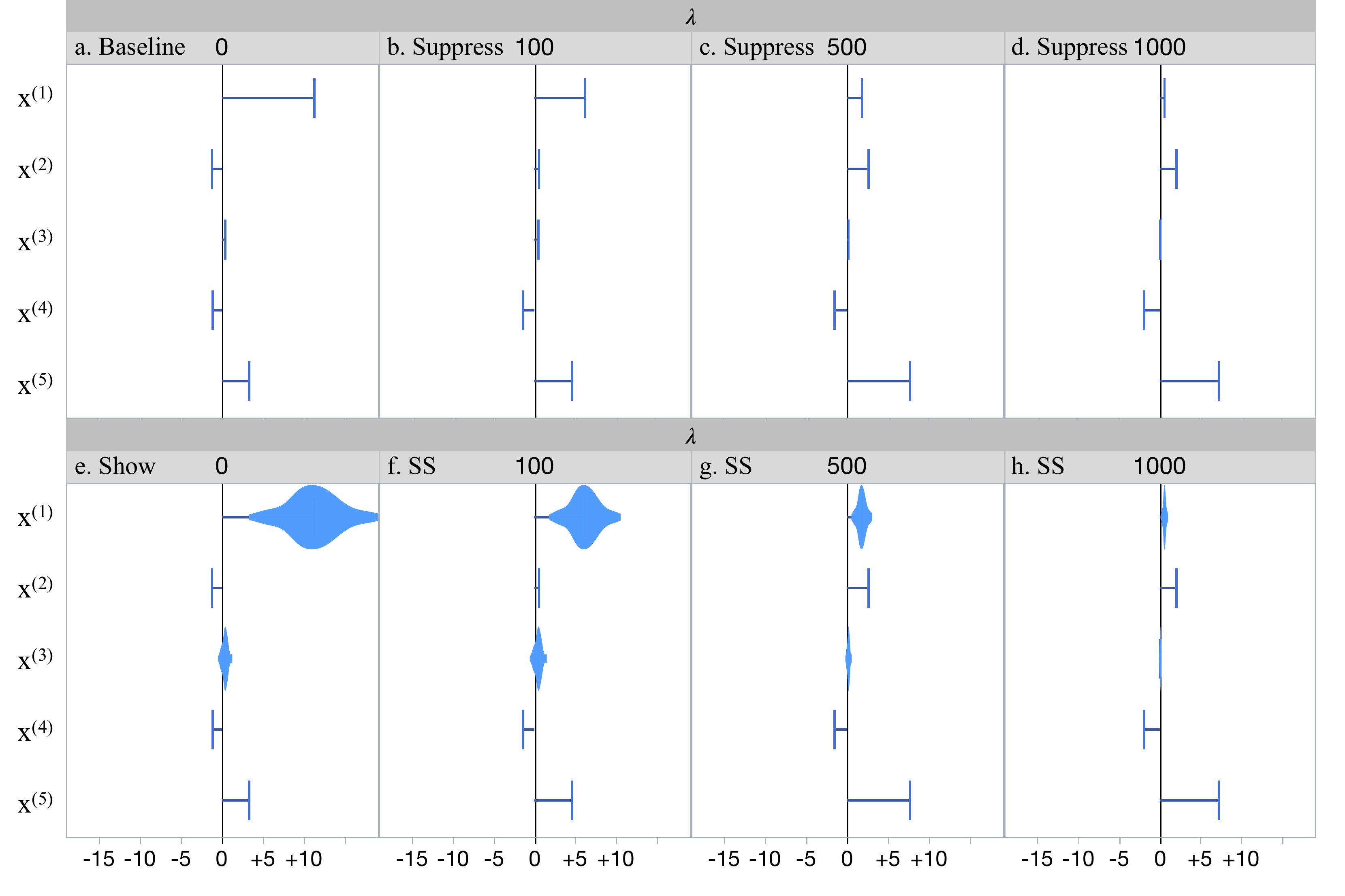}
     \caption{Example of Baseline, Show, Suppress and ShowSuppress (SS) explanations. The attribution explanations in the first row do not show attribution uncertainty while the second row visualizes attribution uncertainty in violin plots. In this case, uncertainty only occurs in features $x^{(1)}$ and $x^{(3)}$, with $x^{(1)}$ having a higher uncertainty. When regularization weight $\lambda$ is zero, no feature attribution is suppressed and the attribution uncertainty is large (see a.Baseline and e.Show). As $\lambda$ increases, the attributions to feature $x^{(1)}$ and $x^{(3)}$ are re-attributed to other features (See Suppress explanations in b,c and d). The bottom row ShowSuppress explanations illustrate that the uncertainty in explanation is suppressed when $\lambda$ increases.}
     \label{fig:adjust_lambda}
\end{figure}

% Adjusting $\lambda$ to suppress explanation uncertainty to different levels. In this case, uncertainty only occurs in features $x^{(1)}$ and $x^{(3)}$, with $x^{(1)}$ having a higher uncertainty. When $\lambda$ equals to zero, no feature attribution is suppressed and the explanation uncertainty is large (see a.Baseline and e.Vis). As $\lambda$ increases, the attributions from feature $x^{(1)}$ and $x^{(3)}$ will be re-attributed to other features (See Reg explanations b,c and d). The explanation uncertainty is also suppressed when $\lambda$ increases, and the bottom row Vis and Reg+Vis explanations show this change.

To reduce the attribution uncertainty, we propose to suppress attribution due to uncertain features in the explainer model. This is not simply concealing the attribution uncertainty but changing the explanation to be less dependent on uncertain features. Specifically, we apply an additional regularization penalty\footnote{Regularization: adding a secondary objective to the training loss function to train the machine learning model so that it performs better with respect to the secondary objective. In our case, we penalize the model from depending too much on inputs with high uncertainty, therefore, we added a loss term regarding input uncertainty.} on the LIME explainer to suppress the attribution uncertainty. This Regularized Explainer $\tilde{g}_f(\cdot)$ decreases the attribution to uncertain features, but this causes a shift to increase attributions to other features to maintain the explainer's inference faithfulness to predictor output. \par
We adapt the objective function in Eq \ref{eq:LIME} for training the explainer to include penalizing input uncertainty:
\begin{equation}
\label{eq:reg_expl}
    \sum_{x \in s_{x_0}} c_{x_0}(x) \cdot \left(f(x)- \tilde{g}_f(x)\right)^{2}+ \overbrace{\lambda\|{\sigma \circ w}\|_{2}^{2}}^\text{\shortstack{Attribution Uncertainty\\penalty}}
\end{equation}
where $\lambda\|{\sigma \circ w}\|_{2}^{2}$ is the weighted L2-norm regularization term that penalizes weights if there is uncertainty in their corresponding input, $\sigma$ is the input uncertainty vector whose $d$-th element is the standard deviation $\sigma^{(d)}$ of the uncertainty $\epsilon^{(d)}$ in $x^{(d)}$, $w$ is the weights vector whose $d$-th element is the linear weight for element $x^{(d)}$, and $\circ$ is the Hadamard product operator. Any larger uncertainty $\sigma^{(d)}$ will penalize the corresponding weight $w^{(d)}$, and higher regularization weight $\lambda$ strengthens the penalization. \par
Figure \ref{fig:adjust_lambda} illustrates the suppression and shifting of attribution due to different $\sigma^{(d)}$ and $\lambda$ values\footnote{The selection of $\lambda$ can be personalized to user preference. It can also be selected by expected faithfulness distance (see Section \ref{sec:exp_faith} and \ref{append:other_sim})}. In this example, since $\sigma^{(1)}\textgreater \sigma^{(3)}\textgreater 0$, $w^{(1)}$ is suppressed more than $w^{(3)}$. The total amount of attributions is conserved during the process to maintain the explanation output faithfulness to the predictor output. The suppressed explanation will be less dependent on uncertain features and less sensitive to changes in the uncertain features, so the explainer will be more robust to uncertainty. This can mitigate users' worry about uncertainty since the explained result remain similar even if a future measurement is noisily different.\par
Since LIME is a model-agnostic explainer that only requires input and output data from the predictor to generate explanation, modifying the LIME explainer will not affect the predictor. The benefit of this is that users can suppress attribution uncertainty only with the access to querying the predictor inference instead of modifying or retraining the predictor. However, an inherent weakness of LIME is that it is a post-hoc estimation of the model, and regularizing post-hoc explanation does not change the underlying predictor behavior, thus the predictor may still be sensitive to input uncertainty. There could be a discrepancy between the uncertainty-aware Regularized Explainer and the underlying predictor. Next, to address this issue, we propose to handle this discrepancy by regularizing predictor directly.

\subsection{Suppressing Attribution Uncertainty in the Predictor}
\label{subsection:Suppress_in_predictor}
Several methods have been proposed to improve the human-interpretability of models by adding regularization constraints on the predictors during training. Regularization terms include elicited user preference for specific feature attributions \cite{Ross2017RightExplanations}, attribution priors such as smoothness in image saliency maps \cite{erion2019learning}, the number of rules and rule length in a rule list \cite{Letham2015}, and depth of a decision tree explanation \cite{Wu2018}. Here, we regularize the predictor to suppress attribution uncertainty. Regularizing the predictor by the attribution uncertainty directly suppresses the predictor model's dependency on uncertain input features. The explanation generated from this regularized predictor should have a smaller discrepancy and higher faithfulness, and yet be uncertainty-aware. However, this requires access to the training process of the predictor; this may not be possible if the prediction and explanation are not implemented by the same developer or stakeholder.\par
Similarly to how we regularized the explainer with an uncertainty penalty, we define the loss function to train the Regularized Predictor $\tilde{f}(\cdot)$ with a penalty based on input uncertainty as
\begin{equation}
\label{eq:reg_pred}
    \sum_{<x,y,\sigma>\in\mathscr{D}}\mathscr{L}\left(\tilde{f}_\theta (x),y \right)+ \overbrace{{\lambda\|\sigma \circ \omega(x,\tilde{f}_\theta)\|}_2^2}^\text{\shortstack{Attribution Uncertainty\\penalty}},
\end{equation}
where $(x,y)$ is the input features and output label of a training instance from training set $\mathscr{D}$, and $\sigma$ is the input uncertainty of $x$, $\mathscr{L}(\cdot)$ is the loss between predictor and ground-truth label, $\omega(x,\tilde{f}_\theta)$ is the attribution explanation function which takes in the predictor model $\tilde{f}_\theta$ and query instance $x$, and outputs the attribution ${\omega}^{(d)}$ for each feature $x^{(d)}$. Note that while Eq \ref{eq:reg_expl} is expressed in terms of linear weights from a LIME linear model explainer, we generalize the attribution explanation technique in the regularization term of Eq \ref{eq:reg_pred}. However, note that the choice of explanation technique is limited to the type of model and its optimization methods. \par
Further note that, for optimal training, $\omega(x,\tilde{f}_\theta)$ should be a function of the model parameters $\theta$. This allows the use of many search optimization methods, such as gradient descent which calculates the derivative of the loss with respect to $\theta$. In this case, the LIME explanation is not suitable, because the linear weight in LIME is model-agnostic and not expressible in $\theta$. Moreover, it is expensive to use LIME to compute attribution explanations, since it needs to sample and train a linear explainer model for every training instance when the loss is computed during the training of the predictor model $\tilde{f}_\theta$. To avoid these issues, we used a fully-connected neural network model and explain it with Integrated Gradients \cite{Sundararajan2017AxiomaticNetworks}, a common technique used to explain neural networks. Integrated Gradients integrates the model output gradient over each input dimension to get the attribution scores of input dimensions. It is fast to calculate during model training and suitable for gradient descent optimizers because it is differentiable with respect to the predictor model parameters $\theta$. We leave the generalization to other types of models or attribution explanation techniques for future work. We can use the same sampling method to estimate the attribution uncertainty generated by Integrated Gradients and show it by violin plots.\par

Having defined techniques to manage uncertainty, we next characterize and evaluate their performance in two experiments --- a simulation study to evaluate the Regularized Uncertainty explanation methods and a user study with human subjects to compare the impact of Showing and Suppressing uncertainty on trust and decision making.

\section{Characterization Study of Regularized Explanation}
\label{section:Simulation_Study}
We seek to understand how Regularized Uncertainty explanations can suppress uncertainty in attribution explanations and whether they are more faithful under input uncertainty than baseline explanations. Thus, we conducted a simulation study with a real-world dataset, where we simulated a large number of hypothetical instances to represent different input uncertainty from query instances.

\subsection{Dataset and Modeling Task}
\label{subsection:dataset_task}
To align the simulation study with the subsequent user study, we used a real-world dataset, the UCI wine quality dataset \cite{Cortez2009ModelingProperties}, for our simulation study. This dataset characterizes 1599 red wines by their quality score (label) based on 11 chemical properties (input features). We split the dataset into a training set (80\%) and a test set (20\%). We focused on a subset of 5 features (alcohol, pH, total SO2, sulphates and volatile acidity) because we want to simplify the task for users in the later user study, and the simulation study should have consistent settings. Instead of extensive testing more datasets whose input uncertainty is unknown or uncontrollable, we chose to further evaluate user perception with a human-subjects study later.\par
Instead of testing with only one uncertain feature, we selected two features --- alcohol and volatile acidity --- to be made uncertain to align with requirements for the later user study; if only one feature is uncertain, it would be too simple and predictable for participants, whereas too many uncertain features would lead to high cognitive load and user confusion. We perturbed the features for two levels of input uncertainty --- high (at 1 standard deviation), medium (at 0.5 standard deviations) and low (at 0.3 standard deviations). For each instance in the dataset and each uncertainty level, we synthesized 50 hypothetical data points by perturbing the two features.\par
Regarding the modeling task, although most interpretable machine learning models are studied with classification problems, we evaluate with regression models, so that we can precisely measure the similarity and differences in the explainer’s, predictor’s and user's inferences. To generalize regression to binary classification, a logistic function can be applied.

\subsection{Trained Models}
\label{subsection:trained_models}
Depending on the explanation regularization approach, we trained various predictor and explainer models: a 3-layer neural network model as the baseline predictor ($f=$NN), baseline and regularized LIME explainers on the baseline predictor ($g_f=$LIME(NN), $\tilde{g}_f=$\textit{Reg}LIME(NN)); the regularized predictor ($\tilde{f}=$\textit{Reg}NN) and Integrated Gradients explanation on the baseline and regularized predictors ($g_f=$IG(NN), $g_{\tilde{f}}=$IG(\textit{Reg}NN)). Note that the predictor model of IG(\textit{Reg}NN) is both regularized and explained by Integrated Gradients. Since two different explanation techniques, LIME and IG, were used to regularize the explainer \textit{Reg}LIME(NN) and predictor IG(\textit{Reg}NN), in order to control the study, the two regularized methods are compared with their baselines LIME(NN) and IG(NN), respectively. Next, we evaluated the performance of the different explanations by defining an Expected Faithfulness metric.

\subsection{New Measure: Expected Faithfulness}
\label{sec:exp_faith}
To investigate explainer faithfulness under uncertain inputs, we have to measure how the explainer can make predictions that agree with the predictor even if the instance has uncertainty or noisy measurements. We defined a new measure --- \textit{Expected} Faithfulness Distance\footnote{Note that higher distance actually refers to less similarity, and lower faithfulness, so this should technically be called "Unfaithfulness Distance", but we term it as "Faithfulness Distance" to simplify nomenclature.}\footnote{We call this metric a ``distance" instead of ``difference", since we calculate it as the aggregate mean squared error (MSE) of differences for all instances, to represent global faithfulness for the explainer model.} --- to measure this effect and show how expected faithfulness distance can be reduced with Regularized Uncertainty explanation to improve expected faithfulness.\par
The inference of an explainer should be consistent and similar to the inference of the predictor; i.e., explainers should be faithful to predictors \cite{Kulesza2015PrinciplesLearning,Lundberg2017APredictions,Ribeiro2016WhyClassifier}.For example, the LIME explainer is a linear model trained to infer the same output as the underlying predictor. The faithfulness distance of baseline explainer for an instance $x$ is defined as the difference between the predictor’s inference $f(x)$ and the baseline explainer’s inference $g_{f}(x)$, i.e., $F_0=\left(f(x)-g_f(x)\right)^2, F_0 \geq 0$. This is a point-estimate metric. When the input feature of instance $x$ is uncertain, we specify that a locally robust explainer should be less reliant on input noise $\epsilon$, thus the explainer’s inference of the noisy input $x+\epsilon$ should be the same as the predictor's inference on the original instance $x$. To measure this local robustness, for the baseline explainer $g_f$, we define the faithfulness distance under uncertainty as $F_{g_f}=\left(f(x)-g_f (x+\epsilon)\right)^2$. \par
For a baseline explanation technique i.e. LIME(NN) or IG(NN), we expect that on average the baseline is less faithful at explaining uncertain instances because its explanation is sensitive to noise. We define its average faithfulness distance over uncertain inputs as the \textit{expected faithfulness distance}: 
\begin{equation}
    E[F_{g_f}]=E_\epsilon\left[\left(f(x)-g_f(x+\epsilon)\right)^2 \right] = \int_{-\infty}^{\infty} \left(f(x)-g_f(x+\epsilon)\right)^2 P(\epsilon) \  d\epsilon ,
\end{equation}
and assert that $E[F_{g_f}]\geq F_0$. In the simulations, we generated 150 hypothetical instances around each test instance by sampling $\epsilon$ from a Gaussian distribution at high (standard deviation = 1), medium (standard deviation = 0.5) and low (standard deviation = 0.3) uncertainties. See the proof in \ref{append:proof}. \par
On the other hand, for the Regularized Uncertainty explainer $\tilde{g}_{f(\cdot)}=$, \textit{Reg}LIME(NN), we expect that its faithfulness to explain uncertain instances is better than the baseline explainer's point-estimated faithfulness distance $F_0$, due to the Regularized Explainer’s robustness. First, we specify that the robust inference outcome should aim to be similar to the prediction for the original instance, i.e., $\tilde{g}_f(x+\epsilon) \approx f(x)$. The faithfulness distance for the Regularized Explainer on an uncertain input instance is thus $F_{\tilde{g}_f}=\left(f(x)-\tilde{g}_f(x+\epsilon)\right)^2$. Although the explanation at some uncertain instances may be less faithful than at the original query instance $x$, on average across all hypothetical uncertain instances, the faithfulness distance of $\tilde{g}_f(\cdot)$ may sometimes be as good as or better than that of the point-estimated explanation on original query instance $x$, i.e., $Prob(E[F_{\tilde{g}_f}]\textless  F_0) \textgreater  0$. For the Regularized Explainer, the expected faithfulness distance is:
\begin{equation}
     E[F_{\tilde{g}_f}]=E_\epsilon\left[\left(f(x)-\tilde{g}_f(x+\epsilon)\right)^2\right] = \int_{-\infty}^{\infty} \left(f(x)-\tilde{g}_f(x+\epsilon)\right)^2 P(\epsilon) \  d\epsilon
\end{equation}	

For the Regularized Predictor $\tilde{f}(\cdot)$ we defined, its prediction on a query instance $x$ is explained by its model-dependent explanation technique $g_{\tilde{f}}(\cdot)$, and $g_{\tilde{f}}(x)$ is the explanation inference of instance $x$ inferred by the explanation. Here the outcome is calculated by summing up all attributions and model bias. The point-estimate explanation faithfulness distance $F_0$ is defined as before, i.e., $F_0=\left(f(x)-g_f (x)\right)^2$. Note that here the baseline explanation is IG(NN). Thus, we define the expected faithfulness distance over uncertain input for the Regularized Predictor IG(\textit{Reg}NN) as:
\begin{equation}
     E[F_{g_{\tilde{f}}}]=E_\epsilon\left[\left(f(x)-g_{\tilde{f}}(x+\epsilon)\right)^2\right] = \int_{-\infty}^{\infty} \left(f(x)-g_{\tilde{f}}(x+\epsilon)\right)^2 P(\epsilon) \  d\epsilon
\end{equation}
Similar to the Regularized Explainer, we hypothesize that $E[F_{g_f}]\geq F_0$ for the baseline predictor $g_f=$IG(NN), and $Prob(E[F_{g_{\tilde{f}}}]\textless  F_0)\textgreater 0$ for Regularized Predictor $g_{\tilde{f}}=$IG(\textit{Reg}NN).

\subsection{Results: Regularized Explainer and Regularized Predictor Improve Expected Faithfulness Distance}
\label{subsection:sim_result_faith_dist}
Next, we conduct a simulation experiment to investigate how likely the explanations of instances are to benefit from improved expected faithfulness distance by suppressing uncertainty with the Regularized Explainer and Regularized Predictor.\par

%\label{sec:simres_regexp}
\begin{figure}[ht]
    \centering
    \includegraphics[width=7cm]{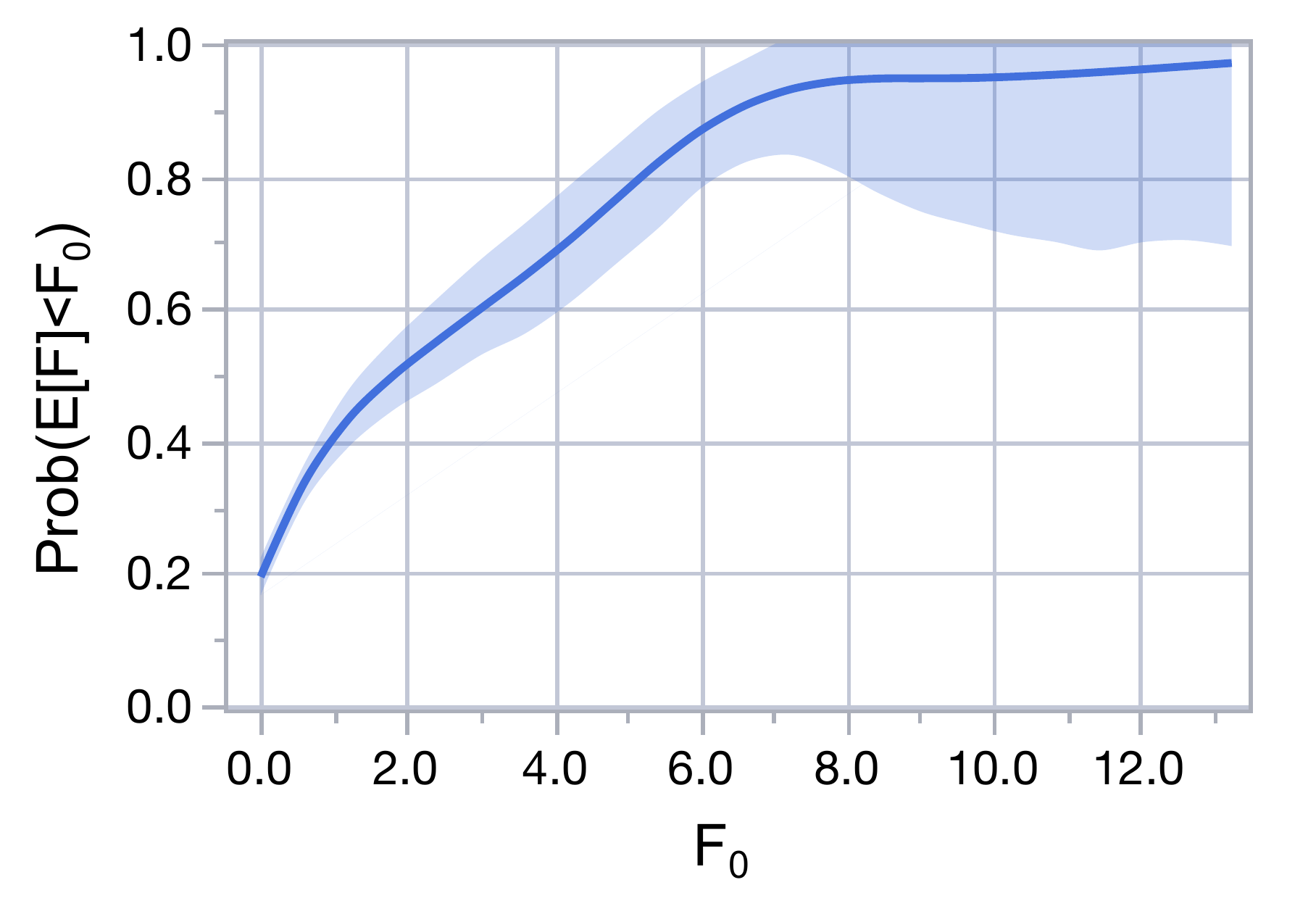}
    \caption{Results of simulation on wine dataset. X-axis $F_0$ is the point estimate faithfulness distance from Baseline explainer LIME(NN), which is the distance between explainer prediction and model prediction. Y-axis is the probability of Regularized Explainer \textit{Reg}LIME(NN) has a smaller expected faithfulness distance on all hypothetical instances than the point estimate faithfulness distance of Baseline ($F_0$). The shaded area is the standard error. This result is calculated by the high input uncertainty level (Gaussian distribution with 0 mean and 1 standard deviation).}
    \label{fig:LIME_E[F]}
\end{figure}

%This experiment examined expected faithfulness distance to test when many uncertain instances were explained by the Regularized Explainer, whether on average they will infer similarly and be faithful to the predictor inferred outcome.  As discussed at the beginning of Section \ref{sec:simulation study}, we trained a fully connected 3-layered neural network as baseline predictor(NN), explained it with baseline LIME baseline explanation $(LIME(NN))$ to calculate baseline faithfulness distance $F_0$, explained NN with regularized LIME explainer as regularized explanation $(RegLIME(NN))$ and calculated expected faithfulness distance $E[F_{\tilde{g}_f}]$ on RegLIME(NN). 

Figure \ref{fig:LIME_E[F]} shows the probability of $E[F_{\tilde{g}_f}]\textless  F_0$ being true, which indicates how often the Regularized Explainer \textit{Reg}LIME(NN) is more faithful than the Baseline explainer LIME(NN) under input uncertainty. The y-axis value, calculated by the number of instances, satisfies $E[F_{\tilde{g}_f}]\textless  F_0$ divided by the number of instances in the dataset. And x-axis represents the different values of baseline explainer faithfulness distance $F_0$ for different $x_0$. We can see that for less faithful baseline explainers (larger $F_0$), regularizing the explainer is more likely to have smaller expected faithfulness distance under uncertainty (i.e., $Prob(E[F_{\tilde{g}_f}]\textless  F_0)$ gets higher). This could be because when $F_0$ is small, the baseline explanation is already very faithful, so there is not much room for regularization to improve. \par%%The expected faithfulness may also be affected by the magnitude of input uncertainty and the choice of regularization weight $\lambda$, but we defer these for future work.
%\subsubsection{Results: Explanation of Regularized Predictor Improved Expected Faithfulness Distance}
\begin{figure}[ht!]
    \leftskip8em
    %\centering
    \includegraphics[width=9cm]{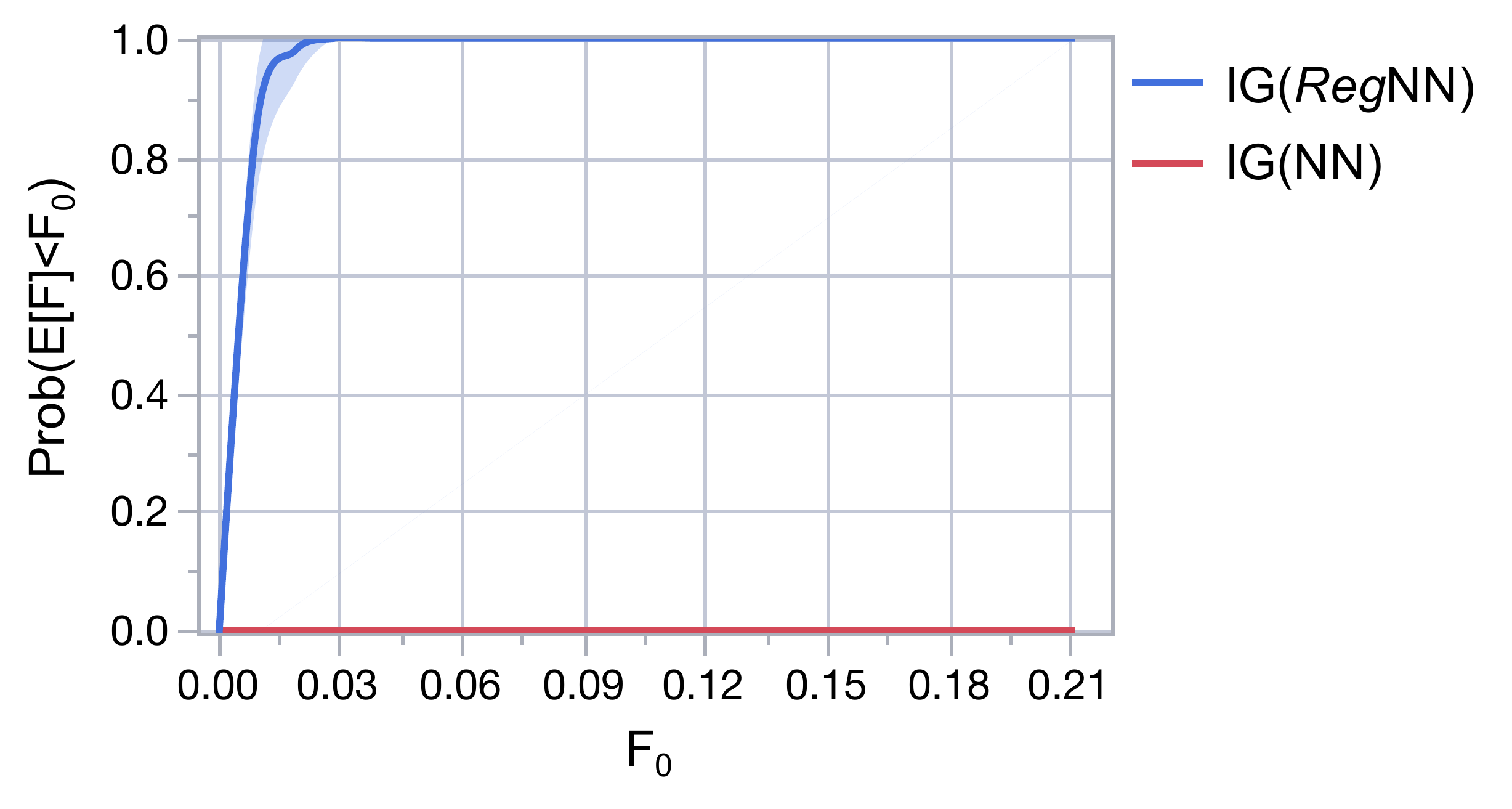}
    \caption{Results of simulation on Regularized Predictor. X-axis $F_0$ is the point estimate faithfulness from Baseline predictor IG(NN). Y-axis is the probability that an explanation has a smaller expected faithfulness over all hypothetical instances than the point estimate faithfulness of explanation from the Baseline predictor. Blue is for the Regularized Predictor IG(\textit{Reg}NN), $Prob( E[F_{g_{\tilde{f}} } ] \textless   F_0 )$. Red is for IG explanation on baseline neural networks predictor IG(NN), $Prob(E[F_{g_f}\textless  F_0])$. Error bar is standard error. This result is calculated by the high input uncertainty level (Gaussian distribution with 0 mean and 1 standard deviation).}
    \label{fig:IGmean_E[F]}
\end{figure}
%As mentioned at the beginning of Section \ref{sec:simulation study}, here we used the same baseline NN predictor from previous Regularized Explainer study, but we explained it with Integrated Gradients as the baseline explanation (IG(NN)) for $F0$ calculation, and then we trained regularized predictor and explained with Integrated Gradients (IG(RegNN)). Then we calculated expected faithfulness distance over noise for both IG(NN) and IG(RegNN). 
Similar to Figure \ref{fig:LIME_E[F]}, Figure \ref{fig:IGmean_E[F]} shows that regularizing on the predictor helps improve the Expected Faithfulness (higher $Prob(E[F]\textless  F_0)$), especially for predictors with poorer baseline explanations (higher $F_0$).
% the results that for an instance $x_0$, when its explanation from a baseline predictor gets less faithful to the model ($F_0$ increases), the probability that the expected faithfulness of explanations from Regularized predictor is better than Baseline over all noisy instances in $x_0$’s neighborhood. Similar to Regularized explainer, the result infers that when the baseline is less faithful, regularizing predictor is more likely to increase the explanation faithfulness because the predictor is robust to input noise, and the expected faithfulness over uncertainty of Baseline can never be better than point-estimation faithfulness due to sensitivity to input uncertainty.
Note that the range of $F_0$ for explainers of the Regularized Predictor IG(\textit{Reg}NN) is much smaller than that for Regularized Explainer \textit{Reg}LIME(NN), which means that the baseline Integrated Gradient is a more faithful explanation method than baseline LIME. Although the Regularized Predictor has better results on expected faithfulness than Regularized Explainer, regularizing the predictor at training is more computationally expensive than regularizing the explainer at inference time, and in many scenarios, developers may not have access to change the predictor. In contrast, regularizing the model-agnostic explainer enables to suppress attribution uncertainty based on users' preference for different uncertainty handling strategies.\par

\subsection{Results: Regularized Explainer and Regularized Predictor Have Similar Explanations}
\label{subsection:sim_result_similar}
\begin{figure}[ht]
    \centering
    \includegraphics[width=10cm]{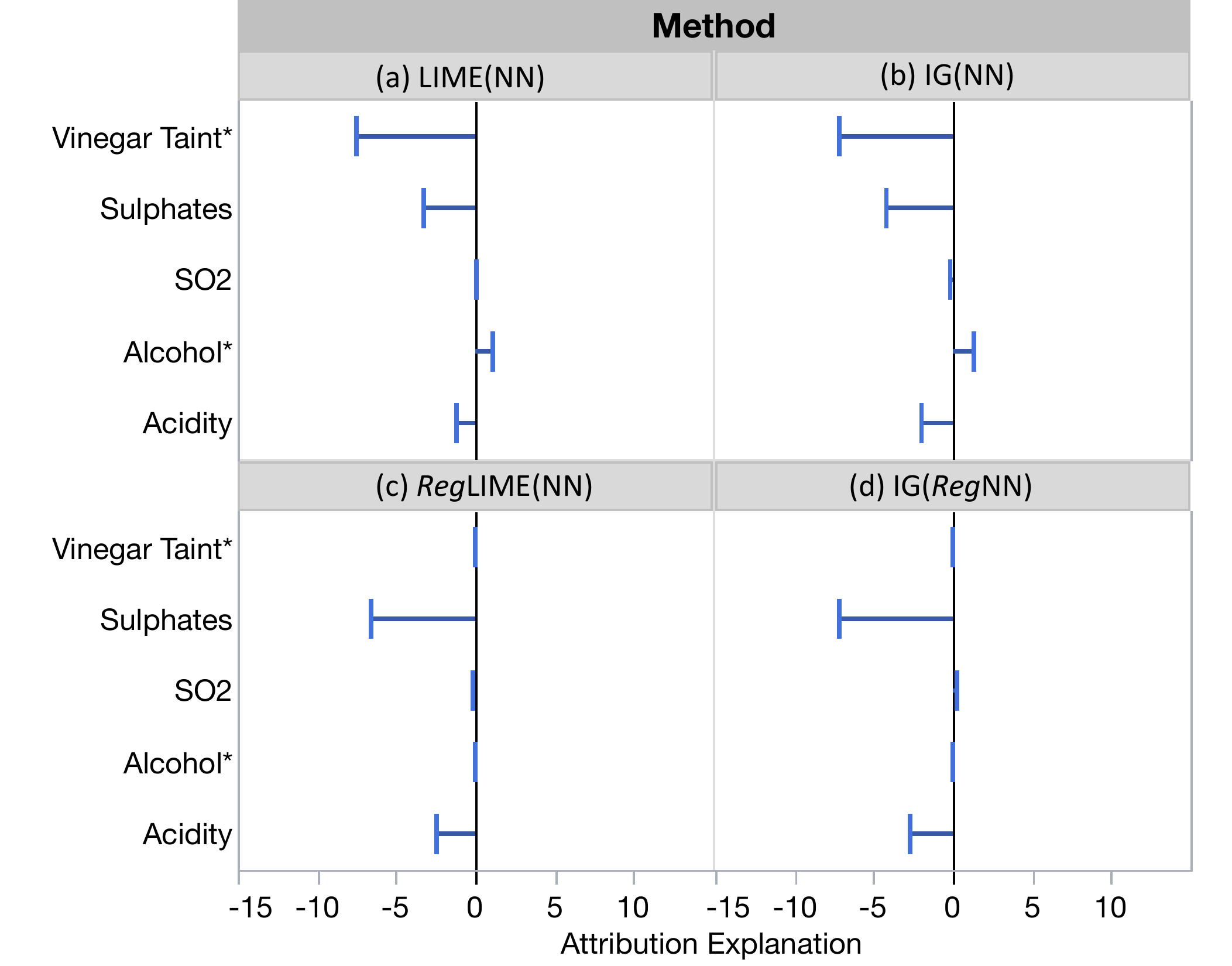}
    \caption{Example explanations from various techniques: (a) Baseline LIME explainer on baseline predictor LIME(NN); (b) Baseline Integrated Gradients explanation on baseline predictor IG(NN); (c) Regularized LIME explainer on baseline predictor \textit{Reg}LIME(NN); (d) Integrated Gradients Explanation on Regularized Predictor IG(\textit{Reg}NN). The two baseline explanations are slightly different, but the two regularized explanations are similar after suppression. * indicates that the Vinegar Taint and Alcohol features had input uncertainty.}
    \label{fig:similar_exp_example}
\end{figure}
\begin{figure}[ht]
    \centering
    \includegraphics[width=14cm]{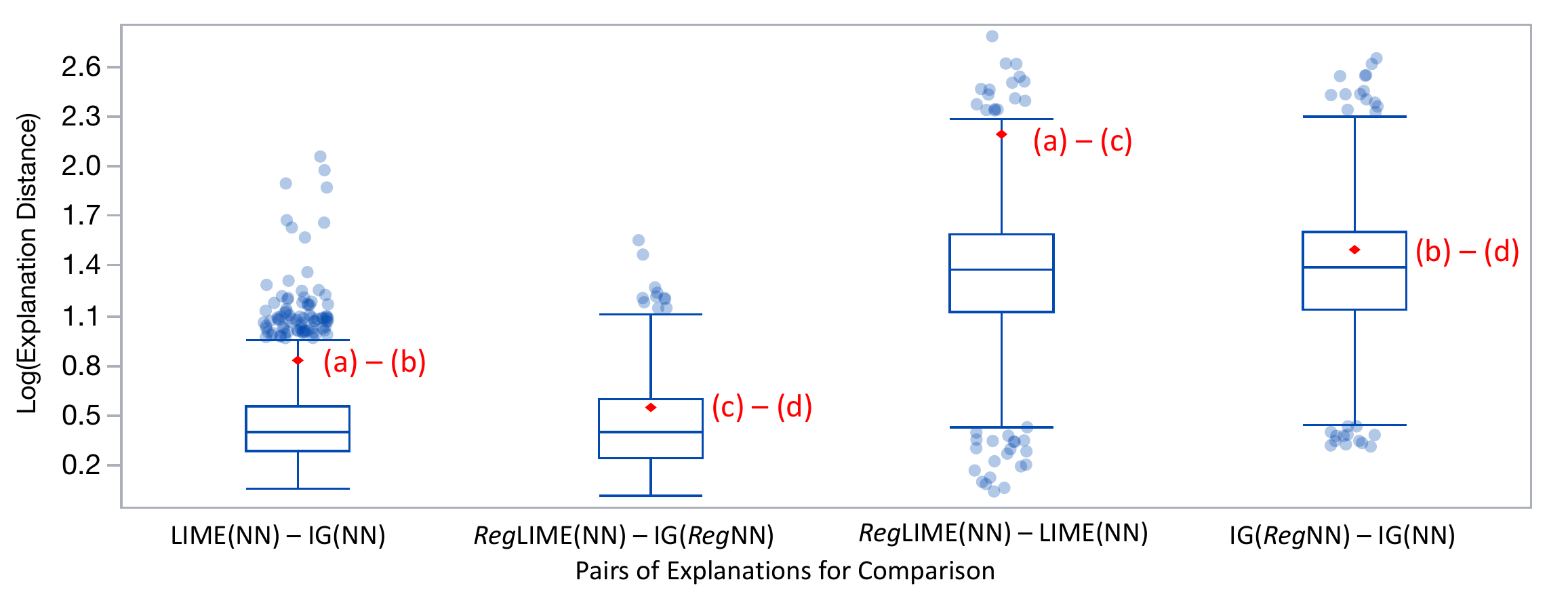}
    \caption{Distribution of logarithm of explanation distance between different techniques, shown as box plots (1st quartile, median, 3rd quartile) and whiskers (1st quartile - 1.5$\times$interquartile range or min, 3rd quartile + 1.5$\times$interquartile range or max). The two baseline explanations are as similar to each other as the regularized explanations to each other (low distance in two left plots). The suppressing effects on attributions by regularizing the explainer and regularizing the predictor are similar too (similar distances in two right plots). Red diamond markers indicate explanation distances for pairs of explanations shown in Figure \ref{fig:similar_exp_example}. For example, (a)-(b) represents the logarithm of Euclidean distance between the explanations in Figure \ref{fig:similar_exp_example}(a) and Figure \ref{fig:similar_exp_example}(b). We used logarithmic transform to provide a compact visualization of the heavily skewed distribution of results.}
    \label{fig:exp_dist_boxplot}
\end{figure}
We have shown that both Regularized Explainer \textit{Reg}LIME(NN) and Regularized Predictor IG(\textit{Reg}NN) methods can generate explanations that are robust to uncertain input and more faithful than their baselines LIME(NN) and IG(NN), but how similar or different are the explanations from these methods? \par

Figure \ref{fig:similar_exp_example} shows four variants of explanations for an instance example. While predictor is explained by different attribution explanation techniques i.e., LIME and IG regularizing either on the explainer or predictor produce a similar suppressed explanation result.The Regularized Explainer and Regularized Predictor methods can suppress attribution to uncertain features in a similar way resulting in similar attribution explanations. To quantify this similarity across all instances, we calculate the similarity between explanations based on the Euclidean distance of their feature attributions.

Figure \ref{fig:exp_dist_boxplot} shows how explanations from the Regularized Explainer \textit{Reg}LIME(NN) and Regularized Predictor IG(\textit{Reg}NN) are also similar. Their similarity is as close the similarity between explanations from the baseline Explainer LIME(NN) and baseline Predictor IG(NN) (compare Figure \ref{fig:exp_dist_boxplot} left two bars), and closer than the similarity between baseline and regularized explanations by more than 5 times\footnote{3.28/0.599=5.48, 3.24/0.599=5.41, 3.28/0.648=5.06, and 3.24/0.648=5, respectively.} (contrast Figure \ref{fig:exp_dist_boxplot} first left bar with right two bars). The differences between regularized and baseline explanation distances were significant in Wilcoxon Rank Sum tests (p<.0001) with large effect sizes ($r=.569$ to $.579$) \cite{pallant2020spss}.

A take-away from this analysis is that uncertainty-aware explanations from the model-agnostic Regularized Explanation can be as faithful as those from Regularized Prediction. The former method also has the benefit of being easier to train.

\section{User Studies to Show and Suppress Uncertainty in Attribution Explanations}
\label{section:User_Study}
We conducted a formative qualitative study and follow-up summative quantitative study to evaluate the two proposed explanation techniques to handle attribution uncertainty. With the  qualitative study, we sought to understand whether and how users could benefit from using either Show or Suppress, how different approaches affect users' trust in the AI and decision making, and how users interpret and use the explanations. With the controlled quantitative user study, we compared the Show and Suppress explanation techniques against baselines to evaluate at scale how using them influences user trust, perceived helpfulness and decision quality on a rating and decision task.\par
For consistency, we employed the same experiment task, experiment apparatus and experiment variables for the qualitative and quantitative studies. That is, the experiment task was used as a probe in the qualitative study. In this section, we first describe the user task, independent variables, control variables, system and explanation user interface (UI) apparatus. We will describe the method, procedure, and results for the qualitative and quantitative studies in the subsequent sections.\par

\subsection{User Study Task: Wine Quality Control Inspection}
\label{subsection:task}
We designed a user decision task that involves estimating a score to make a binary yes/no decision. We defined a scenario to fit the context of the UCI Wine Quality dataset that correlates chemical properties with wine score ratings labeled by human experts. We presented participants with the scenario that they work as a quality control inspector to estimate the quality score (0 to 100) of wine at a production winery and decide whether to accept or reject a wine for sale. If the participant thought the wine score is $\textgreater $50, then he should accept it, otherwise, he should reject (i.e., reject $\leq$ 50). Participants were introduced to the Smart Wine Rating System that estimates (predicts) the wine quality score from 0 to 100 based on five chemical properties, and told that they could use the system to help with decision making. Participants are taught about how to read the system user interface (described in Section \ref{subsection:exp_apparatus}) and also informed about the uncertainty in chemical readings. Depending on the experiment condition, the System provides basic information (readings with uncertainty and predicted quality score with uncertainty), or various forms of additional information. We did not explicitly call these “explanations" to the participants. Participants were told that the wines they were inspecting are borderline cases\footnote{Cases were chosen to emulate the paradigm where the machine learning system would defer to human judgement if it is not confident of its prediction. Hence, with uncertainty, most cases (29/30) had confidence intervals straddling the mean value, i.e., lower bound below mean and upper bound above mean. We expect that participants would make quick decisions that would likely agree with the prediction for non-straddling cases and perceive them to be easy.} and although the system predicts a score, it may be wrong. Wine instances were chosen such that 50\% of them should be rejected. For each decision task, participants indicate their inspection decision, estimated wine score value and uncertainty of their score as a number.\par
More details of the scenario as presented to participants are described in \ref{fig:tutorial_background}. In the qualitative study, we emphasized on understanding the participants' thought processes and preferences across many explanation techniques, so they were asked to take their time to study the UI and think aloud while doing so, and participants were exposed to three system variants. For the quantitative study, we emphasized quick and intuitive decision making, and applied a timed incentive to make at least 12 correct decisions for 15 wines within 15 minutes for each system variant. Participants used two variants of the system one after the other, and were told to aim for at least 80\% correct decisions (i.e., $\geq $12/15) in their inspections each time. \par
We chose this task for several reasons. First, we wanted to minimize the influence of prior knowledge and personal preferences in familiar tasks. With this task, lay people are not familiar with rating wines based on their chemical properties, and we can control the knowledge by training. Second, it is plausible that inputs can be uncertain due to errors when measuring chemical properties. Third, predicting wine quality rating is a classic regression problem in machine learning, and this allows us to more precisely measure the impact on user labeling at a higher resolution than only on a classification task. The dataset and model settings are consistent with the Simulation experiment. \par
\subsection{Experiment Treatment: Independent, Control, and Dependent Variables}
\label{subsection:exp_treatment}
\begin{figure}[ht]
    \centering
    \includegraphics[width=12cm]{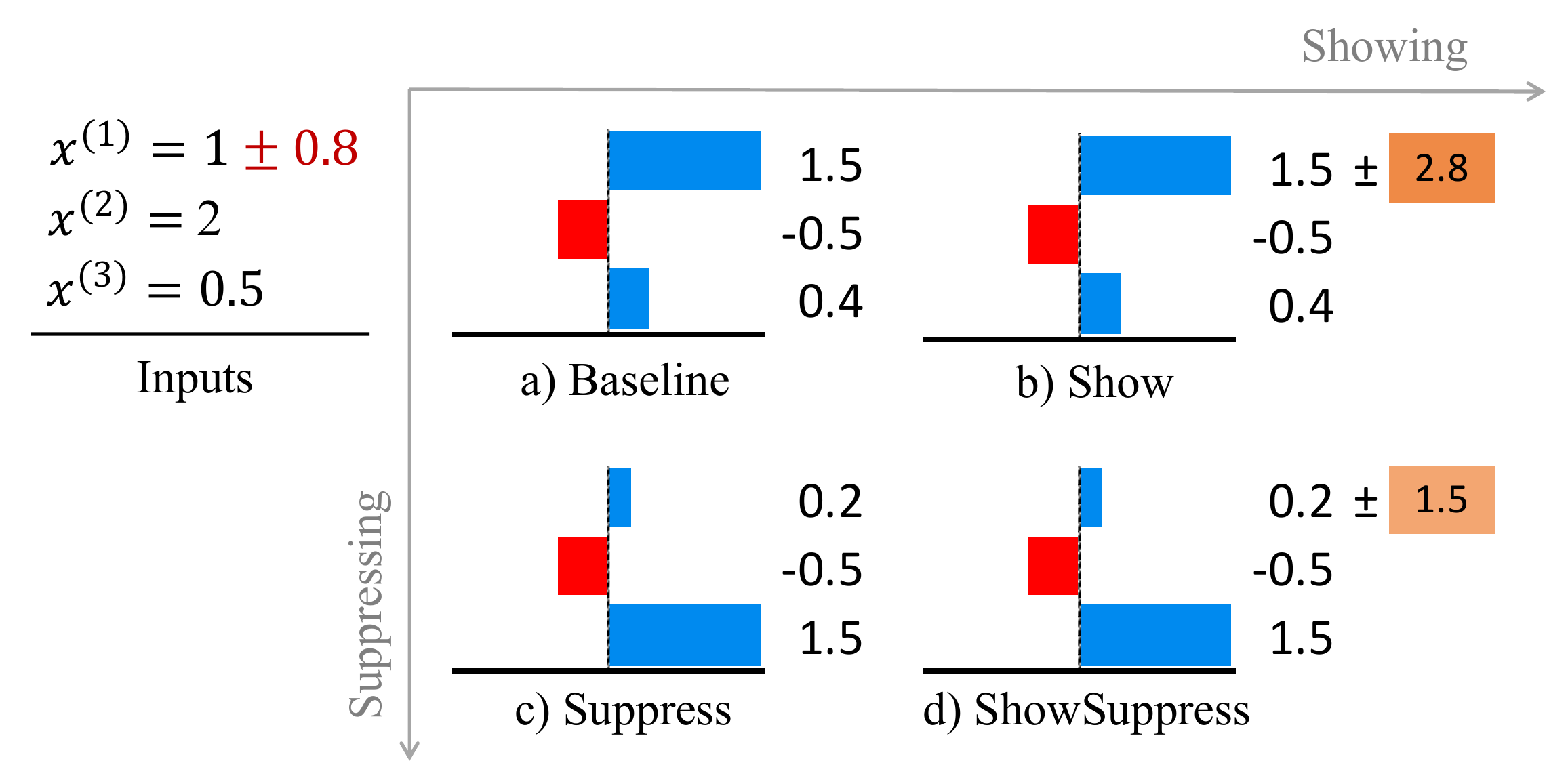}
    \caption{Conceptual examples of simplified visualization in user study.}
    \label{fig:simplified_sample}
\end{figure}

The experiments focus on one primary independent variable about which explanation technique is used for decision making: 

\begin{enumerate}[leftmargin=25pt]
    \item Technique (5 levels): score prediction system with no explanation (None\footnote{We included the None baseline as a reference to verify whether any effects are due to explanations in general or about the specific explanation technique, but we focus our analysis on the four explanation techniques.}), attribution explanation with subscore values in a tornado plot (Baseline, see Figure \ref{fig:simplified_sample}a)), tornado plot including subscore uncertainties (Show, see Figure \ref{fig:simplified_sample}b), tornado plot of suppressed subscores (Suppress, see Figure \ref{fig:simplified_sample}c), tornado plot of Suppressed subscores and shown suppressed uncertainties  (ShowSuppress\footnote{We only used the Regularized Explainer to control the behavior of the predictor model and keep it consistent across explanation conditions; regularizing the predictor will lead to different predictions for the same instances and this is a confounder that affects the user’s trust \cite{Yin2019UnderstandingModels}. Nevertheless, our simulation results have shown that the Regularized Explainer has similar explanations as the Regularized Predictor.}, see Figure \ref{fig:simplified_sample}d).
\end{enumerate}

We treat Technique as a within-subjects design, thus we have 5 conditions in the experiment. Each participant uses more than one technique with multiple trials per technique. To limit experiment variability and present challenging cases for user intervention to simulate when an AI system defers decisions to humans due to its lack of prediction confidence, we set the following variables constant as described.
\begin{enumerate}[resume,leftmargin=25pt]
    \item Input Features with Uncertainty (Alcohol and Vinegar Taint): we set only two input features to have uncertainty, and with equal relative amounts. All other input features have no uncertainty.
    \item Input Uncertainty (medium): we fixed the uncertainty of each input feature as a Gaussian distribution with a standard deviation that is half the standard deviation of the feature values.
    \item System Correctness (all instances' system prediction is correct): For each selected instance, we made sure the predicted and actual (ground truth) scores to be consistent, i.e. on the same side of the decision threshold (i.e., both above or both below). 
    \item Closeness to Decision Threshold (between 40 and 60, i.e., $|$system score - $50|\textless 10$): For all selected instances, we made sure the system prediction score is close to the decision threshold 50 (within 40 to 60) to present challenging ambiguous cases.
\end{enumerate}

% Each participant was randomly assigned to two of the five Technique conditions (within subjects). We randomly selected a subset of 30 data instances to have participants view the same instances in a controlled study. The selected instance satisfied two criteria:1)system In each condition, 15 wine instances were presented to each participant in a random order, which was recorded by Trial Sequence (within subjects). In each trial, participants made their estimation twice before and after explanation plots were available (within subjects).

To choose a representative sample of hypothetical instances with uncertainty, we processed and selected data based on the following steps:
\begin{enumerate}[label=\roman*.,leftmargin=25pt]
    \item For each instance from the dataset in the simulation study, add noise according to a Gaussian distribution to create 50 hypothetical instances with uncertainty.
    \item Filter instances with the aforementioned control criteria.
    \item Divide instances into separate clusters by performing hierarchical clustering on their feature values, attribution explanations, model predictions, the uncertainty of these values and ground truth scores.
    \item Perform stratified sampling across clusters to select final instances for the experiment.
\end{enumerate}
Ultimately, we selected 34 wine instances for our user studies, 4 for practice, and 30 for the main trials. The mean Closeness to Decision Threshold is 1.95 (SD=1.05).\par

We measured user performance \textit{per trial} and overall system impression of the explanation technique \textit{per condition}. For each trial, we measured
\begin{enumerate}[leftmargin=25pt]
    \item Decision Time: the amount of time in seconds for the participant to decide to accept or reject the wine, estimate the wine quality score and score uncertainty. This includes time to see and study the input values and uncertainty, system prediction and explanation (if available). Decision Time is only recorded and analyzed for the quantitative study.
    \item User quality score value: the wine quality score estimated by the participant, denoted as $Score_{User}$.
    \item User quality score uncertainty: the $\pm$ error bound of the participant's wine quality score value, denoted as $ScoreUncertainty_{User}$. This should refer to half the 90\% confidence interval width to correspond to what the system displayed.
\end{enumerate}
For each condition, we measured
\begin{enumerate}[resume, leftmargin=25pt]
    \item Confidence: the participant's self-reported confidence in decision making on a 7-point Likert scale from strongly disagree to strongly agree.
    \item Trust: self-reported trust of the system for an accurate score prediction, on a 7-point Likert scale from strongly disagree to strongly agree. Note that the participant may trust the system, and yet not feel confident about her decision due to other factors.
    \item Helpfulness: self-reported overall helpfulness of the system and any available information or explanations for decision making, on a 7-point Likert scale from strongly disagree to strongly agree.
    \item Helpfulness of specific system features: self-reported helpfulness of different features of the system on 7-point Likert scales from strongly disagree to strongly agree. System features include reading value, reading uncertainty, subscore value, subscore uncertainty, suppressed subscore value, suppressed subscore uncertainty,score value, and score uncertainty. These helpfulness questions were asked based on Technique condition.
\end{enumerate}	
These measures were recorded for the qualitative interview and quantitative study and used to compute dependent variables (See Section \ref{subsubsection:DVs}). \par

\subsection{Experiment Apparatus}
\label{subsection:exp_apparatus}

\begin{figure}[!t]
    \centering
    \includegraphics[width=13cm]{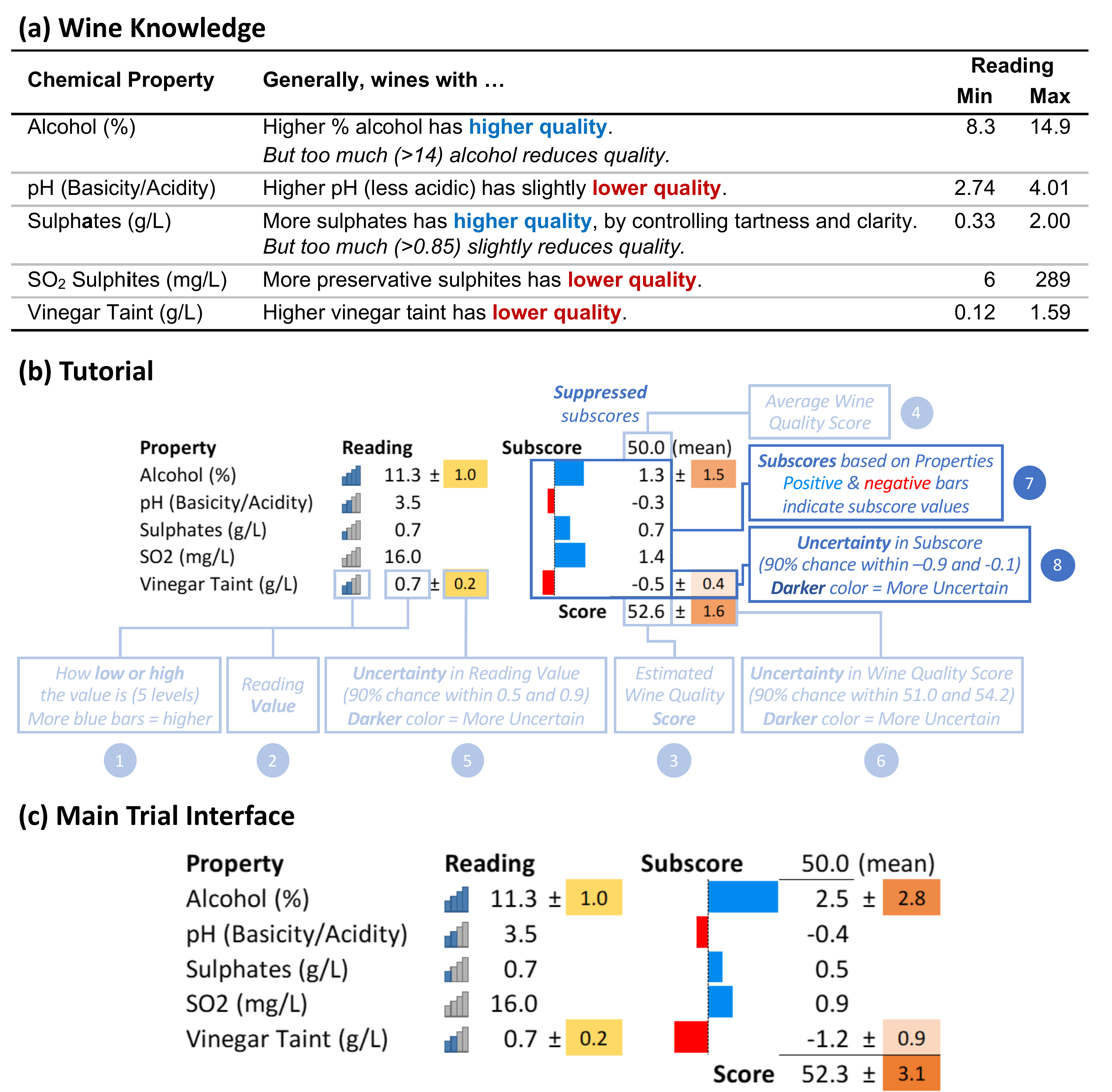}
    \caption{(a) Background Knowledge table shown to participants to teach them how each chemical property affects the wine quality score. The description is obtained by analyzing partial dependence between input features and the rating score, where a trend is flat if the slope of between wine quality score and the standardized property is less than 2.0/unit. (b) Annotated screenshot of the system user interface (ShowSuppress version with all 8 components displayed) used to teach participants on how to read the UI properly. Other system variants will have fewer UI components. (c) System UI as seen during practice or main trial without annotation clutter (Show version displayed). With the Suppress and ShowSuppress conditions, participants can hover the mouse over the figure to compare the explanation before and after suppression (subscore and score values and uncertainties will switch). Yellow and orange highlights indicate the amount of uncertainty in the reading and subscore, respectively.}
    \label{fig:walk_through}
\end{figure}

We designed a simplified user interface to communicate attribution explanations with uncertainty for the Show and Suppress techniques to be usable by lay users from Amazon Mechanical Turk. We prioritized usability and familiarity over more precise and expressive visualizations. In an online pilot study with the visualization in Figure \ref{fig:sample}, we found that many lay users had difficulty in learning to read the violin plots to pass comprehension tests. Hence, we chose a familiar table layout with input values as numbers, attribution values, and uncertainties represented as ($\pm$) numbers. Uncertainty numbers indicate the 90\% confidence interval half-width, suggesting the range within which the true value has a 90\% chance to lie. with small violin plots. To show the attribution explanation, we kept the tornado plot, but as a simple bar chart embedded in the table, rather than a large T-bar chart. Attribution uncertainty is simply presented as a single number, rather than a distribution curve or error bars. For simplicity to lay users, we call the attribution values as “subscores". \par

Figure \ref{fig:walk_through} shows the modular table user interface that can show different UI components depending on the explanation technique condition. 1) \& 2) the chemical property reading value, which is the system input, and the indicator of how high or low the reading value is; 3) the estimated wine quality score, which is the system output; 4) the average wine quality score for all the wine, which is also the decision boundary; 5) the uncertainty of the property reading values, which is the input uncertainty; 6) the uncertainty of wine quality score, which is the output uncertainty; 7) subscores for each property, which is the attribution explanation; and 8) the uncertainty in subscores, which is the attribution uncertainty. All participants only learned the system features 1) to 6) in the training, and would not learn about the system explanation until the main study.\par
The experiment was implemented as an online Qualtrics survey with different images for each system UI trial, and different trials on subsequent pages. We describe the experiment procedure next.

\subsection{Experiment Procedure}
\label{subsection:procedure}

We describe the procedure for participants in the quantitative study, then describe the truncated procedure for qualitative study participants. In the quantitative study, participants from Amazon Mechanical Turk follow the procedure:
\begin{enumerate}[leftmargin=25pt]
    \item Read a welcome message and consent to the study.
    \item Study a tutorial and answer corresponding screening questions on:
    \begin{enumerate}[label=\roman*.]
        \item The task description and task background knowledge (Figure \ref{fig:walk_through}a).
        \item System user interface versions with the incremental introduction to input readings and output score, readings and score uncertainty (Figure \ref{fig:walk_through}b).
    \end{enumerate}
    \item If participants had \textless 5 out of 7 screening questions correct, they will be disqualified from the study. Those that pass proceed to the next step and answer 6 questions about Uncertainty Tolerance and Uncertainty Decisive (7-point Likert Scale, see Figure \ref{fig:uncertainty_strategy}). 
    \item Task reminder to describe the user task and introduce bonus incentives (12 correct at \$0.25, with \$0.25 increments until 15 correct at \$1.00). 
    \item Randomly assigned to two (of 5) explanation Technique conditions, where for each condition, the participant sees:
    \begin{enumerate}[label=\roman*.]
        \item Pre-Condition tutorial on the specific explanation UI (Figure \ref{fig:walk_through}b, Figure \ref{fig:tutorial_showsuppress}), with comprehension test questions for data quality checking, not to screen participants.
        \item Two practice trials, where for each trial, the participant
        \begin{itemize} 
            \item Estimates his score value and uncertainty\footnote{We asked for a simple number, and did not ask participants to estimate a 90\% confidence interval to avoid technical jargon and they have already learned from the tutorial that the uncertainty refers to a 90\% chance of the value being within the range. We had piloted asking uncertainty with the bin and balls question \cite{Goldstein2014LayDistributions}, but wanted a much faster data entry method, and did not need very high accuracy.} using sliders\footnote{We recorded with sliders instead of number text entry to speed up data entry and allow easy entry of mixed decimal numbers (see Figure \ref{fig:main_trial}).}, and
            \item Decides to accept or reject the wine. 
            \item Sees a displayed timer to indicate how much time has passed.
            \item Sees a second page (See Figure \ref{fig:trial_feedback}) that reviews her answers, gets feedback on the ground truth wine score and whether her decision of accepting/reject is correct, and answers a question about her rationale\footnote{Her answers are repeated for reference. The rationale question serves two purposes: i) manipulation check to ensure that users are paying attention and reasonably understand the user interface, ii) stimulate users to self-explain to raise their level of understanding before proceeding to the main trials.} in the decision and rating estimation. This page was not included for the main trial to not interrupt rapid task completion.
        \end{itemize} 
        \item Task and incentive bonus reminder.
        \item 15 main trials (randomly ordered) with the same format as practice trials but without the second page of feedback.
        \item Post-Condition page where the participant reviews her performance (how many trials correct) and answers questions to rate her Confidence in decision making, Trust in the system estimation, and Helpfulness of the system information for decision making.
        \item The participant takes a one-minute break after finishing the first condition.
    \end{enumerate}
    \item After both conditions, the participant answers demographics questions (e.g.  age, gender, ethnicity, education, employment status and occupation) and receives feedback on how much bonus she would get. 
\end{enumerate}

The qualitative study is conducted online via a Zoom audio call with screen capture recording, which the participant consents to. They also go through the tutorial and can ask the experimenter clarification questions. They do not receive a bonus for their task performance, since we do not incentivize for speed and accuracy. Finally, the focus of inquiry was to observe their usage and rationale using the think aloud-protocol \cite{nielsen2002getting} and interview questions, rather than their answers to the rating questions.% After attempting the task and answering some questions, the participants were provided with the actual score answers to facilitate further discussion.
Mummolo et al. found little evidence for the experimenter demand effect for survey studies on Amazon Mechanical Turk \cite{mummolo2019demand}. Nevertheless, to control for the demand effect, we randomized the order of testing explanation techniques, regularly informed participants at every trial that the system could be wrong, so that they would not necessarily want to copy or trust the system.

\subsection{Qualitative Study: Usage and Usefulness of Showing and Suppressing Attribution Uncertainty}
\label{subsection:qualitative_study}
Although many prior studies have investigated the benefits and limitations of showing explanations of AI, it is unclear how communicating uncertainty in explanations impacts the user experience. In particular, we propose two different explanation techniques to handle attribution uncertainty, so it is important to understand how users will comparatively perceive and understand, and apply them. Therefore, we conducted a qualitative study by observing how users made decisions on the wine inspection task and interviewing their experience of different explanation techniques.\par
In particular, we were interested in the following objectives:
\begin{enumerate}[=\alph*), leftmargin=25pt]
    \item Verify that users can easily acquire a basic understanding of the baseline tornado plot for attribution explanations. We also explored any further needs on the explanation of the AI.
    \item Pilot a training method to teach users how to understand the different explanation techniques (Show or Suppress) and identify usability issues and misunderstandings of the visualizations. This helped us to further improve the experiment apparatus for evaluation with remote MTurk workers.
    \item Observe and enquire how users interpret and employ the various information (UI component) in each explanation technique to estimate their wine score and uncertainty, and to make decisions. For example, 
    \begin{itemize}
        \item Showing: Would showing subscore uncertainty influence the user’s score estimate, would she use both sides of the uncertainty bounds? Would showing uncertainty raise or lower the user’s confidence? 
        \item Suppressing: Would suppressing subscores and subscore uncertainties raise the user’s confidence and trust? How would it impact user's score estimation? For each objective, we ask follow-up questions to gain insight into the interviewees’ rationale.
    \end{itemize}
    \item Whether and why users have a preference for different explanation techniques, and for what circumstances.
\end{enumerate}

We recruited 12 interviewees from a university mailing list, aged 21 to 53 (M=25.9), 9 females. 6 were undergraduate students, 5 were graduate students, and 1 was alumni. They studied various majors, including nursing, medicine, psychology, real estate, biology, accountancy, chemical engineering, mechanical engineering, project and facility management, and computer science. Interviewees were conducted via the Zoom call with a shared screen to maintain social distancing during the COVID pandemic. Audio and screen movements were recorded with participant consent. Interviews took 35 to 57 minutes (M=48.5) and participants were compensated with a \$10 SGD Starbucks gift card for their time.\par
Through a semi-structured interview with participants using the explanation techniques in a web interface, we conducted the qualitative study with the following procedure: After consenting to the study, participants were introduced to the wine inspection task, and went through the relevant tutorials as described in Section \ref{subsection:exp_apparatus}. Each participant rated 1 to 2 wines, and, for each wine, sequentially used three out of four explanation techniques with different orders (e.g., Baseline $\rightarrow$ Show $\rightarrow$ ShowSuppress, Baseline $\rightarrow$ Suppress $\rightarrow$ ShowSuppress, Baseline $\rightarrow$ Show $\rightarrow$ Suppress, Baseline $\rightarrow$ Suppress $\rightarrow$ Show). We assigned interviewees to different arrangements based on whether we have saturated on our understanding of comparison between different explanation techniques. We asked the participant to think aloud while using each explanation technique. After each trial (one technique for one instance), we asked users questions regarding our aforementioned objectives. Due to limited time, we only asked questions that were relevant to the interviewee’s comments and that we had not saturated on. We ended by collecting demographic and background information.\par
Next, we discuss our findings and organize them in several themes.

\subsubsection{Understanding of Baseline, Show \& Suppress Attribution Explanations}
\label{subsubsection:understand_attribution}
We probed the participants' mental models of how the AI system works. We found that most participants understood that the system makes predictions by adding up the subscores in the attribution explanation of all chemical properties to the total quality score, and that subscores can contribute positively or negatively. This is consistent with our tutorial. P6 understood that \textit{“it gives some weightage for each 5 components, and the weightage for sulphites seems to be quite large $(subscore=-2.9)$ ... It shows quite clearly how those properties contribute negatively or positively to the overall wine quality.”} Participants could comfortably describe their decision-making process using the system subscores; e.g., P10: \textit{“For alcohol, I can see the subscore is highly negative $(-6.1)$. Then for pH $(0.3)$, Sulphates $(0.9)$ and SO2 $(1.1)$ are all slightly positive. But Vinegar Taint $(-1.3)$ is slightly negative. ... Based on these, I will likely reject this wine, because the contribution of Alcohol and Vinegar Taint in the negative range plays a very high contribution to the total score.”}\par

P10 found that seeing subscores and tornado plot increased her understanding of the system “because the reading and range of each property are quite different, we are not sure whether the value contribution is high or low to the final score and how much the reading contributes to the final score, but with the subscore we can get how much each property contributes to the final score. And by comparing the height of the red bars and the blue bars, it can allow us to make the final decision especially when it comes to uncertainty." From the interview, we learned that most participants are able to understand the model by reading attribution explanations, used the attribution to make decisions, trusted or questioned the system because of the attribution explanation.\par

However, when relating the system subscores to their general knowledge from the background table (Figure \ref{fig:walk_through}a), some participants questioned the validity of subscore values in the system. From the background table, P12 learned how the chemical property reading values could affect the wine quality score, and disagreed with the system on one of the subscores: \textit{“The percentage of the Alcohol is in the lower range, so I think it will affect more (than other chemical properties). …  I think the alcohol subscore should be affecting it even bigger like $-$2.8 (instead of the system shown $-$2.1)."} \par

Participants generally understood the Show and Suppress techniques regarding subscore uncertainty. Participants could understand how input uncertainties lead to subscore uncertainties. P6 observed that \textit{“(the subscore uncertainty) is proportional to which property has larger uncertainty ... there’s more uncertainty associated with (the uncertain) Alcohol in determining the quality, it is helpful in that way.”} P9 noted that subscore uncertainty may not be monotonically increasing with input uncertainty, and saw \textit{“that a small variation in alcohol can lead to larger subscore uncertainty."} Participants understood that Suppress would reduce the subscores of uncertain input properties and reallocate the attribution. P11 understood that \textit{“the system makes the subscores nearer to zero, so the alcohol and VT won’t affect the wine quality."}, and P2 noticed that \textit{“the Sulphate subscore (magnitude) becomes larger to compensate (the suppressing)."}\par

\subsubsection{Showing Uncertainty Is Diversely Helpful}
\label{subsubsection:show_uncertainty_help}
We found that participants had different approaches to adjust their score estimation after seeing the additional subscore uncertainty (compared to a previous baseline explanation).  P8 increase his estimate of the score uncertainty compared to his earlier estimate when seeing the Baseline explanation, because \textit{“apart from (the output uncertainty being) $\pm 1.9$, the subscores also have high uncertainty.”} Although subscore uncertainty is actually the decomposition of the score uncertainty, and the output score uncertainty does not change, showing uncertainty gave these participants the impression that the system is more uncertain than before, thus a few participants also increased their decision uncertainty after seeing subscore uncertainty. However, perceiving more uncertainty does not mean that the participants will be less confident in their decision. Most participants did not change their confidence after seeing the subscore uncertainty, with some even increasing their confidence. P3: \textit{“It (subscore uncertainty) makes me even more confident of my decision to reject because I have seen that alcohol (subscore) can fall anywhere within this range. It is even more uncertain, and I am more sure that I have made the safe decision to reject the wine.”} P11 increased her trust than Baseline after seeing subscore uncertainty \textit{“because without subscore uncertainty, just by reading the reading uncertainty, I can’t really guess how the reading affects the subscore. With the subscore uncertainty, I can understand how much the subscore will change.”} She also said that with subscore uncertainty \textit{“I am able to guess the score(output) more accurately.”} \par

Indeed many participants felt that showing subscore uncertainty helped them to understand the system and to make decisions. P6 mentioned that \textit{“it seems that (the system) quantifies the uncertainty contributed by each property more clearly (than Baseline).”} Showing subscore uncertainty helped participants to make better use of input uncertainty and understand where the output uncertainty is from. P7 mentioned that knowing subscore uncertainty makes the input uncertainty more helpful than Baseline, because subscore uncertainty helps him to \textit{“deduce”} the rating of the wine from the input uncertainty. With subscore uncertainty, P11 did \textit{“(find) the wine score uncertainty more helpful, because I can understand better where the 1.9 (output uncertainty) comes from. Previously (in Baseline), I thought it was a random estimation of uncertainty.”} \par

Instead of just considering uncertainty as an error, some participants used the uncertainty to interpret one-sided bounds to consider the best-case and worst-case scenarios. P3 and P12 wanted to be safe with their accept or reject decisions, and decreased their score after seeing subscore uncertainty. According to Prospect Theory \cite{kahneman2013prospect}, they demonstrated risk aversion to weight a negative outcome as more likely. For the uncertain property readings, i.e. Alcohol, they focused on the worst subscore that Alcohol could get by subtracting the subscore uncertainty value from their previous score estimate when viewing the Baseline explanation. On the other hand, some participants decided to accept borderline wine because the score \textit{“has a chance to be above 50”} (P1). P1 added the subscore uncertainty to the displayed score (upper bound estimate) and increased his score estimate. \par

One participant, P11, used the uncertainty to provide flexibility in her interpretation to align the system behavior with her belief. For an instance with low Vinegar Taint (VT), P11 felt that VT should have a positive subscore to be consistent with the background knowledge table. However, the subscore was calculated to be slightly negative ($-0.1$). Having seen the VT subscore uncertainty of 1.0, she happily increased her score estimate by 1.0, since she imagined that the VT subscore could actually be positive. With subscore uncertainty, P11 was \textit{“able to guess the score more accurately, because I think for the Vinegar Taint, if (the score) drops by 0.1, then I think it would be a bit inaccurate because (the knowledge table) says lower (VT reading value) will increase (the score), so I used the +1. I increase my score by 1 because for vinegar I plus one because I think it should increase the wine score. So I used the uncertainty of subscore to estimate my score. My estimation of VT subscore is 1.1.”} Therefore, the subscore uncertainty gave credence to her belief that the subscore could be positive, allowing her to be more assertive in her estimation.

\subsubsection{Suppressing Uncertainty Generally Improves Trust and Confidence}
\label{subsubsection:suppress_uncertainty_improve_trust}
We found that suppressing attribution uncertainty improved users' confidence in the decision and trust of the system. 
Compared to Baseline, when the subscore is suppressed, P8 was more confident in her decision \textit{“because the (score) uncertainty is lower, and I know that even if (I subtracted) 0.9 the (lower bound) score would be still above 50.”} P11 also increased her trust of the system compared to Baseline \textit{“because I think with the suppressing, it will eliminate the uncertainty in Alcohol and Vinegar Taint measurement, it will be more accurate.”} When comparing Show to ShowSuppress versions of the system, participants also increased their confidence and trust due to the decrease in score and subscore uncertainty. 

Some participants were comfortable with and appreciated the re-attribution due to uncertainty suppression. P7 thought the \textit{“suppressed subscore is pretty good because it can readjust and recalibrate the subscores when the uncertainty is high.”} However, some participants found that this "compensation" could be confusing because the attributions become mismatched with the background knowledge. For example, P2 compared the background table and system explanation and said: \textit{“If I have the knowledge table, I can see that although the (Alcohol subscore) uncertainty is low, it doesn’t mean the reading is a good one. ... Because after the suppression, the subscore is $-$0.1. Although subscore is less now (closer to 0, higher than Baseline subscore $-$6.1), but the reading of the property is quite far away from the reading of good wine"}\par

When participants estimated the score value and uncertainty with Suppress or ShowSuppress, they gave smaller uncertainty bounds than with Baseline and Show, which is consistent with the lower uncertainty due to suppression. P11 \textit{“(thought) the suppressed subscore will make the wine score more accurate.”} Participants were more decisive to give an estimate further from decision boundary 50, with lower uncertainty. After viewing that the scores with Suppress were higher above 50 than with Baseline, P6, P8, and P11 further increased their estimate.\par

\subsubsection{Preference to Suppress Uncertainty, but Divergent Opinions to Show Suppressed Uncertainty}
\label{subsubsection:strategy_preference}
We found that most participants appreciated the suppressed subscore, and some participants liked to see the subscore uncertainty when subscores are suppressed. P1 preferred Suppress over Show because \textit{“before suppressing, (the Show version) is not very accurate.”} P9 preferred ShowSuppress over Show because \textit{“I don’t really have to pay attention to how much the (input) uncertainty is anymore. I just assume the system already accounts for it to calculate the subscore and final score.”}  When comparing between Suppress and ShowSuppress, participants had divergent opinions. P8 thought the hybrid of showing and suppressing is better than suppressing only. He found the subscore in ShowSuppress to be more helpful than those of Suppress because \textit{“now I am given the uncertainty (of subscores), I can engage by how much the discrepancy there is, and I can take that into account in my overall decision of whether to accept or reject the wine.”} However, although he found ShowSuppress more helpful than Suppress, he had a lower trust of ShowSuppress than Suppress \textit{“because now that I’m given the uncertainty of the subscore, I know that the machine has a certain discrepancy, so I can’t fully trust the machine.”} Relatedly, P4 preferred not being shown the subscore uncertainty (i.e., preferred Suppressed over ShowSuppressed) \textit{“because there will be too much information.”}\par

\subsubsection{Opportunities to Improve Interpretability}
\label{subsubsection:opportunity_on_interpretability}
Although we have found that participants understood and made good use of different versions of explanations, we noticed that they may occasionally misuse the user interface, questioned the system’s uncertainty handling and asked for more information. \par
Although attribution explanations are popular and intuitive \cite{Ribeiro2016WhyClassifier,Lundberg2017APredictions,bach2015pixel}, some participants were curious to learn deeper explanations to fully understand the system. With Baseline, P6 found the subscores unhelpful for her decision making because she did not know how each subscore was calculated and what was its relationship between the input value and subscore value. However, she could not articulate what other information would help her. We prompted that a line chart visualization\footnote{For example, with partial dependence plots or generalized additive models (GAM).} or rules\footnote{As generated by association rule mining, decision tree learning, etc.} could provide deeper explanations, but she felt they would be too much information. We avoided mentioning of machine learning concepts, such as multi-factor linear regression, feature vector spaces, and neural networks to avoid jargon with lay users. This is beyond the scope of our study and relates to AI literacy in the general public \cite{long2020ai}. Nevertheless, we limited our study to the 2nd level of the Self-Explaining Scorecard \cite{Klein2020} to more deeply investigate the impact of uncertainty even at such a low level of explanation.\par

Similarly, some participants wanted to know how subscore uncertainties (in Show) were calculated. We avoided discussing error propagation with Monte Carlo simulation, hypothetical instances and sampling, since that would require a more advanced understanding of statistics which is not common in lay knowledge. P9 said that it is good to know the subscore uncertainty, but she wanted to know more: \textit{“I don’t know how much the change in reading would lead to the change in the subscore. ... I see that a small variation in alcohol can lead to larger subscore uncertainty, but I don’t know the direct relationship.”} This suggests an interest in counterfactual explanations \cite{miller2019explanation}, which is beyond the scope of our study.

As for Suppress and ShowSuppress, we identified two issues: 1) it requires some effort to understand how suppressing works, and 2) participants would like to personalize the extent of suppression. Regarding the learning curve, participants appreciated the suppressing but some of them found it confusing at first. P7, who used Suppress, felt that \textit{“initially it is confusing to me, it takes me a while to understand the suppression.”} Within about a minute, P7 switched repeatedly between Baseline and Suppress to understand. This inspired us to implement the hovering interaction to help users perceive the changes in subscores with and without suppression. Regarding the personalized suppression preference, most participants were fine with strong suppression. For an instance where the Alcohol subscore was suppressed to almost zero, P8 found it acceptable, because \textit{“the uncertainty of Alcohol subscore is very high, so I wouldn't use Alcohol in my decision making."}. However, P6, who used ShowSuppress, felt that the subscore should not be suppressed too much (i.e., all the way to zero), and suppressing should be a little or by half because \textit{“I don’t know if the subscore suppressing is actually good and accurately evaluate the quality of each component.”}\par

\subsubsection{Clarity and Usability Issues and Fixes}
\label{subsubsection:usability_issues}
We identified several minor clarity and usability issues that we fixed in the subsequent quantitative study. P4 was confused about why the subscores were so small and believed that they had to be larger to be consistent with the background knowledge. We clarified that the presented cases were borderline rather than typical, so the subscores would tend to be small. P3 also was interested to know how the subscore is calculated and tried to rationalize it by counting the number of positive bars and subtracting the number of negative bars, but not their subscore values. We clarified that the total score is derived from the sum of subscores, not the count of properties. Another participant preferred to see subscores as additive partial attributions from 0 that sum to the total score, rather than an average value and attribution differences from average. Designers can consider this alternative representation in the future. However, we point out that this may limit an intuitive comparison between suppressed and default attributions as deviations from the mean. P1 found that the subscore uncertainty did not add up to the total score uncertainty as a simple sum. We clarified that the subscores should be a sum of squares to get the square of total score uncertainty. Other minor fixes include amplifying the contrast in the highlight colors of the uncertainty numbers, scaling all tornado plots to the same range.\par

\subsection{Quantitative Study: Impact of Showing and Suppressing Attribution Uncertainty on User Performance and Perception}
\label{subsection:quantitative_study}
Following the formative qualitative study where we interviewed participants on the usage and rationale of the explanation techniques, we conducted a summative quantitative study to evaluate how Showing or Suppressing attribution uncertainty affects user decision performance and perception about the AI system and explanation. In this section, we describe our measures, hypotheses, and results. We used the same primary independent variable of explanation techniques in the quantitative study as in the qualitative study.

\subsubsection{Random Variables}
\label{subsubsection:RVs}
Besides the Independent Variables and Controlled Variables described in Section \ref{subsection:exp_treatment}, we tracked several Random Variables in the quantitative study to help calculate dependent variables and to account for potential confounds. 

We varied system predicted and actual scores as follows:
\begin{enumerate}[leftmargin=25pt]
    \item Trial Sequence (1 to 15): to describe the order of the tasks received by participants. This is only used in the quantitative study.
    \item System Score: The system wine quality score, denoted as $Score_{System}$. For Suppress and ShowSuppress which show the suppressed and default subscores, we also track the baseline score, denoted as $Score_{SystemBaseline}$.
    \item System Score Uncertainty of Baseline: The uncertainty of system quality score from the Baseline, denoted as $ScoreUncertainty_{SystemBaseline}$.
    \item Actual Score: The ground truth quality score, denoted as $Score_{Actual}$.
    \item Actual Decision (2 levels): Accept or Reject. Whether the Actual Score is greater than 50 or not. The Actual Decision of half of the selected wine instance in the study is Accept.
\end{enumerate}

Confounds due to user background include:
\begin{enumerate}[resume,leftmargin=25pt]
    \item Uncertainty Decisiveness: We ask participants three 7-point Likert scale questions about their ability to make decision uncertainty \cite{Greco2001CopingMeasure}, and this measures whether they were higher or lower than neutral.
    \item Uncertainty Tolerance: We ask participants three 7-point Likert scale questions about their uncertainty Tolerance \cite{Carleton2007FearingScale}.
\end{enumerate}

\subsubsection{Dependent Variables}
\label{subsubsection:DVs}
We measured and computed various dependent variables to understand the participants’ decision correctness, task time, and confidence in decision making, trust\footnote{We did not evaluate trust appropriateness on the system and controlled the system correctness to be good for all instances because: 1) to learn when it is appropriate to trust, the user will need ground truth feedback to assess system correctness; 2) it takes many trials to observe and learn even a simple system failure pattern of a toy system \cite{Bansal2019}; 3) if ground truth feedback is unknown, users will need strong domain knowledge or technical background to identify system errors, which is not easy even for experts without any social discussion and experimenting, not to mention the lay users. Refer to \cite{Kaur2020, Lim2011InvestigatingApplications} for some results that users tend to over-trust AI even when they may be inappropriate; this is beyond the scope of our current study. We defer the study of uncertain explanations of inappropriate model behavior to future work.} in the system’s inference and helpfulness of the system. We recorded measures of the system prediction and actual (ground truth) scores used to calculate subsequent dependent variables: We further computed dependent variables to more determine the impact on relevant outcomes, such as decision quality. See Table \ref{tab:DV} for definitions and calculations.

\begin{table}[]
\footnotesize
\begin{tabular}{lp{9.4cm}}
\hline
\textbf{Dependent Variable} & \textbf{Definition} \\ \hline
Log(Time) & Logarithmic transform of task time per wine decision trial. \\ \hline
Decision Closeness & Negative of difference between the user’s score and system’s displayed score, i.e., $-|Score_{User}-Score_{System}|$ \\ \hline
\multirow{2}{*}{\makecell[l]{Relative\\ Decision Uncertainty}} & Ratio of user’s score uncertainty and system’s baseline uncertainty, \\
& i.e., $ScoreUncertainty_{User}/ScoreUncertainty_{SystemBaseline}$. A ratio $\textless  1$ means that the user perceives lower uncertainty than the system. \\ \hline
Decision Quality  & Negative of difference between the user’s score and actual (ground truth) score, i.e., $-|Score_{User}-Score_{Actual}|$ \\ \hline
\multirow{2}{*}{\makecell[l]{Sum(Decision\\ Correctness)}} & Sum of correct decisions (accept/reject) for all 15 trials per system version. \\ \hline
Trust & \multicolumn{1}{l}{\multirow{3}{*}{\makecell[l]{7-point Likert scale rating\\(strongly disagree to strongly agree).}}} \\ \cline{1-1}
Confidence & \multicolumn{1}{l}{} \\ \cline{1-1}
Helpfulness & \multicolumn{1}{l}{} \\ \hline
\end{tabular}
\caption{Dependent variables measured and analyzed in the quantitative user study.}
\label{tab:DV}
\end{table}

\subsubsection{Hypotheses}
\label{subsubsection:hypo}
We hypothesized how different explanation techniques would influence each dependent variable. See Table \ref{tab:hypotheses_result} middle column. For Decision Time (H1), both Show and Suppress provide additional information\footnote{Show adds subscore uncertainty numbers, Suppress displays the UI twice to compare with Baseline}, so we hypothesized that they will take more time to read than Baseline, and their combination of them ShowSuppress (SS) will take longer than either separately. Since Show has higher uncertainty than Suppress, we hypothesize that it would make users more uncertain in decision making and would slow them down than Suppress and Baseline. We hypothesize that Decision Closeness (H2), Trust (H6), Confidence (H7) and Helpfulness (H8) will be correlated and be ordered by the most suppressed and transparent (SS) as the best, followed by Suppress, Baseline, and then Show as last, because showing uncertainty can lead to distrust \cite{Lim2011InvestigatingApplications}. For Relative Decision Uncertainty (H3), we hypothesized that showing uncertainty (Show, ShowSuppress) would lead to higher reported uncertainty than unshown (Baseline, Suppress, respectively) because of increased awareness about uncertainty; suppressing (Suppress, ShowSuppress) will have lower reported uncertainty due to the uncertainty that is communicated being smaller. We hypothesize that Decision Quality (H4) and Decision Correctness (H5) will be correlated and suppression will lead to better decisions due to the model correctness is controlled as good and the suppressing improved expected faithfulness to model in Section \ref{subsection:sim_result_faith_dist} (Figure \ref{fig:LIME_E[F]}), and that showing more information can help the decision to be correct.\par

\begin{table}[!h]
  \includegraphics[width=13.8cm]{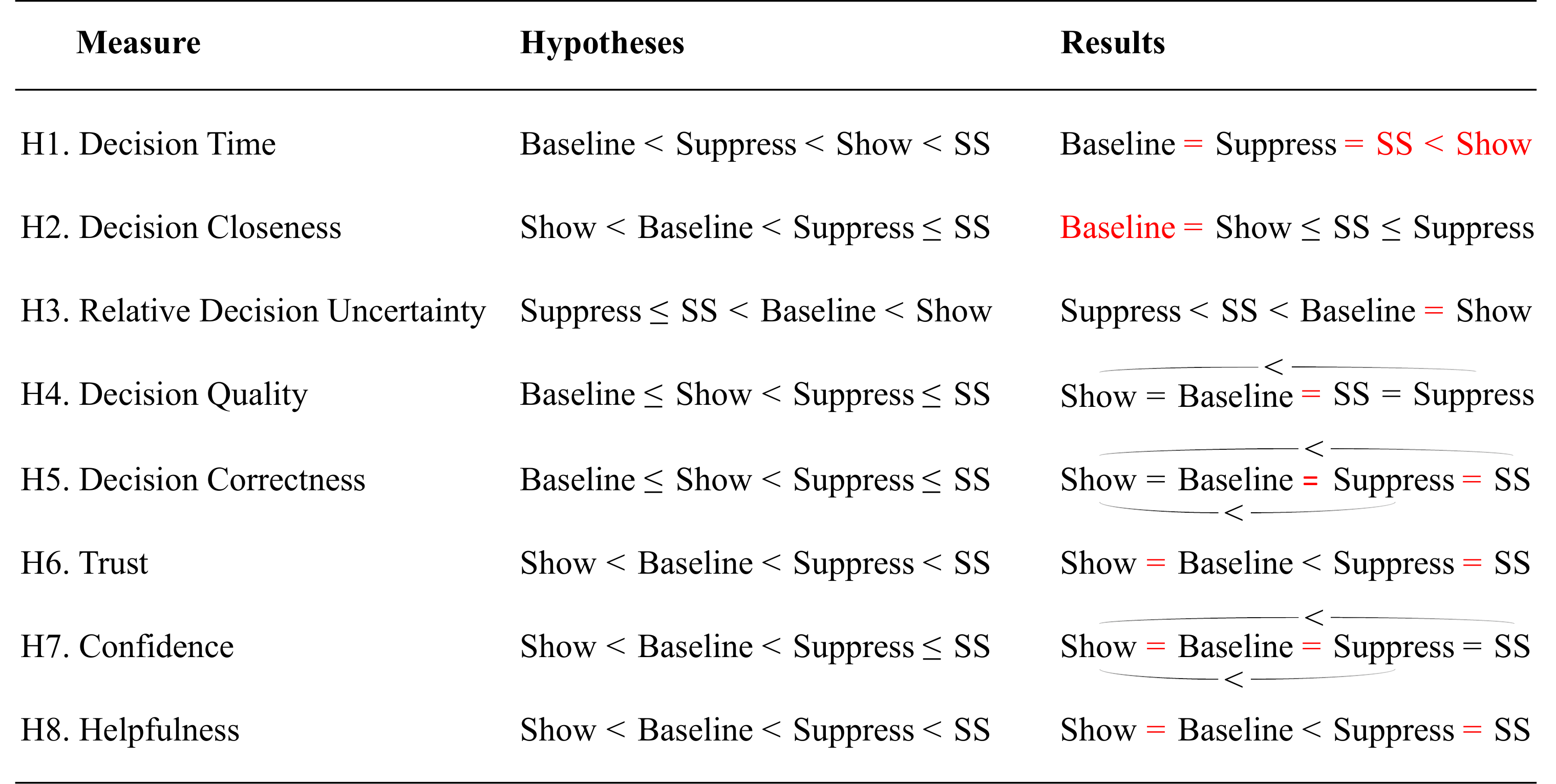}
   \caption{Hypotheses and results from quantitative analysis in terms of experiment hypotheses. SS refers to the ShowSuppress technique. Sign “$\textless $” indicates a significant difference at ${\rm p\textless .0001}$, “$\leq$” indicates marginal difference at ${\rm p\textless .001}$, and “$=$” indicates no significant difference at ${\rm p\textgreater .01}$. Although adjacent comparisons are not significant, other comparisons may be significantly different and this is indicated as branched lines with “$\textless $”. {\color[HTML]{ea3223}Red} text and signs indicate results that do not agree with the hypotheses. }
  \label{tab:hypotheses_result}
\end{table}

%\begin{figure}[ht]
%    \centering
%    \includegraphics[width=12cm]{fig:hypotheses.pdf}
%    \caption{Hypotheses. Each row is the hypothesis for a dependent variable. SS refers to the ShowSuppress version of system explanation. }
%    \label{fig:hypotheses}
%\end{figure}

We pre-registered our experiment variables, hypotheses, data exclusion criteria and analysis methods at AsPredicted\footnote{The anonymized pre-registration document is available at https://aspredicted.org/blind.php?x=ub3mt2} before we collected data. Decision Correctness is an additional dependent variable that we investigated besides what we pre-registered. It is the classification version of Decision Quality, and they have the same hypothesis.\par

\subsubsection{Statistical Analysis and Results}
\label{subsubsection:analysis_results}
We recruited participants from Amazon Mechanical Turk (MTurk). To ensure the data quality, our survey was only available to Mturk workers with high qualification (at least 5,000 completed HITs with above 97\% approval rate). Participants were compensated US\$ 5.00 once they pass the screening quiz and completed the survey in around 35 minutes. Of the 270 MTurk workers who attempted the survey, 147 passed the screening quiz to complete the survey (54.4\% pass rate). If a participant made correct decisions for more than 24 instances (out of 30), she would get a bonus of \$0.25 per correct instance. A participant can get up to US\$7.00 payment. To manage MTurk quality issues \cite{Kennedy2018TheCrisis}, we had the following pre-registered data exclusion criteria:
\begin{enumerate}
    \item Per-trial responses with abnormal outlier completion times (too long or too short).
	\item Per-trial responses with inconsistent answers (e.g. accept a wine but rating score is below 50)
	\item Per-condition responses from participants who failed more than one third pre-condition comprehension questions
	\item Responses that are too identical across trials and questions, which suggests participants rushing through and not answering questions carefully.
	\item Participants who provide meaningless or nonsensical free text answers
\end{enumerate}
We recruited 147 participants (32.43\% female, median age 36 years) who passed the screening quiz and completed the survey but eliminated a few responses not satisfying the exclusion criteria. Ultimately, 131 respondents were included for analysis. \par
\begin{table}[t!]
\small
\centering
\begin{tabular}{llrcc}
\hline
\textbf{Response} & \textbf{Linear Mixed Effects Model} & \textbf{p\textgreater{}F} & $\mathbf{R^2}$ & $\bm{f^2}$\\ 
& (+ Participant as random effect) & & &\\ \hline
Log(Time) & Technique + & \textless .0001 & .530 & 1.13\\ 
 & Trial Sequence & \textless .0001 & & \\
\hline
Decision Closeness & Technique + & .0003 & .495 & 0.98\\
 & {\color[HTML]{9B9B9B} Trial Sequence} & {\color[HTML]{9B9B9B} \textit{n.s.}} & & \\ \hline
 & Technique + & \textless .0001 & .649 & 1.85\\ 
\multirow{-2}{*}{\begin{tabular}[c]{@{}l@{}}Decision Relative \\ Uncertainty\end{tabular}} & {\color[HTML]{9B9B9B} Trial Sequence} & {\color[HTML]{9B9B9B} \textit{n.s.}} & & \\ \hline
Decision Quality & Technique + & \textless .0001 & .594 & 1.46 \\ 
 & {\color[HTML]{9B9B9B} Trial Sequence +} & {\color[HTML]{9B9B9B} \textit{n.s.}} & & \\
 & Actual Score Closeness + & \textless .0001 & & \\ 
 & {\color[HTML]{9B9B9B} Technique $\times$ Actual Score Closeness} & {\color[HTML]{9B9B9B} \textit{n.s.}} & & \\ \hline
Sum(Decision & Technique & \textless .0001 & .714 & 2.49\\ 
Correctness) & & & & \\ \hline
Trust & Technique & \textless .0001 & .721 & 2.59\\ \hline
Confidence & Technique & .0002 & .744 & 2.91\\ \hline
Helpfulness & Technique & \textless .0001 & .524 & 1.10\\ \hline
\end{tabular}
\caption{Statistical analysis of dependent variable responses and factors (one per row), fit as linear mixed effects models, all with Participant as random effect. Per-trial dependent variables were modeled with Technique and Trial Sequence as fixed effects, and per-condition dependent variables were modeled with only Technique as fixed effect. Decision Correctness was modeled as the per-condition metric Sum(Decision Correctness per-trial) to allow more insightful analysis with parametric modeling, Decision Correctness is a binary variable (0 or 1). Decision Quality depends on the actual (ground truth) score of the instance, so we add Actual Score Closeness (negative absolute difference of actual score to decision boundary 50) as fixed effect and Technique $\times$ Actual Score Closeness as interaction effect. {\color[HTML]{9B9B9B} \textit{n.s.}} refers to no significant difference at ${\rm p\textgreater .01}$. ${\rm p\textgreater F}$ refers to the significance level of an ANOVA for each fixed effect. ${\rm R^2}$ and $f^2$ refer to the model’s coefficient of determination and Cohen’s $f^2$, respectively. All models have strong effect size with $f^2\textgreater 0.35$ and ${\rm R^2} \gtrapprox 0.5$ indicating that they explain at least 50\% of variance.}
\label{tab:stat_analysis}
\end{table}
We divided the dependent variables into per-trial and per-condition variables. We fit one multivariate linear mixed effects models for per-trial variables, and another for per-condition variables. Table \ref{tab:stat_analysis} describes the details of the models with fixed, interaction, and random effects. Generally, all models had good to very good fit (high R2).\par
Due to a large number of comparisons in our analysis, we consider differences with ${\rm p\textless .001}$ as significant and ${\rm p\textless .005}$ as marginally significant. Most significant results are reported as ${\rm p\textless .0001}$. This is stricter than a Bonferroni correction for $k=50$ comparisons (significance level$=.05/50$).\par
Following advice from Dragicevic \cite{dragicevic2016fair}, we performed a logarithmic transform on time, Log(Time), which is commonly done for time measurements \cite{Keene1995TheSpecial} to correct for positive skewness in time measurements and mitigate timing outliers \cite{Sauro2010AverageReport}. We had analyzed task time with a logarithmic transform to be able to fit a linear mixed effects model with Participant as a random effect. To verify the effect without the log-normality assumptions, we performed non-parametric tests on task time to compare the medians between conditions\footnote{A Kruskal-Wallis H test by ranks found a significant difference between the task Time across Technique (${\rm p\textless .0001}$), and pairwise Wilcoxon Rank Sum tests found that Show technique had higher task Time than all other techniques (all ${\rm p\textless .0001}$). The less statistically efficient Median test of the number of points above the median was also significant (${\rm p\textless .0001}$), and pairwise Median tests also found that Show had higher task Time than None, Baseline, Suppress, and ShowSuppress (${\rm p=.0082}$, ${\rm \textless .0001}$, ${\rm \textless .0001}$, $.0067$, respectively).}. These verified the significant differences found in the parametric model. Note that since the non-parametric tests do not account for the Participant as a random effect, so it does not consider the reduced variance due to individuals with the within-subjects experiment design.\par
\begin{figure}[ht!]
    \centering
    \includegraphics[width=14cm]{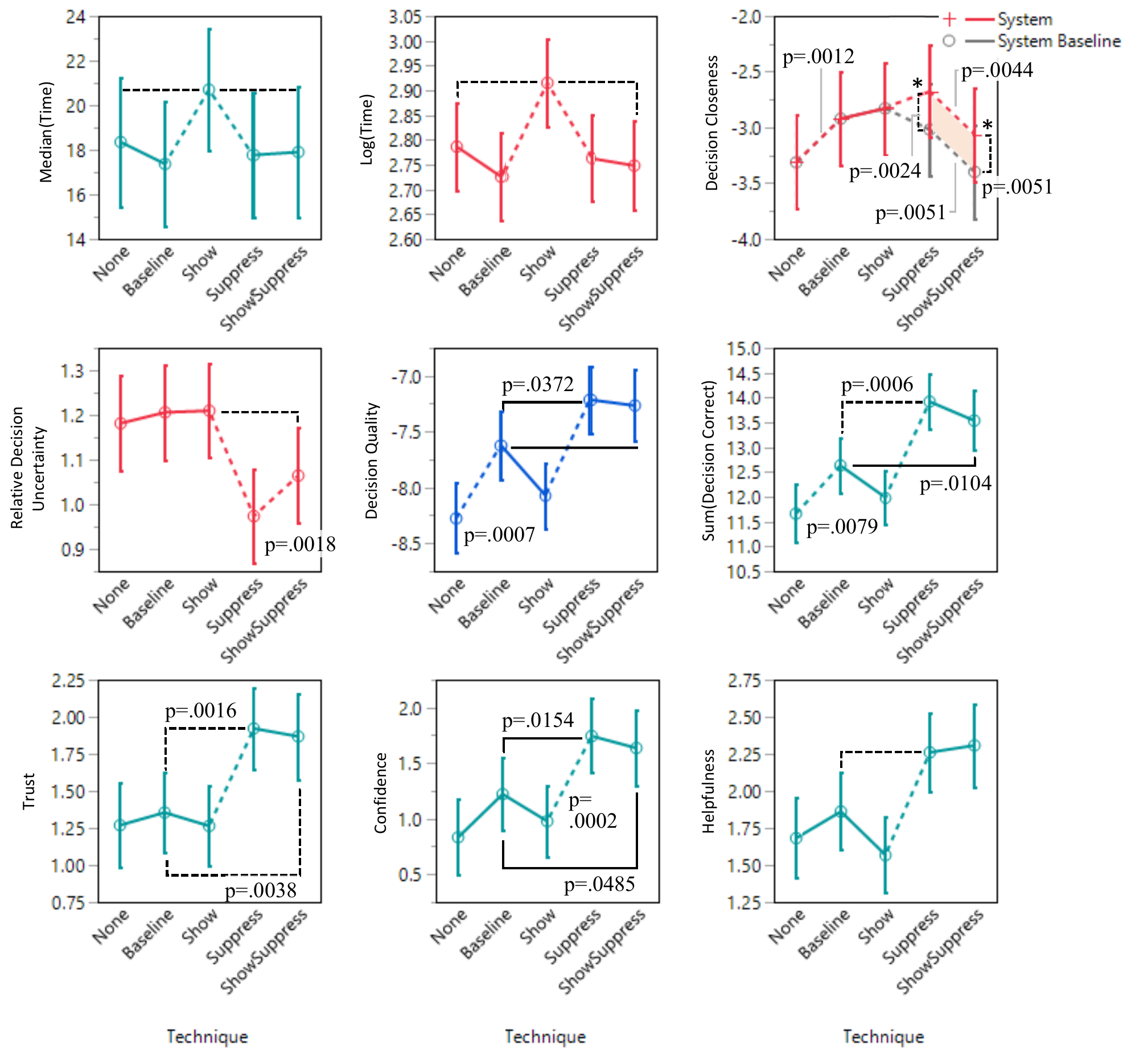}
    \caption{Results showing how explanation types influence 8 dependent variables. Dotted lines indicate extremely significant ${\rm p\textless .0001}$ comparisons, otherwise very significant as stated; solid lines indicate no significance at ${\rm p\textgreater .01}$. Error bars indicate a 90\% confidence interval. Distance Closeness includes an additional {\color[HTML]{9B9B9B} \textbf{grey}} line to indicate closeness to the Baseline score that is shown with the suppressed score for suppressed explanation types. Per-trial dependent variables are shown as {\color[HTML]{ee3223} \textbf{red}} lines, and per-system dependent variables are shown as {\color[HTML]{439897} \textbf{teal}} lines. Decision Quality was modeled with Actual Score Ambiguity (how close to decision boundary 50) as a fixed effect, and this is shown as a {\color[HTML]{1a50ff} \textbf{blue}} line.}
    \label{fig:stat_result}
\end{figure}
Figure \ref{fig:stat_result} shows the details of statistical analysis results with Technique as the main fixed effect. We describe key insights in terms of practical differences. Note that many of the numbers are in terms of our experiment apparatus (15 trials per condition, with borderline wine instances with high system estimate uncertainty) and the wine rating task (e.g., score from $0-100$, but likely within $30-70$). Therefore, we also interpret the results in terms of percent differences to give clarity to how the results could scale to other contexts. We describe findings for each dependent variable:\par
\textbf{Task Time:} Using Show explanations is slower than Baseline explanations by 3.01s (M=18.3 vs. 15.3s), which is 19.7\% slower. Suppress or ShowSuppress explanations do not take significantly more time than Baseline.\par
Decision Closeness and Decision Quality have units in terms of score difference, for scores that vary from 0 to 100. We relate them to the system decision quality (the negative difference between system baseline score and actual score, i.e.,$-|Score_{SystemBaseline} - Score_{Actual}|$, which has Medians -4.01. \par
\textbf{Decision Closeness:} Using Suppress and ShowSuppress increase Decision Closeness from Baseline -2.92 by 0.350 and 0.328 (11.1\% and 9.7\%), respectively. In the suppressed conditions, since both suppressed versions and corresponding unsuppressed versions were provided for comparison, we compared the two distances from user score to suppressed and unsuppressed system scores by ratio: $|Score_{User}-Score_{System}|/|Score_{User} - Score_{SystemBaseline}|$, and the median is 0.889 (for 67.5\% of participants the distance ratio is smaller than 1), indicating that participants tend to choose a score closer to the suppressed score than the baseline score when in any suppressed condition, and there was no significant difference between ShowSuppress and Suppress. Using Show explanations do not change Decision Closeness. For Suppress and ShowSuppress, Distance Closeness to System is marginally higher than Distance Closeness to Baseline, suggesting that participants did prefer to follow the suppressed score (System) than the unsuppressed one (Baseline).\par
\textbf{Decision Quality:} Using any explanation improves Decision Quality compared to None. Using Suppress or ShowSuppress increases Decision Quality more than using Show from -8.08 by 0.860 and 0.810 (10.7\% and 10.0\%), respectively. However, Show, Suppress and ShowSuppress are not significantly different from Baseline.\par
\textbf{Relative Decision Uncertainty:} Using Suppress and ShowSuppress decreases user perceived Relative Decision Uncertainty from Baseline at 1.21 by 0.232 and 0.142 (19.3\% and 11.7\%), respectively. When using explanations without suppression (None, Baseline, Show), users perceived higher decision uncertainty than what the system showed, i.e., $\textgreater 1$. Relative Decision Uncertainty for Suppress is significantly lower than ShowSuppress (M=0.97 vs. 1.06).\par
\textbf{Sum(Decision Correct):} Using Suppress and ShowSuppress increases decision correctness compared to Show from 12.0 correct answers by 1.94 and 1.56 (16.2\% and 13.0\%), respectively. Only Suppress significantly improves decision correctness compared to Baseline, and from 12.6 correct answers by 1.29 (10.2\%). Show explanations do not improve decision correctness compared to Baseline.\par
\textbf{Trust:} Using Suppress and ShowSuppress increases Trust rating compared to Baseline from 1.35 (above 1=“Slightly Agree”) to 1.92 and 1.87 (both almost 2=“Agree”). Show does not improve Trust compared to Baseline.\par
\textbf{Confidence:} Using Suppress increases Confidence rating compared to Baseline from 1.22 (above 1=“Slightly Agree”) to 1.74 (towards 2=“Agree”). Show and ShowSuppress do not improve Trust compared to Baseline.\par
\textbf{Helpfulness:} Using Suppress and ShowSuppress increases Helpfulness rating compared to Baseline from 1.86 (below 2=“Agree”) to 2.26 and 2.30 (above 2=“Agree”). Show does not improve Trust compared to Baseline.\par

\subsubsection{Interpreting Results With Respect to Hypotheses}
\label{subsubsection:results_interpret}
Next, we interpret our results in terms of our hypotheses of how different explanation techniques could be beneficial or detrimental to user performance and opinion. Table \ref{tab:hypotheses_result} compares the results with the hypotheses and highlights the differences. The key insights are:
\begin{enumerate}
    \item ShowSuppress is not the slowest to use, but Show is instead, suggesting that the trust and confidence gained due to suppressing uncertainty helped participants to be more decisive. Indeed, participants are no slower using ShowSuppress than Baseline or Suppress.
    \item Though we expected the Show technique to have the lowest Decision Closeness, Trust, Confidence, and Helpfulness because it reveals model uncertainty, it is not significantly worse than Baseline.
    \item Suppress and ShowSuppress do reduce perceived Decision Uncertainty, improve Trust and Helpfulness compared to Baseline, its impact on Decision Quality, Decision Correctness and Confidence is less significant.
\end{enumerate}

In summary, showing attribution uncertainty that is large may not improve user performance and perception, and yet costs more task time. Therefore, it is more important to mitigate uncertainty first. Showing the uncertainty suppression achieves the same performance and perception improvement as suppressing uncertainty compared to baseline, and does not suffer from the detrimental effects of showing unsuppressed attribution uncertainty.

\subsubsection{Factor Analysis of Showing and Suppressing}
\label{subsubsection:factor_analysis}
\begin{table}[ht]
\small
\centering
\begin{tabular}{lrrr}
\hline
\multirow{2}{*}[-3mm]{\textbf{Response}} & \multicolumn{3}{c}{\textbf{Fixed Effects (p \textgreater F)}} \\ \cline{2-4} 
 & \textbf{Showing} & \textbf{Suppressing} & \textbf{\begin{tabular}[c]{@{}l@{}}Showing $\bm{\times}$ \\ Suppressing\end{tabular}} \\ \hline
Log(Time) & \textless .0001 & .0014 & \textless .0001 \\ \hline
Decision Closeness & {\color[HTML]{9B9B9B} \textit{n.s.}} & {\color[HTML]{9B9B9B} \textit{n.s.}} & {\color[HTML]{9B9B9B} \textit{n.s.}} \\ \hline
Decision Closeness (to Baseline) & {\color[HTML]{9B9B9B} \textit{n.s.}} & .0004 & {\color[HTML]{9B9B9B} \textit{n.s.}} \\ \hline
Relative Decision Uncertainty & {\color[HTML]{9B9B9B} \textit{n.s.}} & \textless .0001 & {\color[HTML]{9B9B9B} \textit{n.s.}} \\ \hline
Decision Quality & {\color[HTML]{9B9B9B} \textit{n.s.}} & \textless .0001 & {\color[HTML]{9B9B9B} \textit{n.s.}} \\ \hline
Sum(Decision Correctness) & {\color[HTML]{9B9B9B} \textit{n.s.}} & \textless .0001 & {\color[HTML]{9B9B9B} \textit{n.s.}} \\ \hline
Trust & {\color[HTML]{9B9B9B} \textit{n.s.}} & \textless .0001 & {\color[HTML]{9B9B9B} \textit{n.s.}} \\ \hline
Confidence & {\color[HTML]{9B9B9B} \textit{n.s.}} & \textless .0001 & {\color[HTML]{9B9B9B} \textit{n.s.}} \\ \hline
Helpfulness & {\color[HTML]{9B9B9B} \textit{n.s.}} & \textless .0001 & {\color[HTML]{9B9B9B} \textit{n.s.}} \\ \hline
\end{tabular}
\caption{Results of t-tests and 2-way ANOVA on the factors of the explanation techniques for Showing or Suppressing attribution uncertainty in explanations. }
\label{tab:2by2}
\end{table}
The four explanation Techniques (excluding None) can also be considered in terms of the two factors Showing and Suppressing. For technique Show and ShowSuppress, factor Showing is true; for technique Baseline and Suppress, factor Showing is false. For technique Suppress and ShowSuppress, factor Suppressing is true; for technique Baseline and Show, factor Suppressing is false. We performed contrast t-tests with fixed effects Showing, Suppressing, and contrast interaction effect Showing $\times$ Suppressing. Table \ref{tab:2by2} shows the results of contrast t-tests and 2-way ANOVA on these factors. There are significant fixed and interaction effects for Log(Time) as previously seen in Figure \ref{fig:stat_result}. All explanation Techniques achieve indistinguishable levels of Decision Closeness. For all other dependent variables, there is no effect of Showing attribution uncertainty and no interaction effects, but there are significant effects Suppressing attribution uncertainty. 

\subsubsection{Some Differences Due to Uncertainty Intolerance}
\label{subsubsection:uncertainty_tolerance}
\begin{table}[!ht]
\small
\centering
\begin{tabular}{llrcc}
\hline
\textbf{Response} & \textbf{Linear Mixed Effects Model} & \textbf{p\textgreater{}F} & $\mathbf{R^2}$ & $\bm{f^2}$ \\ 
& (+ Participant as random effect) & & & \\ \hline
 Decision & Technique + & .0002 & .496 & 0.99\\ %\cline{2-5} 
 Closeness & {\color[HTML]{9B9B9B} Trial Sequence +} & {\color[HTML]{9B9B9B} \textit{n.s.}} & & \\ %\cline{2-5} 
 & {\color[HTML]{9B9B9B} Uncertainty Tolerant +} & {\color[HTML]{9B9B9B} \textit{n.s.}} & & \\ %\cline{2-5} 
 & {\color[HTML]{9B9B9B} Uncertainty Decisive +} & {\color[HTML]{9B9B9B} \textit{n.s.}} & & \\ %\cline{2-5} 
 & Technique $\times$ Uncertainty Tolerant + & \textless .0001 & & \\ %\cline{2-5} 
 & {\color[HTML]{9B9B9B} Technique $\times$ Uncertainty Decisive} & {\color[HTML]{9B9B9B} \textit{n.s.}} & & \\ \hline
 Confidence & Technique + & \textless .0001 & .744 & 2.91\\ %\cline{2-5} 
 & {\color[HTML]{9B9B9B} Uncertainty Tolerant +} & {\color[HTML]{9B9B9B} \textit{n.s.}} & & \\ %\cline{2-5} 
 & {\color[HTML]{9B9B9B} Uncertainty Decisive +} & {\color[HTML]{9B9B9B} \textit{n.s.}} & & \\ %\cline{2-5} 
 & Technique $\times$ Uncertainty Tolerant + & .0032 & & \\ %\cline{2-5} 
 & {\color[HTML]{9B9B9B} Technique $\times$ Uncertainty Decisive} & {\color[HTML]{9B9B9B} \textit{n.s.}} & & \\ \hline
\end{tabular}
\caption{Statistical analyses including user uncertainty intolerance and decisiveness as linear mixed effects models with fixed and interaction effects (one per row), and with Participant as a random effect. {\color[HTML]{9B9B9B} \textit{n.s.}} refers to no significant difference at ${\rm p\textgreater .01}$. ${\rm p\textgreater F}$ refers to the significance level of an ANOVA for each fixed effect. ${\rm R^2}$ and $f^2$ refer to the model’s coefficient of determination and Cohen’s $f^2$, respectively. All models have a strong effect size with $f^2\textgreater 0.35$ and ${\rm R^2} \gtrapprox 0.5$ indicating that they explain at least 50\% of the variance.}
\label{tab:intolerance_analysis}
\end{table}
As an additional analysis, we investigated whether differences in individual risk tolerance influenced their perception and use of the explanation techniques. We note that the survey questions on personality types are context-free and prone to be inaccurate to characterize real-world intolerance perception. Nevertheless, this is exploratory analysis can provide some formative insights. We found that the six survey questions on risk tolerance were correlated and performed the common factor analysis using maximum likelihood with varimax rotation to group responses into two factors — uncertainty tolerance (first 3 questions) and uncertainty decisiveness (last 3 questions), see Figure \ref{fig:uncertainty_strategy}. The final number of factors are statistically significant by the Bartlett Test of Sphericity (all ${\rm p\textless .0001}$) and explains 66.5\% of the response variance. To support the analysis of interaction effects, we further binarized the factors into whether the response is above the median to derive the fixed effects Uncertainty Tolerant and Uncertainty Decisive, respectively. We fit a multivariate linear mixed effects model as described in Table \ref{tab:intolerance_analysis}. \par
\begin{figure}[ht!]
    \centering
    \includegraphics[width=14cm]{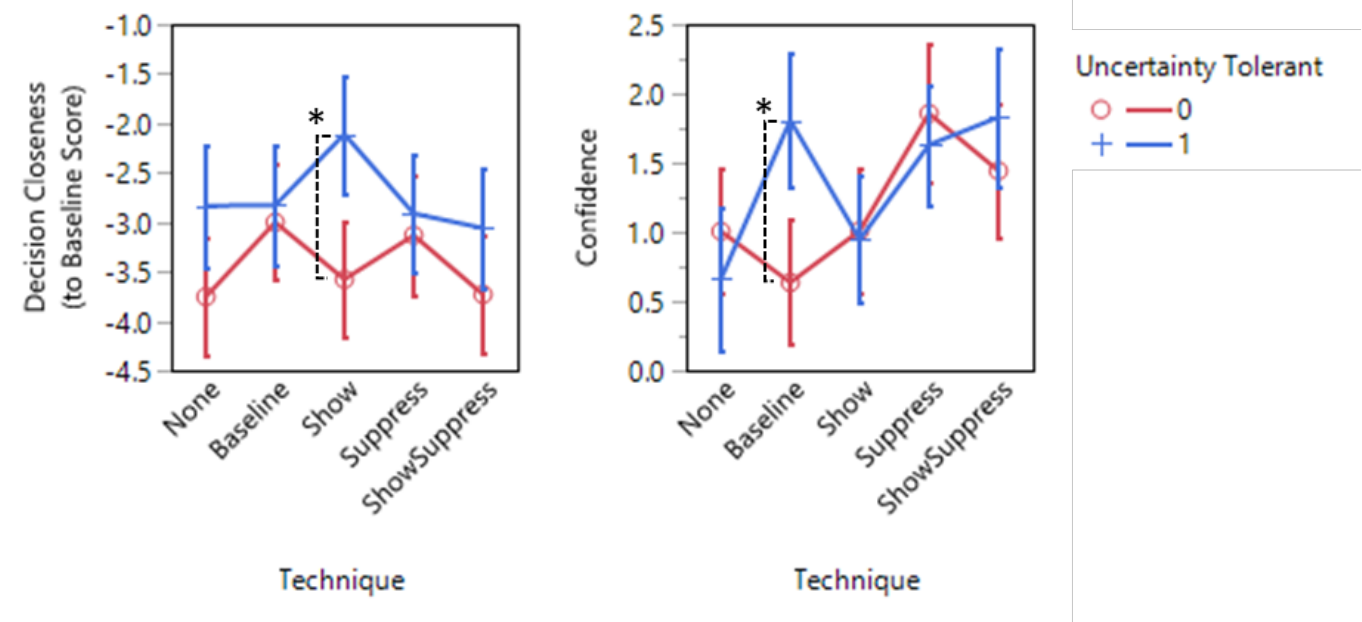}
    \caption{Results showing how participants who are more Uncertainty Tolerant or less tolerant are differently influenced by different explanation types. Significant contrast t-test effects are indicated with \* and dotted line. Error bars indicate a 90\% confidence interval.}
    \label{fig:tolerance_result}
\end{figure}

We found several significant interaction effects for per-trial Decision Closeness and per-condition Confidence (see Figure \ref{fig:tolerance_result}). We performed contrast t-tests for specific differences identified. When viewing explanations that Show attribution uncertainty, participants who were more Uncertainty Tolerant had higher Decision Closeness (agreement with the AI system) than those who were less tolerant (M=$-2.11$ vs. $-3.58$, ${\rm p=.0007}$). This suggests that uncertainty intolerant users are less trusting of AI when they see that the AI is uncertain. When viewing Baseline explanations that do not show or suppress attribution uncertainty, Confidence in decision making is higher for more Uncertainty Tolerant users than less tolerant ones (M= $1.81$ vs. $0.64$, ${\rm p=.0006}$). This suggests that uncertainty tolerant users become more confident when viewing the AI’s explanation, though without uncertainty in its explanations; Baseline explanations do not raise the confidence of uncertainty intolerant users. However, for uncertainty intolerant users, they had significantly higher confidence using Suppress compared to Baseline (M=1.86 vs. 0.64, p$=$.0001), suggesting that the suppression helped make them more decisive. \par
\subsection{Summary of Qualitative and Quantitative Results}
\label{subsection:sum_qualitative_quantitative}
We summarize key findings unified from our qualitative and quantitative user studies. Compared to baseline explanations, showing attribution uncertainty helps users to be more aware of and understand internal attribution uncertainty and, in general, does not increase model (score output) uncertainty or decrease decision confidence, though it requires more decision time. However, this improvement in understanding did not improve user decision correctness or tendency to copy system predictions (decision closeness). Although, on average, user trust in and perceived helpfulness system predictions with shown attribution uncertainty was the same as for baseline explanations, our interviews revealed  divergent opinions regarding trust and helpfulness. From interviews, we learned that users who rated more positively appreciated understanding the relationship between input uncertainty and output uncertainty, while users who rated less positively were concerned about knowing more sources of high uncertainty. We also saw a similar divergence in our quantitative analysis that shown attribution uncertainty, users who were more uncertainty tolerant were more likely to copy the system score than users who were less tolerant. This suggests that uncertainty intolerance can lead to more distrust in predictions when more details of uncertainty are shown.\par

Compared to baseline explanations, suppressing attribution uncertainty helps to improve user decision quality, correctness, and confidence, without costing more decision time. Users found it more helpful and trustworthy too, because of the visibly reduced model (output) uncertainty and reduced concern regarding uncertain inputs due to their smaller attribution. Many users were appreciative of the "compensation" to re-allocate attributions due to the suppression. With suppressed attributions, showing the attribution uncertainty (ShowSuppress) did not suffer from performance and perception issues of show unsuppressed uncertainty, because users were less worried about the small uncertainties. Thus, showing suppressed attributions with uncertainty did not even cost more decision time than using baseline explanations. Finally, through interviews, we found interest in personalizing the extent of suppression to see a more subdued effect or to customize based on applications.

\section{Design Implications: Show or Suppress?}
\label{section:design_implication}
In this work, we have investigated uncertainty that is propagated into feature attributions and compared two alternative approaches of showing uncertainty within explanations or suppressing attributions to make explanations less reliant on uncertain input. We note that there are many needs for a variety of explanations and deeper mechanistic explanations as demonstrated in our interview results and in literature \cite{Ribeiro2016WhyClassifier,Dodge2019,lai2020chicago,Cheng2019ExplainingStakeholders,ghai2020explainable}. Our approaches are complementary to these explanations, but further study is needed to investigate the impact of showing or suppressing uncertainty in those explanations. Here, we discuss recommendations for whether to show or suppress attribution explanation uncertainty in explanations based on findings from our simulation experiments and user studies.

\subsection{Communicating Explanation Attribution Uncertainty}
\label{subsection:benefit_uncertainty_attribution}
Much research to visualize uncertainty has focused on helping users to identify uncertain inputs \cite{McCurdy2019AVisualization,Correa2009AAnalytics}, but this requires users to apply their domain expertise to understand how they relate to the analysis task. With machine learning models or other automation systems, research on communicating uncertainty has focused on the model, inference or outcome uncertainty, typically expressed as model confidence \cite{Lim2011InvestigatingApplications,Yin2019UnderstandingModels,Kay2016WhenSystems}. However, these show an overall uncertainty due to a single decision and does not provide additional details for the source of uncertainties. Communicating both input uncertainty and model uncertainty can provide more insights to users, but the relationship between input and model uncertainty remains unclear. In this work, we have proposed and demonstrated the benefits and issues of communicating uncertainty \textit{within} an explanation model to provide more transparency uncertainty. We found that users could understand the model better by being able to trace how uncertainty propagates from input, to attributions, to model output. Therefore, attribution uncertainty can provide additional insight into the internal workings of AI models.

%{\color{red}I DO NOT SEE HOW THIS PARAGRAPH IS RELEVANT. ---
%Research on machine learning model stability, robust learning, and Bayesian inference seeks to exploit uncertainty to improve model predictive performance \cite{doshi2017towards,gosiewska2019ibreakdown,koh2017understanding,zhang2019should}. Although there are some works on the explanation uncertainty, they are mainly about the instability of explanation methods and are used by AI developers to inspect and evaluate AI methods. Instead, we call attention to communicating uncertainty within explanation to end users.}

% {\color{red}
% Discuss Input vs. Output vs. Attribution uncertainty.
% Attribution uncertainty is more useful and insightful than Input uncertainty. And how people would still want to adjust the Output uncertainty after seeing the Attribution uncertainty.
% There are papers talk about attribution uncertainty but they are all about the instability of explanation method not due to input uncertainty.}

\subsection{Show Attribution Uncertainty When Suppressed}
\label{subsection:show_uncertainty_when_suppressed}
A key recommendation from our results is to show attribution uncertainty when it is suppressed to be low. Otherwise showing high attribution uncertainty would worry the users, increase the uncertainty of their decision, reduce the confidence in their decisions, slow down their decision making, and ultimately limit their trust towards the system. Our findings are similar to Lim and Dey \cite{Lim2011InvestigatingApplications} who found that when model uncertainty is low, showing explanations improves user trust, but when model uncertainty is high, showing explanations hurts user trust. In our work, instead of requiring low uncertainty, our proposed Suppress technique can reduce model uncertainty to low enough levels and achieve the gains in perceived helpfulness and trust towards the system. Though, we recommend being judicious in how much suppression to exercise, and to consider allowing users to interactively control the suppression level.

%Despite the benefits of suppressing uncertainty, our findings only limit to the cases when the system prediction is correct. If the system can be wrong, when uncertainty is suppressed, users may experience an association fallacy on the reduced uncertainty and misattribute improved decision making of the model. Suppressing uncertainty can also cause overconfidence and over-trust. Readers can refer to these works \cite{Bansal2019, Kaur2020, Lim2011InvestigatingApplications} for the scenario when the system can be wrong. Nevertheless, when uncertainty is suppressed by regularizing predictor, the predictor model is more robust to noise, and it can also be more accurate in prediction (See Figure \ref{fig:other_results}), thus the cases of wrong system prediction become rarer.

% Another limitation of suppressing uncertainty is that the Regularized Explainer method can be misleading when the users task is to predict the system output. When the uncertainty is only suppressed in the explainer model, users would incorrectly believe that the underlying predictor model is also robust to noise, even though this is not true. In this case, the Regularized Predictor method is recommended.
We note that with ShowSuppress explanations, some users may be misled to think that some input features would not affect the predictor model output much. We highlight that baseline (Show) and regularized (ShowSuppress) explanations serve different purposes. Baseline explanations aim to faithfully represent the Predictor for each instance even with the noise $\epsilon$, i.e., $g_f(x+\epsilon) \approx f(x+\epsilon)$, and provide representative feature weights of how the Predictor thinks. In contrast, Regularized Explainers prioritize a more robust explanation to discount the effect of uncertain features, i.e., $\tilde{g}_f(x+\epsilon) \approx f(x)$, and provide an interpretation that defers influences to more reliable features. Under uncertainty, even though the Predictor could produce an ambiguous prediction, ShowSuppress would inform the user with a more confident explanation that we have shown to be more accurate on average. To ameliorate the confusion between the two explanation types, users should be reminded of the distinction between Show and ShowSuppress explanations, i.e., $g_f(x+\epsilon) \neq \tilde{g}_f(x+\epsilon)$. They should be shown together as we have done, with Show as the primary explanation and ShowSuppress as supplementary. Providing explanations from a Regularized Predictor will avoid this issue, but would require Predictor models to be retrained.
Furthermore, while ShowSuppress explanations increased decision quality for correct system predictions, they may not increase decision quality for wrong predictions, since users may not be able to perceive uncertainty to support counterfactual reasoning. Similar to prior works \cite{Bansal2019, Kaur2020, Lim2011InvestigatingApplications}, we expect that ShowSuppress explanations would also lead to inappropriate trust for wrong predictions and defer a fuller investigation to future work.

\subsection{Alternative Visualizations of Attribution Uncertainty}
\label{subsection:visualization_generalization}
Among the many ways to visualize uncertainty \cite{hullman2015hypothetical,Kay2016WhenSystems}, in this work, we have employed the violin plot and confidence interval ($\pm$number) to show attribution uncertainty. On the one hand, we demonstrated augmenting a tornado plot with many violin plots to visualize details of each uncertainty distribution. This is informative for savvy users with good technical, statistical, and graph literacy. On the other hand, we evaluated the confidence interval number with lay users online to ensure simple, quick interpretation, since this does not directly require graph literacy or familiarity with probability distributions. In pilot studies, with participants screened for graph literacy, we found that users of violin-tornado plots had higher trust in the system than users of baseline tornado plots. This discrepancy from our current results could be because these users could appreciate the mathematical detail provided in these explanations and found them more trustworthy. Future work can investigate showing attribution uncertainty to different target users with different user-friendly uncertainty visualization methods, such as quantile dot plots \cite{Kay2016WhenSystems} or hypothetical outcome plots (HOP) \cite{hullman2015hypothetical}.

\section{Conclusion}
\label{section:conclusion}
We have highlighted and investigated how attribution explanation uncertainty can impact explanations in machine learning, particularly in terms of user trust, confidence, and decision-making. Informed by different uncertainty coping strategies, we proposed two techniques to manage attribution explanation uncertainty: showing uncertainty in attribution explanations and suppressing  attribution uncertainty by reducing and reallocating attributions. We conducted a simulation study, qualitative interviews, and quantitative user evaluation to compare these techniques with baseline explanations. We found that showing attribution uncertainty helps with user understanding of the prediction models, but does not improve trust, confidence and decision making, yet slows down decision making; suppressing uncertainty reduces decision uncertainty, improves trust towards prediction models, user confidence in decision making, and decision quality. This demonstrates the importance to carefully communicate uncertainty \textit{within} attribution explanations to improve the trustworthiness of artificial intelligence systems. \par
\section{Acknowledgement}
 We thank Ashraf Abdul, Dr. Lyu Yan, Dr. Zhang Yehong and Dr. Wang Shengyu for their assistance in the discussion, piloting study, providing language help and proof-reading. We thank all our participants for their dedication of time and precious feedback. This work was supported in part by the Ministry of Education, Singapore and was carried out at the NUS Institute for Health Innovation and Technology (iHealthtech) under grants T1 251RES1804 and R-722-000-004-731.\par

\newpage
\bibliographystyle{elsarticle-harv} 
\bibliography{paper}

\newpage
\appendix
\label{Appendix}
\renewcommand{\figurename}{Appendix Figure}
\section{Proof of Baseline Explainer’s Expected Faithfulness}
\label{append:proof}
Here we prove that the expected faithfulness distance $E[F_{g_f}]$ of baseline LIME explainer $g_f$ over noisy input $x+\epsilon$ is the same as or worse than point-estimated baseline faithfulness distance $F_0$.\\

\begin{proof}
$E_{\epsilon}[F_{g_f}]\geq F_0$.
\begin{equation*}
\begin{aligned}
F_{g_f}-F_{0}&=\left(g_{f}(x+\epsilon)-f(x)\right)^{2}-\left(g_{f}(x)-f(x)\right)^{2}\\
&=\left(g_{f}(x+\epsilon)+g_{f}(x)-2 f(x)\right)\left(g_{f}(x+\epsilon)-g_{f}(x)\right) \\
&=\left(w^\top x+w^\top \epsilon+w^\top x-2 f(x)\right)\left(w^\top x+w^\top \epsilon-w^\top x\right) \\
&=\left(2 w^\top x-2 f(x)+w^\top \epsilon\right) w^\top \epsilon \\
&=2\left(w^\top x-f(x)\right) w^\top \epsilon+w^\top w \epsilon^\top \epsilon^\top,\\
E_{\epsilon}\left[F_{g_f}-F_{0}\right]&=2\left(w^\top x-f(x)\right) w^\top E_{\epsilon}[\epsilon]+w^\top w E_{\epsilon}\left[\epsilon^\top \epsilon\right],\\
\text{Since } \epsilon^{(d)} &\sim \mathcal{N}\left(0,\left(\sigma^{(d)}\right)^{2}\right),\\
E[F_{g_f}]-F_{0}&=E_{\epsilon}\left[F_{g_f}-F_{0}\right]=0+w^\top w \sigma^\top \sigma \geq 0.
\end{aligned}
\end{equation*}
\end{proof}

% \newpage
\section{Other Measures of Regularized Explanations}
\label{append:other_sim}
\setcounter{figure}{0}
\begin{figure}[H]
    \centering
    \includegraphics[width=8cm]{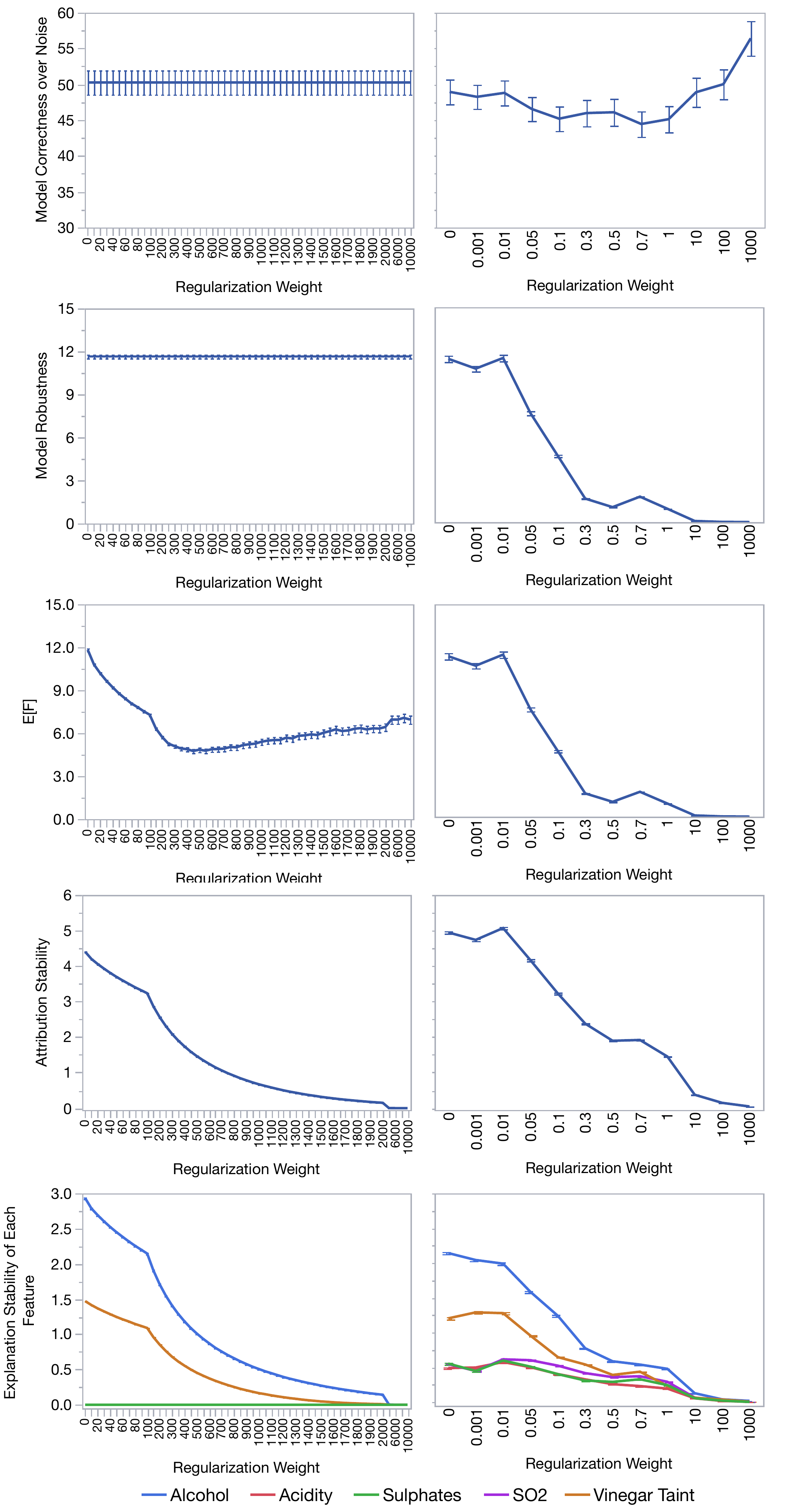}
    \caption{Other measures of Regularized Explainer (Left) and Regularized Predictor (Right) in simulation study. X-axis is the regularization weight $\lambda$. Regularized Explainer does not affect model correctness over noise or model robustness while Regularized Predictor can improve them. Both regularized methods improve expected faithfulness and attribution stability under input uncertainty. Smaller value on y-axis is better.}
    \label{fig:other_results}
\end{figure}

Appendix Figure \ref{fig:other_results} demonstrates how regularization weight $\lambda$ influences measurements on model prediction and explanation, including model prediction correctness over noise, model prediction robustness, explanation expected faithfulness and attribution explanation stability.\par
Here we describe how the measures are defined for an instance, and the results in \ref{append:other_sim} are the average of these measures over all instances in the wine data set. The model correctness over noise of an instance is defined as the mean squared error between model prediction and ground truth over all hypothetical instances drawn from the input uncertainty distribution of the instance. Model prediction robustness of an instance is defined as the mean squared error between the model prediction of the instance and those of hypothetical instances draw from the input noise distribution. Explanation expected faithfulness distance is the same as the definition in Section \ref{sec:exp_faith}. Attribution explanation stability of an instance is the sum of the standard deviation of each feature attribution over the hypothetical instances. The fourth row in Appendix Figure \ref{fig:other_results} is the sum of explanation stability over all features, and in the last row different features’ attribution explanation stability is overlaid. All the error bars represent the standard error of the average over all instances. For all these measures, the smaller the values are the more correct, faithful, robust or stable the method is.\par
The left column in Appendix Figure \ref{fig:other_results} shows that when increasing the regularization weight $\lambda$ of the explainer, model correctness and robustness are not affected because LIME explainer is post-hoc and model-agnostic. The expected faithfulness distance of explanation increases with $\lambda$, and when $\lambda$ becomes too big, the explainer's faithfulness distance gets worse. The stability of attribution keeps increasing with $\lambda$.\par
The right column illustrates that the model correctness of Regularized Predictor first improves with $\lambda$ and then gets worse when $\lambda$ becomes too big. Model robustness, explanation’s expected faithfulness distance and stability keep increasing with $\lambda$. \par
\newpage
\section{User Interface of Survey}
\label{append:user_interface}
This appendix shows the various pages in the quantitative survey implemented in Qualtrics for Amazon Mechanical Turk workers as participants. Note that branching logic was used to manage the random assignment to different conditions and trial sequences. Each figure illustrates a page in the survey.

% \label{append:welcome}
\setcounter{figure}{0}
\begin{figure}[!ht]
    \centering
    \includegraphics[width=13cm]{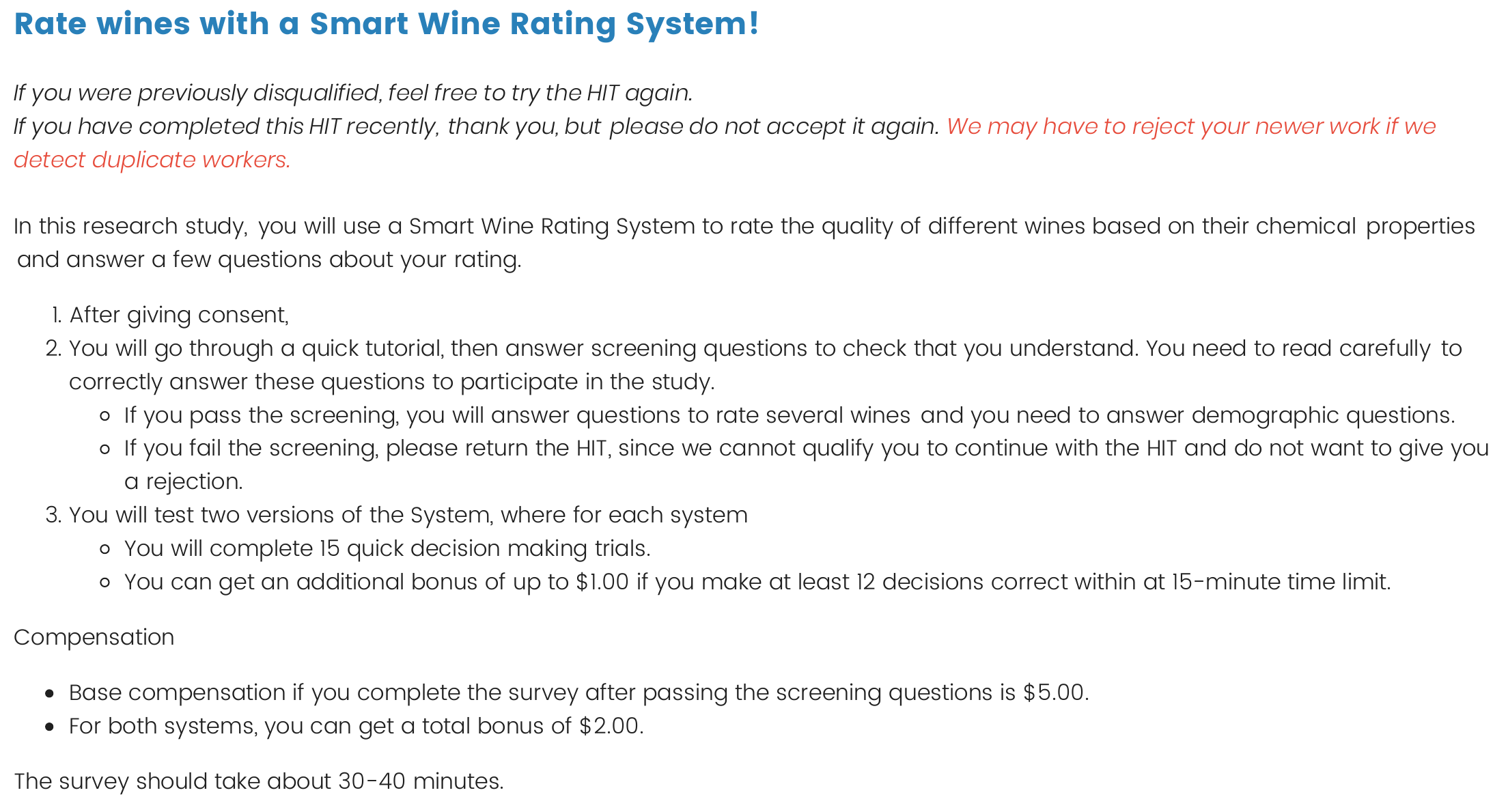}
    \caption{Welcome page introducing the tasks in the survey and compensation with bonus incentive.}
    \label{fig:welcome}
\end{figure}

% \label{append:task_description}
\setcounter{figure}{1}
\begin{figure}[!ht]
    \centering
    \includegraphics[width=13cm]{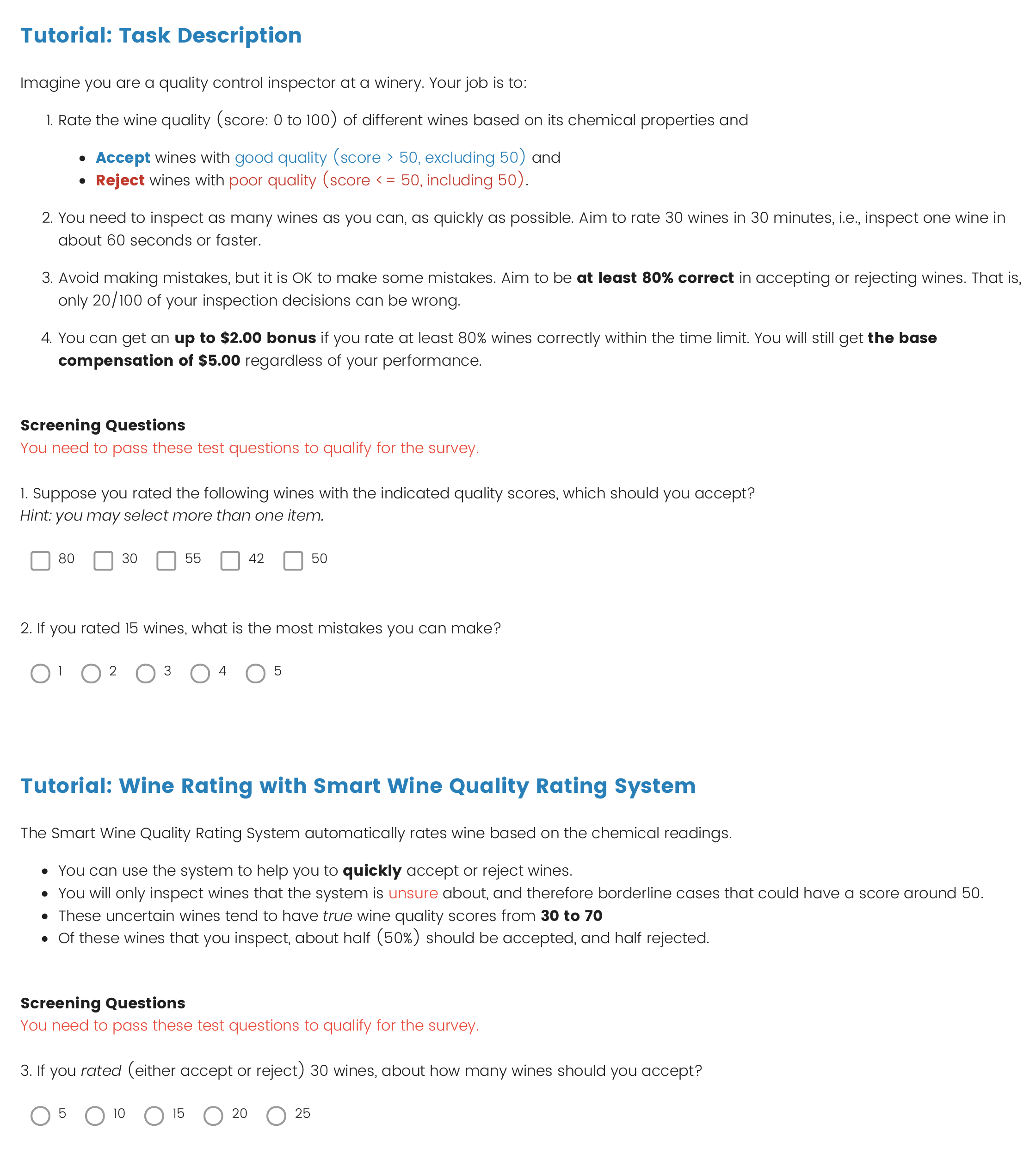}
    \caption{First tutorial on task description and introduction to the AI system with screening questions.}
    \label{fig:task_description}
\end{figure}

\setcounter{figure}{2}
\begin{figure}[!ht]
    \centering
    \includegraphics[width=13cm]{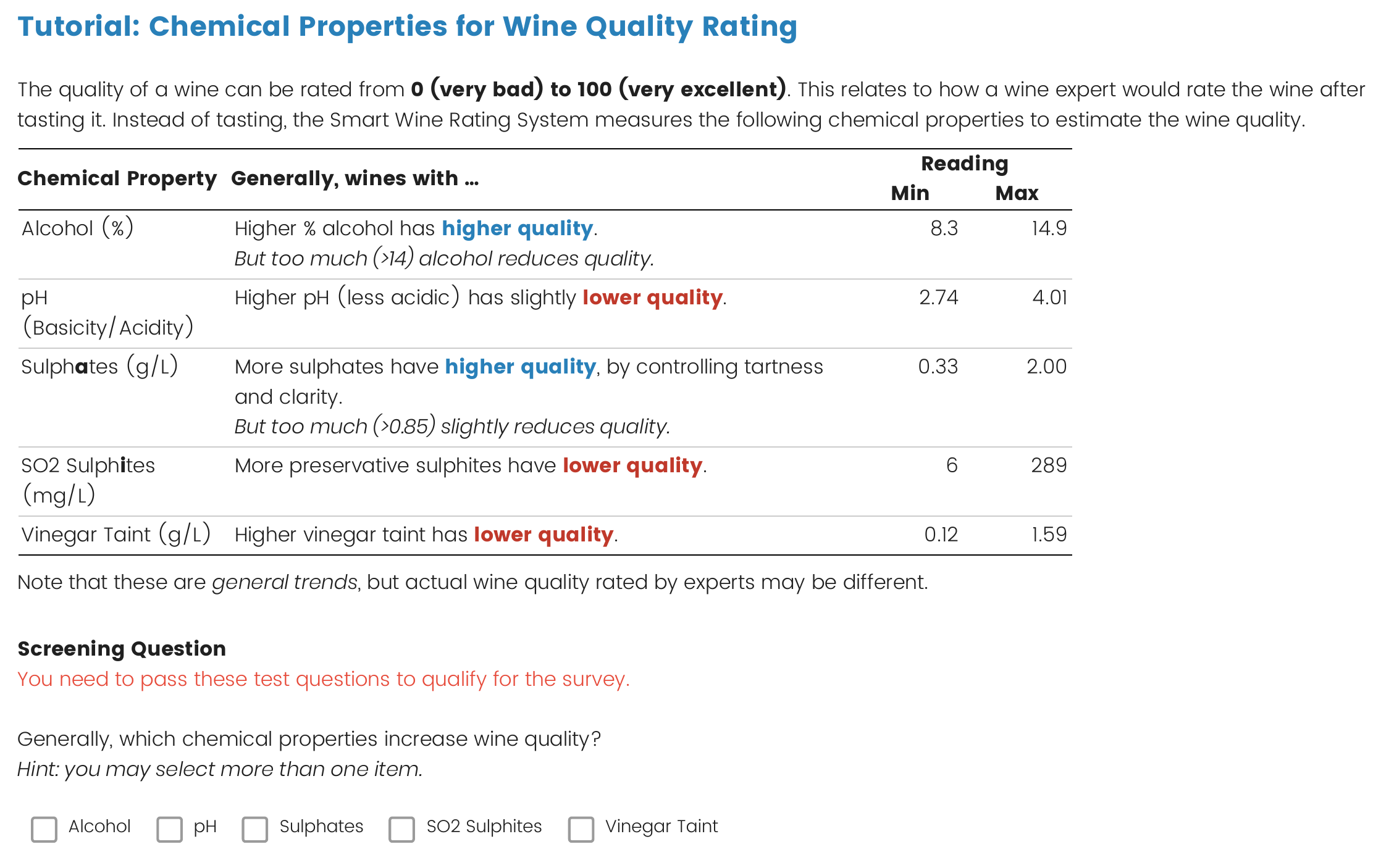}
    \caption{Tutorial on background knowledge on estimating wine quality rating based on chemical properties, with screening question.}
    \label{fig:tutorial_background}
\end{figure}

% \label{append:tutorial_reading+scores}
\setcounter{figure}{3}
\begin{figure}[!ht]
    \centering
    \includegraphics[width=13cm]{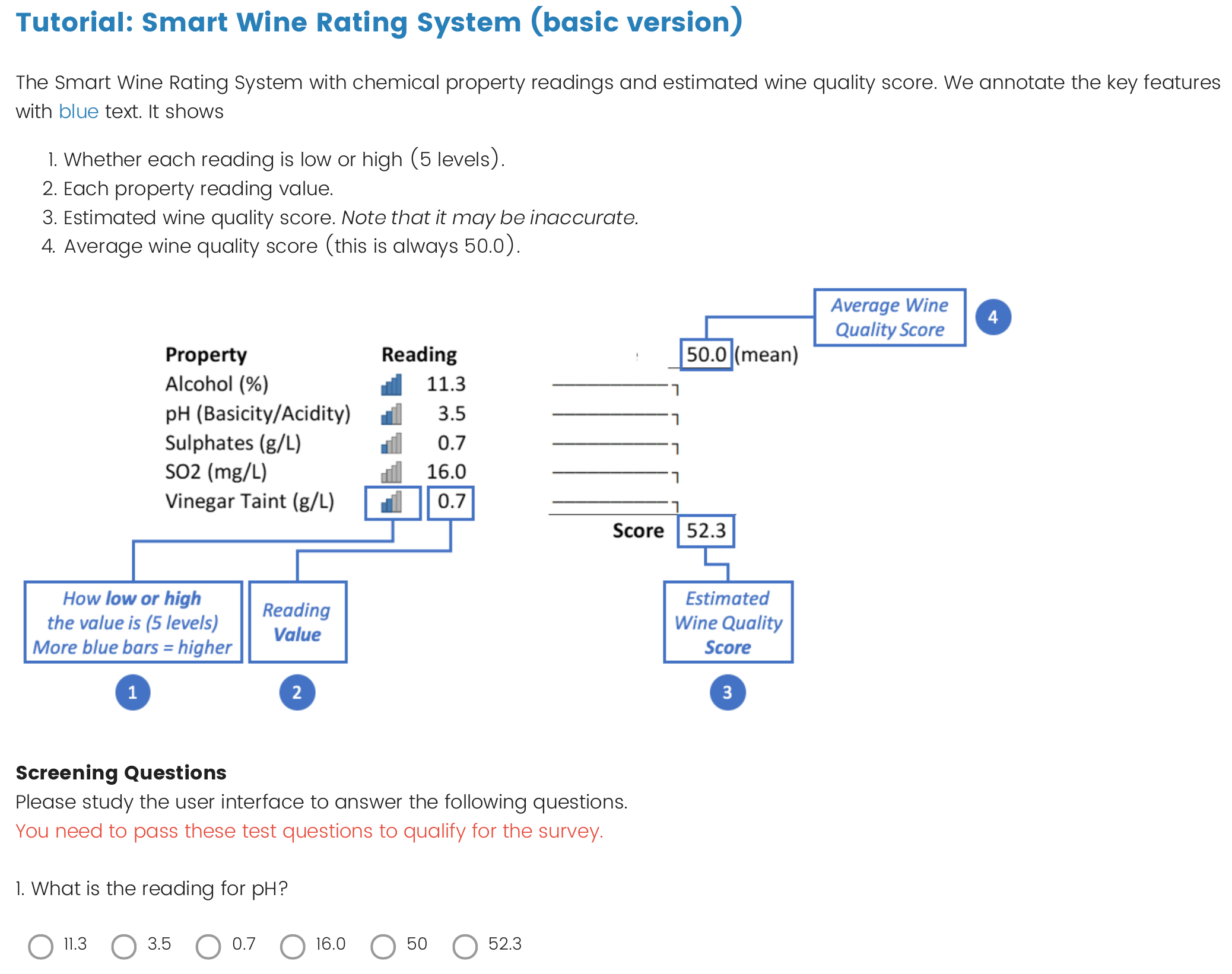}
    \caption{First tutorial on AI system, regarding the basic version with no explanation (None), with screening question.}
    \label{fig:tutorial_reading+scores}
\end{figure}

% \label{append:tutorial_noexp}
\setcounter{figure}{4}
\begin{figure}[!ht]
    \centering
    \includegraphics[width=13cm]{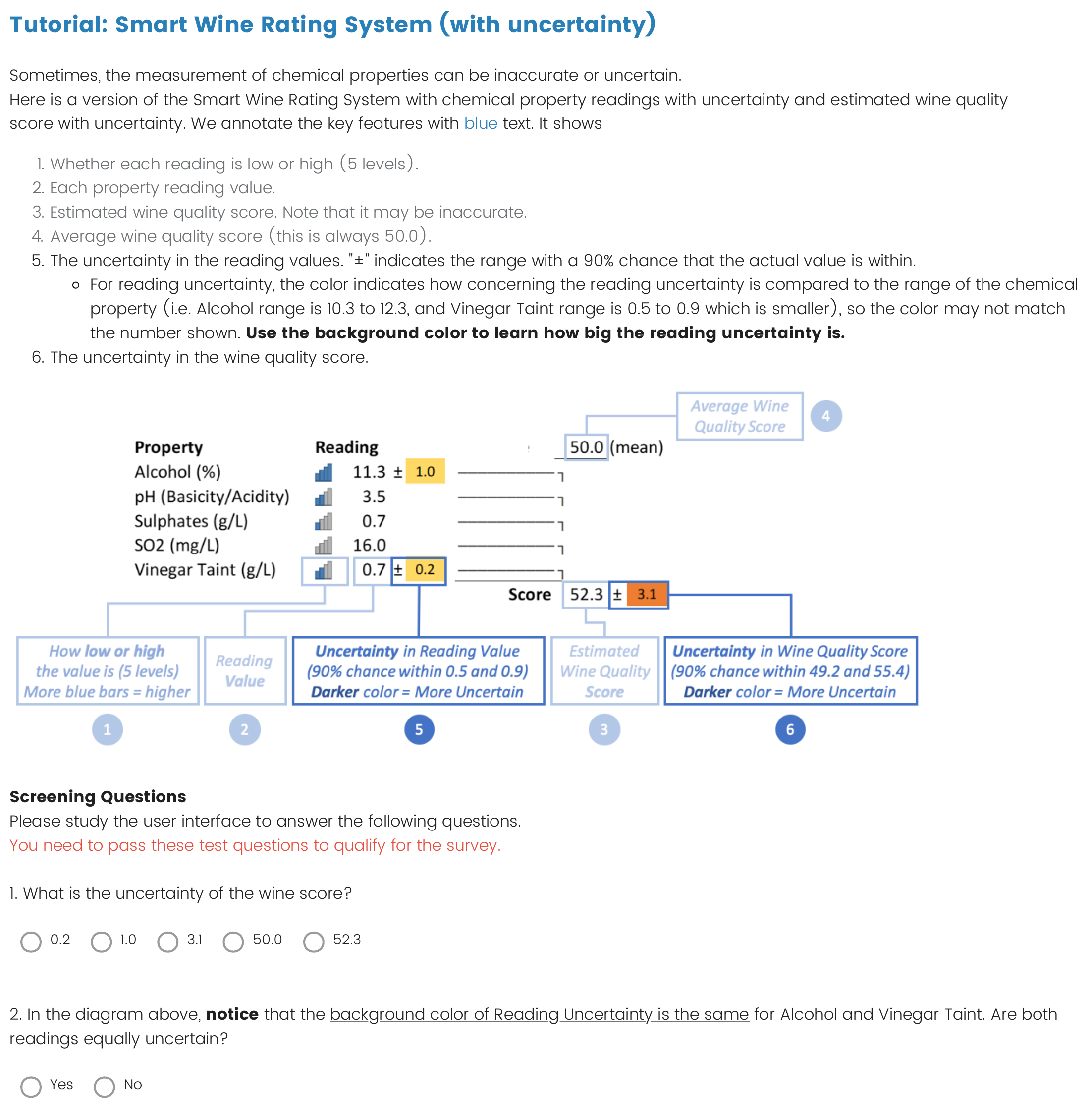}
    \caption{Second tutorial on AI system, regarding the basic verison with no explanation (None), now also discussing uncertainty in input readings, with screening question.}
    \label{fig:tutorial_noexp}
\end{figure}

\setcounter{figure}{5}
\begin{figure}[!ht]
    \centering
    \includegraphics[width=13cm]{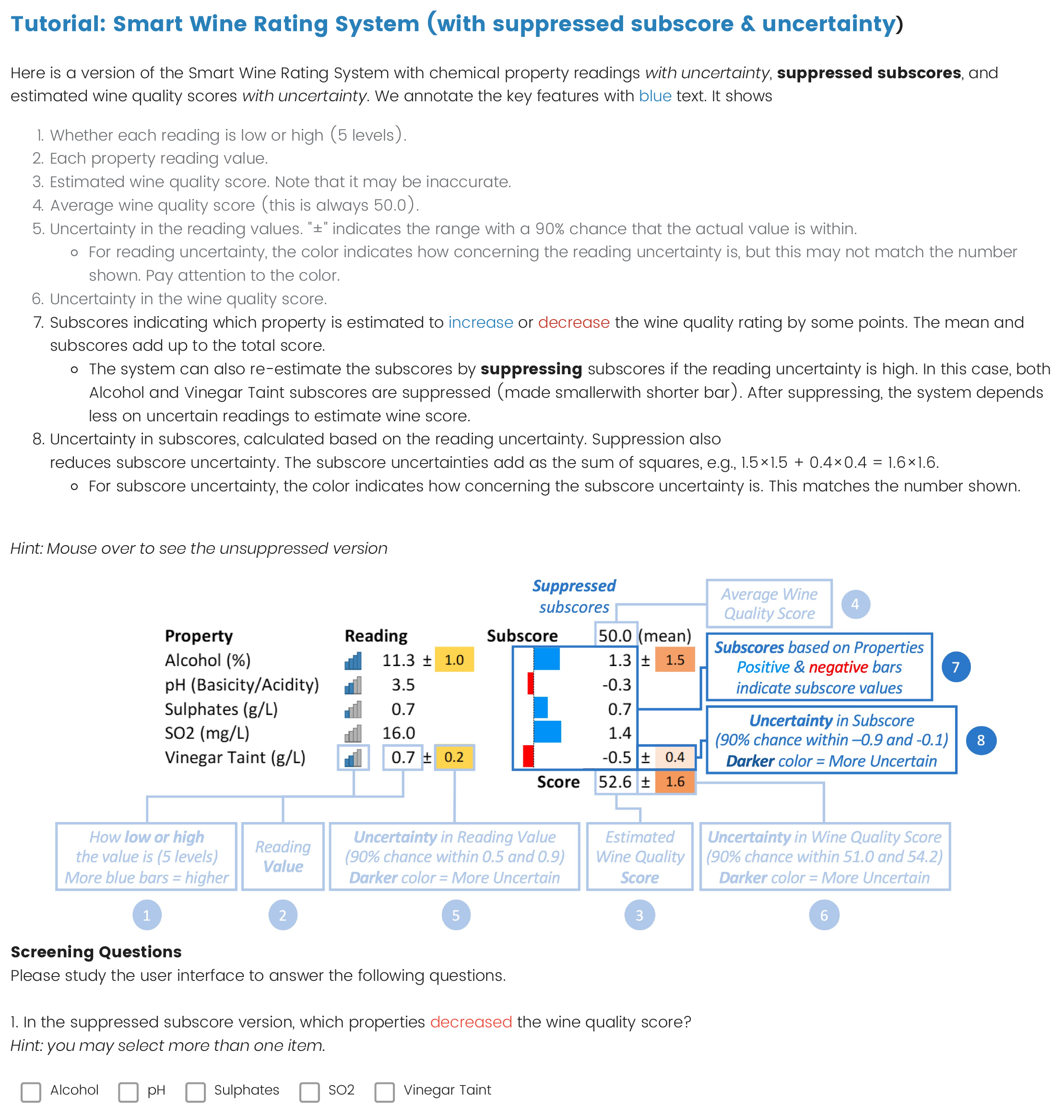}
    \caption{Pre-Condition tutorial on the system user interface showing relevant explanation components (ShowSuppress version shown with all components). Participants are incrementally introduced to different UI components to facilitate a gentle introduction. Questions are for manipulation check of comprehension, but not used for screening.}
    \label{fig:tutorial_showsuppress}
\end{figure}

% \label{append:uncertainty_strategy}
\setcounter{figure}{6}
\begin{figure}[!ht]
    \centering
    \includegraphics[width=13cm]{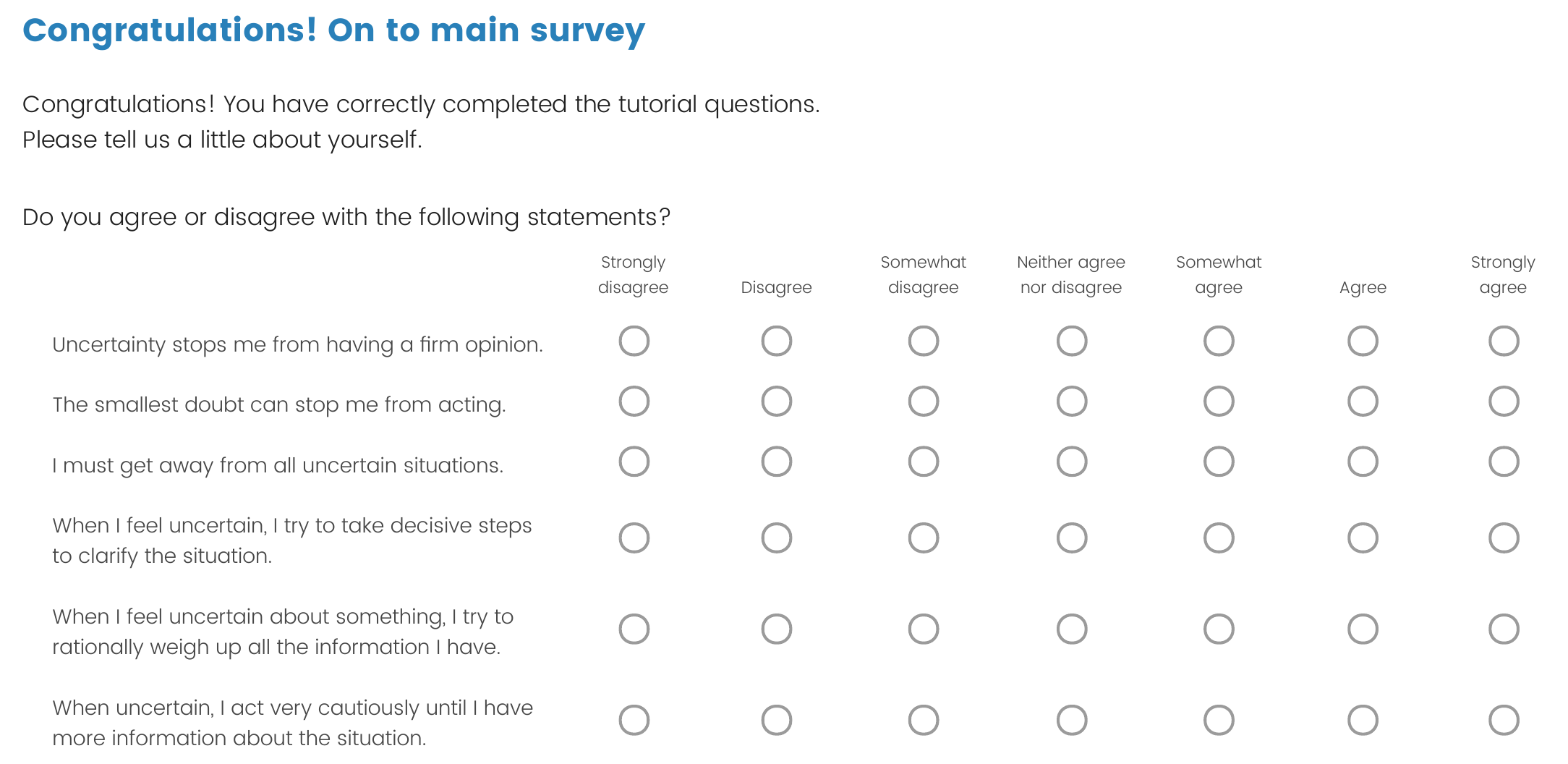}
    \caption{Post-screening qualification page with survey on Uncertainty Intolerance to measure confound of user personality background. This is asked before any experiment trials to avoid measuring a change in opinion due to performing potentially difficult tasks.}
    \label{fig:uncertainty_strategy}
\end{figure}

% \label{append:screening_failure}
\setcounter{figure}{7}
\begin{figure}[!ht]
    \centering
    \includegraphics[width=13cm]{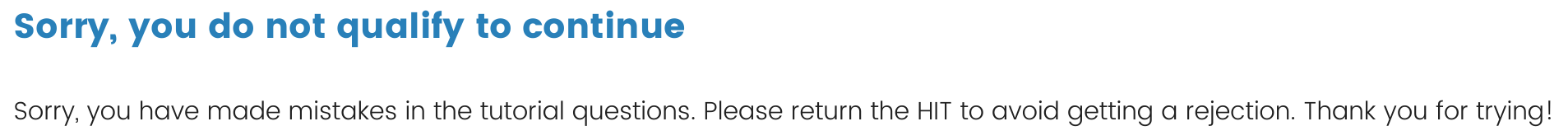}
    \caption{Post-screening disqualification page for participants who had <5 out of 7 correct answers to screening questions.}
    \label{fig:screening_failure}
\end{figure}

\setcounter{figure}{8}
\begin{figure}[!ht]
    \centering
    \includegraphics[width=13cm]{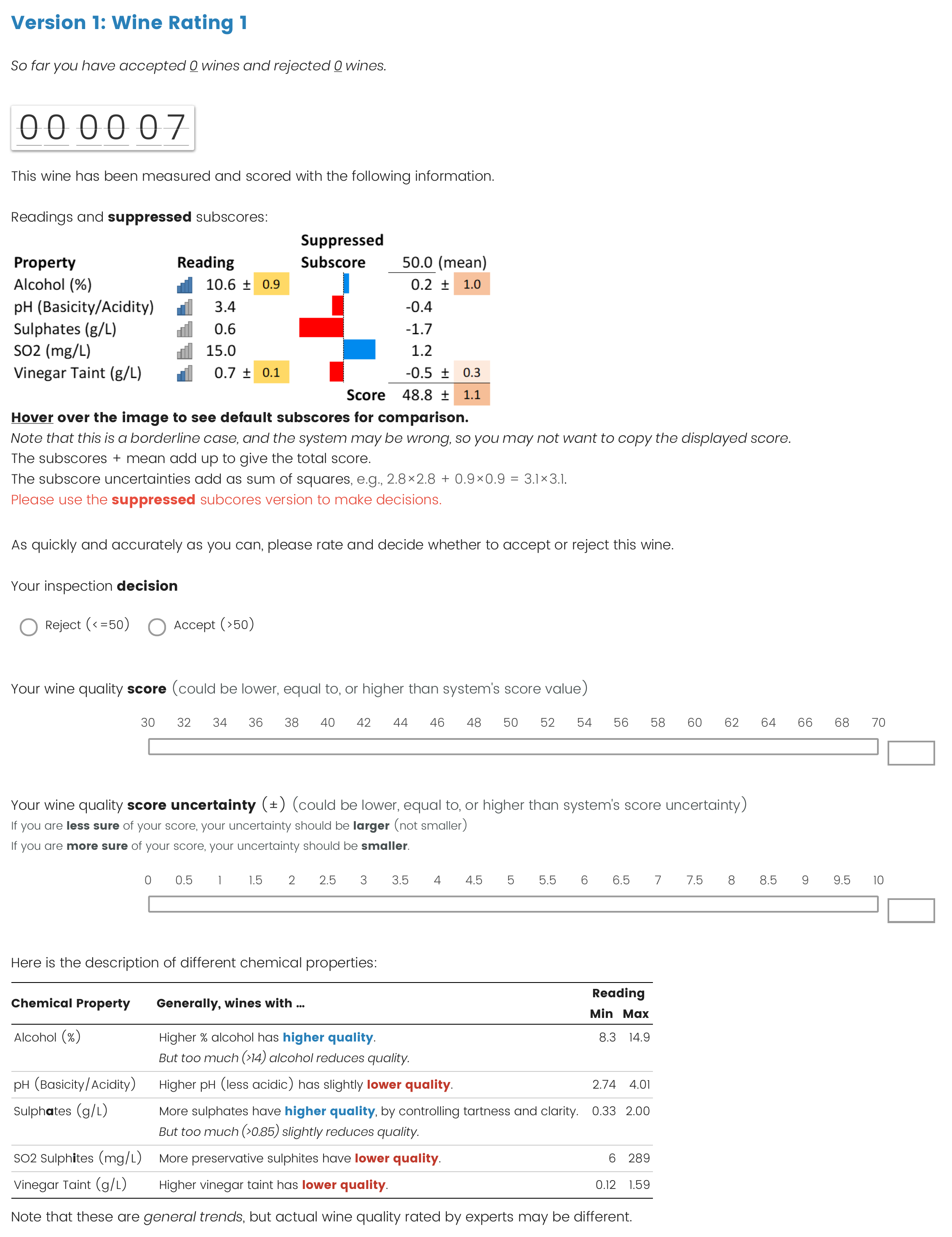}
    \caption{First page of a trial task (same for practice and main trials) showing timer, system user interface (UI), and three questions regarding inspection decision, score value, and score uncertainty. The UI is slightly different for different Explanation Technique conditions, showing the corresponding explanation components. The background table is included for reference.}
    \label{fig:main_trial}
\end{figure}

% \label{append:trial_feedback}
\setcounter{figure}{9}
\begin{figure}[!ht]
    \centering
    \includegraphics[width=13cm]{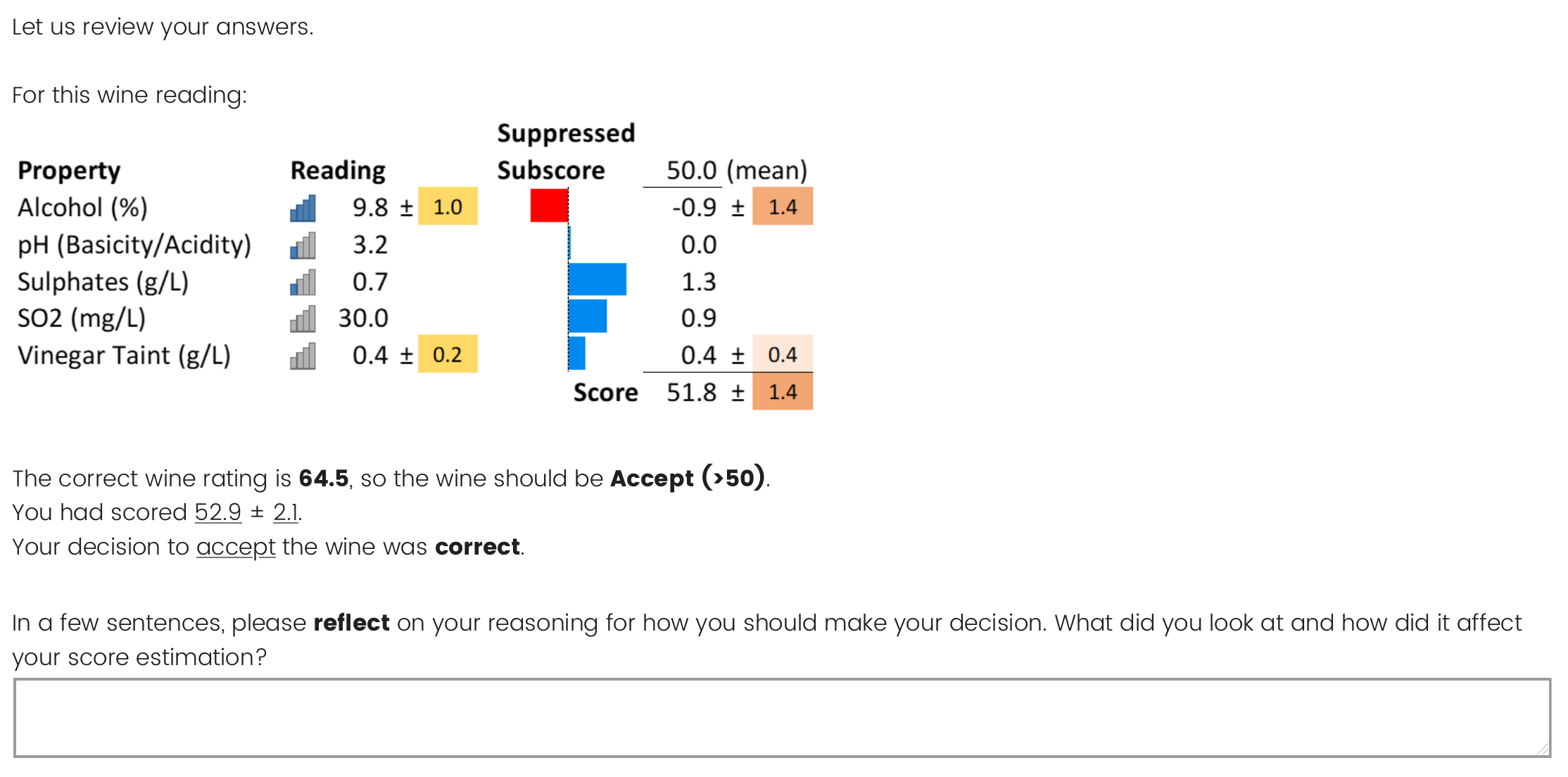}
    \caption{Second page of a Practice trial task showing the participant's answers as a reminder and asking about their rationale in decision and rating estimation. This page was not included for the main trial. It serves two purposes: i) manipulation check to ensure that users are paying attention and reasonably understand the user interface, ii) stimulate users to self-explain to raise their level of understanding before proceeding to the main trials.}
    \label{fig:trial_feedback}
\end{figure}

% \label{append:post_trial}
\setcounter{figure}{10}
\begin{figure}[!ht]
    \centering
    \includegraphics[width=13cm]{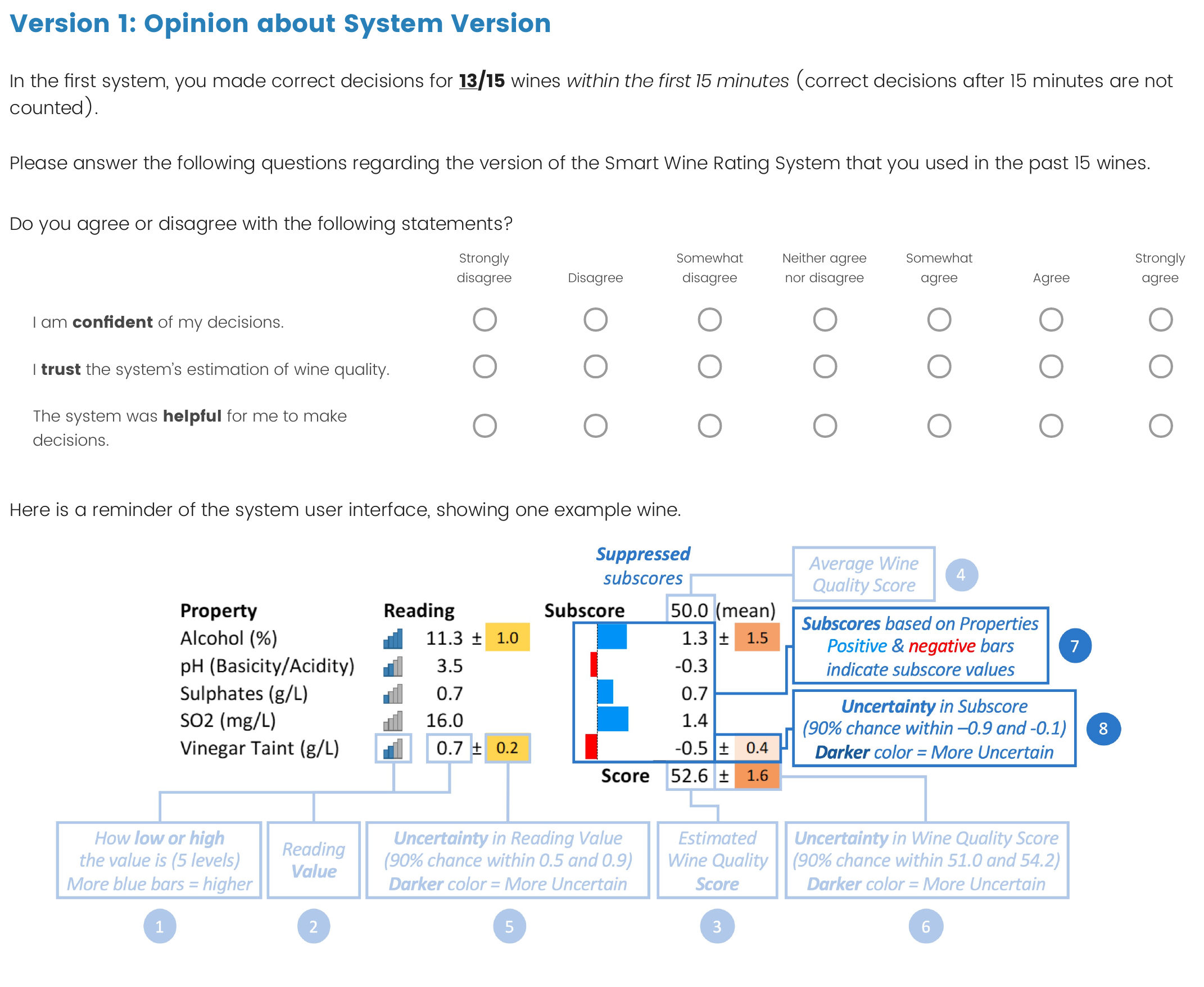}
    \caption{Post-Condition (after main trials) survey page with rating questions about the overall system.}
    \label{fig:post_trial}
\end{figure}

% \label{append:demographics}
\setcounter{figure}{11}
\begin{figure}[!ht]
    \centering
    \includegraphics[width=13cm]{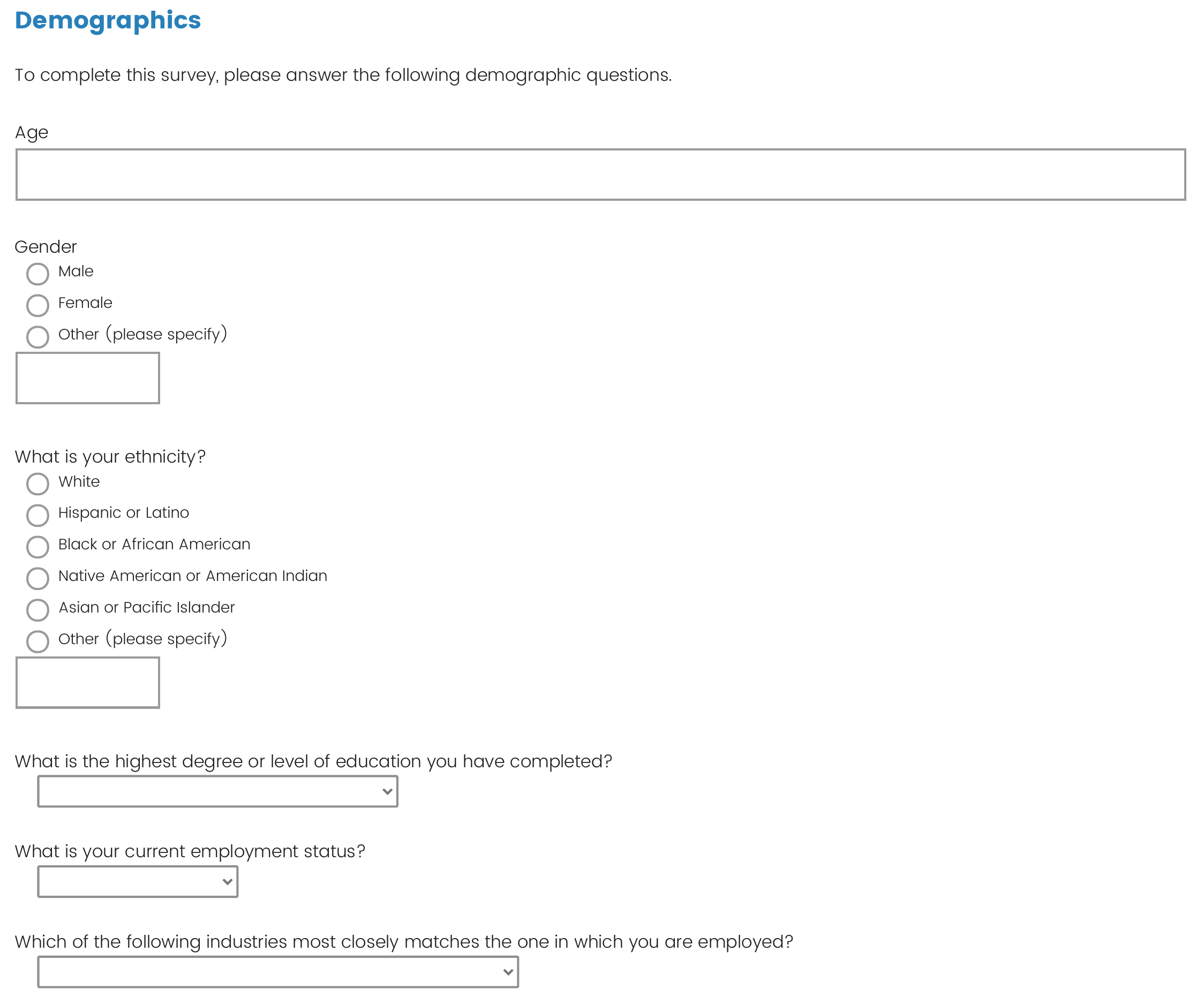}
    \caption{Final survey page with questions on demographics.}
    \label{fig:demographics}
\end{figure}

% \label{append:demographics}
\setcounter{figure}{12}
\begin{figure}[!ht]
    \centering
    \includegraphics[width=11cm]{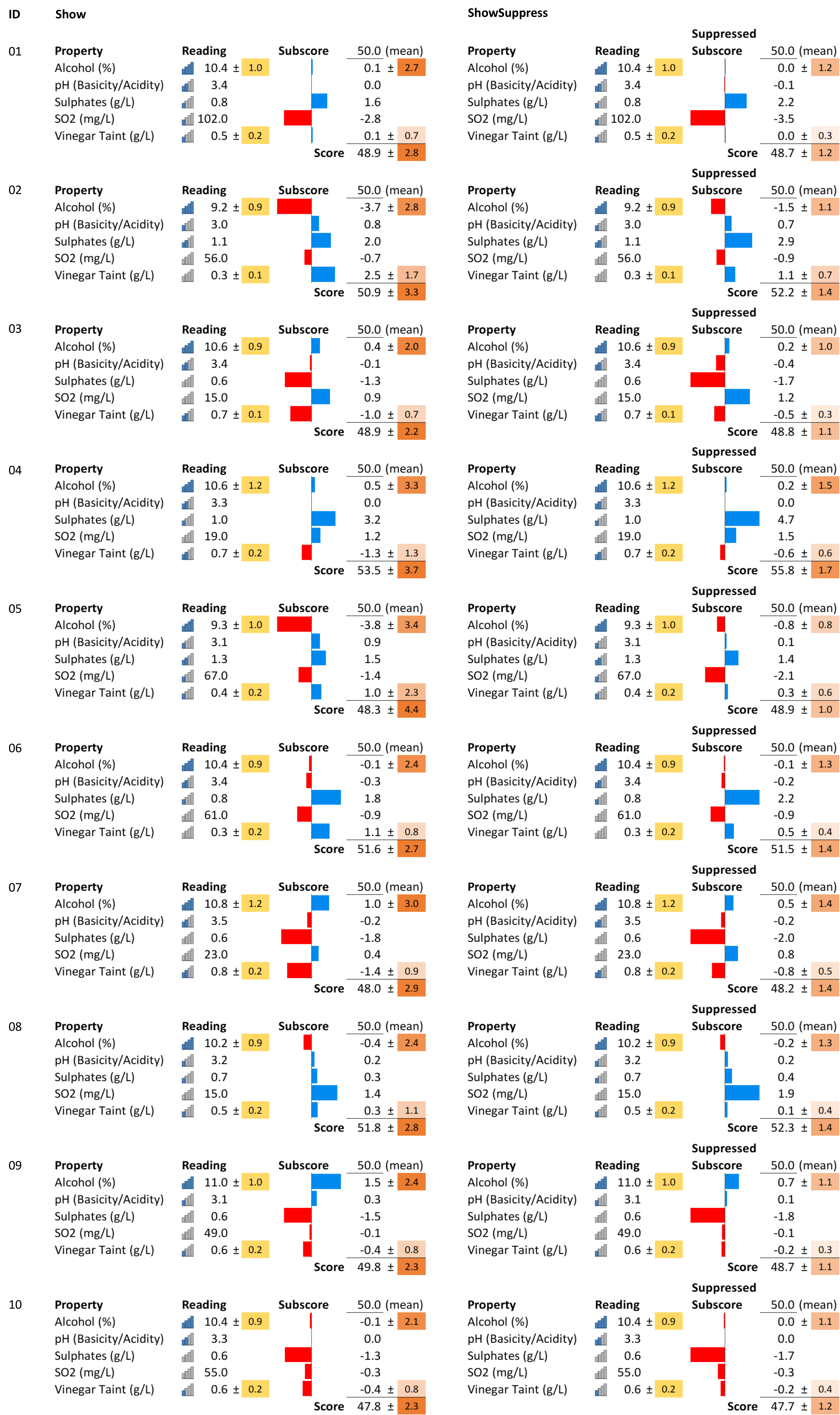}
    \caption{Instances used in user study. Here we show the interface of Show (left) and ShowSuppress (right) explanation.}
    \label{fig:all_instances_1}
\end{figure}

\setcounter{figure}{12}
\begin{figure}[!ht]
    \centering
    \includegraphics[width=11cm]{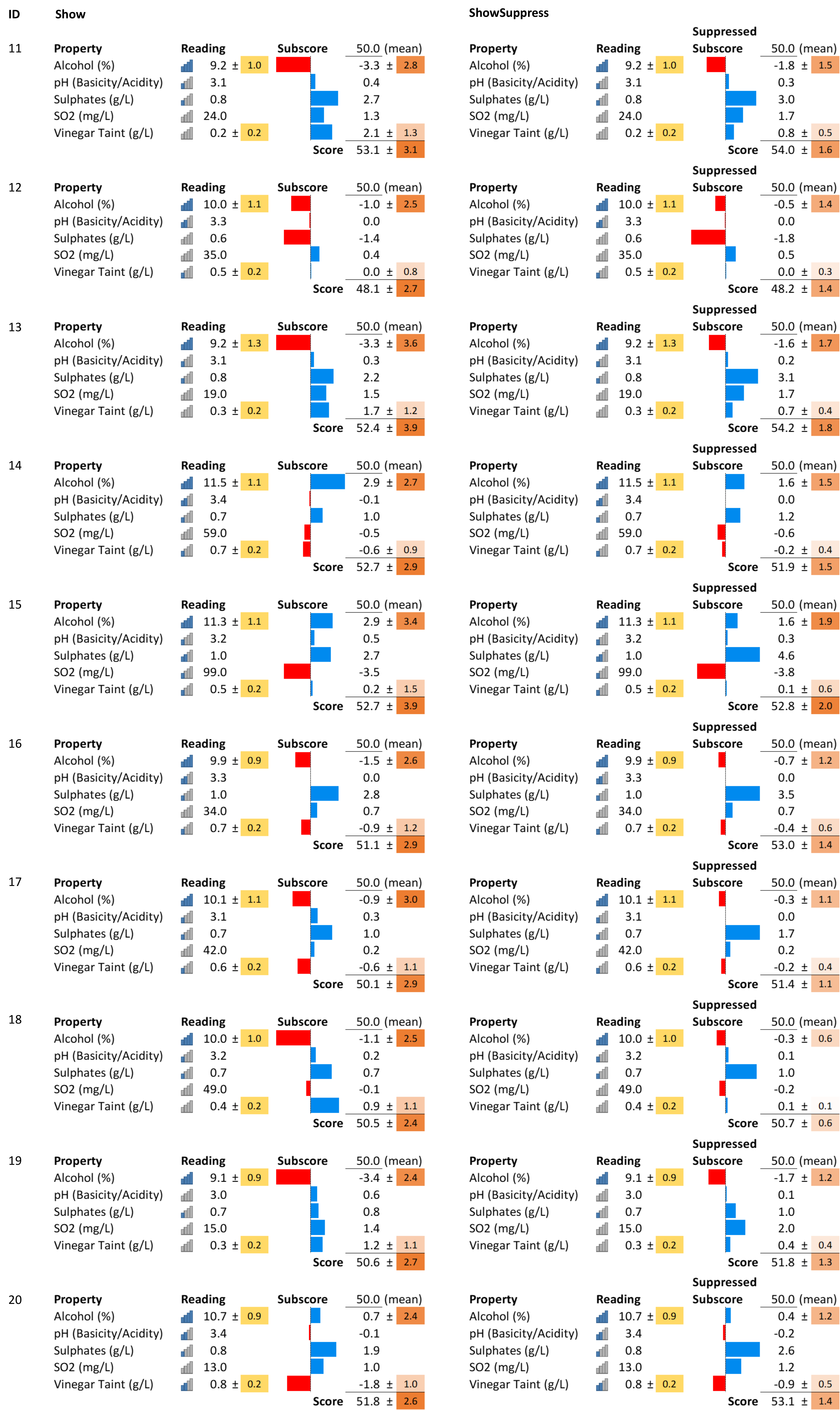}
    \caption{(continued) Instances used in user study.Here we show the interface of Show (left) and ShowSuppress (right) explanation.}
    \label{fig:all_instances_2}
\end{figure}

\setcounter{figure}{12}
\begin{figure}[!ht]
    \centering
    \includegraphics[width=11cm]{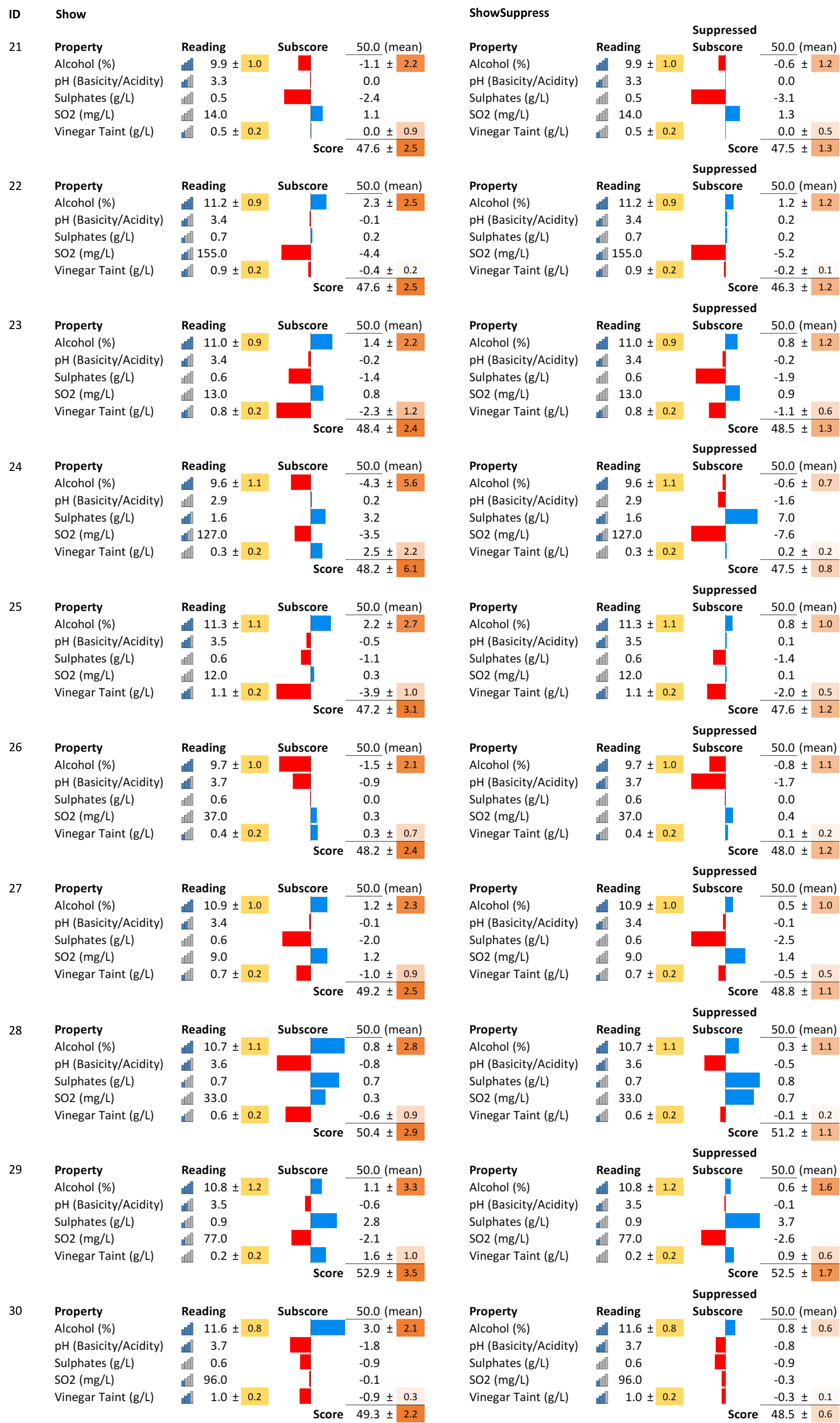}
    \caption{(continued) Instances used in user study.Here we show the interface of Show (left) and ShowSuppress (right) explanation.}
    \label{fig:all_instances_3}
\end{figure}

\end{document}